\documentclass[10pt,journal,letterpaper,compsoc]{IEEEtran}
\usepackage{caption}
\usepackage{times}
\usepackage{amsmath}
\usepackage{amssymb}
\usepackage{graphicx}
\usepackage{amsfonts}
\usepackage{subfig}
\usepackage{tabularx}

\usepackage{latexsym}
\usepackage{color}
\usepackage{multirow}

\usepackage{xspace}
\makeatletter
\DeclareRobustCommand\onedot{\futurelet\@let@token\@onedot}
\def\@onedot{\ifx\@let@token.\else.\null\fi\xspace}

\ifCLASSOPTIONcompsoc
   \usepackage[nocompress]{cite}
\else
\fi
\ifCLASSOPTIONcompsoc
\else
\fi
\ifCLASSINFOpdf
\else
\fi

\newcolumntype{C}[1]{>{\centering\let\newline\\\arraybackslash\hspace{0pt}}m{#1}}

\newcommand{\figref}[1]{fig.\ref{#1}}
\newcommand{\Figref}[1]{Fig.\ref{#1}}
\newcommand{\tabref}[1]{table \ref{#1}}

\begin{document}
\title{Actions in the Eye: Dynamic Gaze Datasets and Learnt Saliency Models for Visual Recognition}

\author{Stefan~Mathe,~\IEEEmembership{Member,~IEEE},
        Cristian~Sminchisescu,~\IEEEmembership{Member,~IEEE}
\IEEEcompsocitemizethanks{\IEEEcompsocthanksitem Stefan Mathe is with the Institute of Mathematics at the Romanian Academy and the Computer Science Department at the University of Toronto. {\it Email: stefan.mathe@imar.ro}. \protect\\
\IEEEcompsocthanksitem Cristian Sminchisescu is with the Department of Mathematics, Faculty of Engineering, Lund University and the Institute of Mathematics at the Romanian Academy. {\it Email: cristian.sminchisescu@math.lth.se (c.a.)}}%
\thanks{}}


\markboth{ } %IEEE Transactions on Pattern Analysis and Machine Intelligence, Submitted, March 2013.}
%Actions in the Eye: Dynamic Gaze Datasets and Learnt Saliency Models for Visual Recognition}%
{Shell \MakeLowercase{\textit{et al.}}: Bare Demo of IEEEtran.cls for Computer Society Journals}

%\IEEEpubid{\makebox[\columnwidth]{\hfill 0000--0000/00/\$00.00~\copyright~2007 IEEE}%


% PRIOR to the title within the \IEEEcompsoctitleabstractindextext IEEEtran
\IEEEcompsoctitleabstractindextext{%
\begin{abstract}
Systems based on bag-of-words models from image features collected at maxima of sparse interest point operators have been used successfully for both computer visual object and action recognition tasks. While the sparse, interest-point based approach to recognition is not inconsistent with visual processing in biological systems that operate in `saccade and fixate' regimes, the methodology and emphasis in the human and the computer vision communities remains sharply distinct. Here, we make three contributions aiming to bridge this gap. First, we complement existing state-of-the art large scale dynamic computer vision annotated datasets like Hollywood-2\cite{MarszalekEtAl2009} and UCF Sports\cite{RodriguezEtAl2008} with human eye movements collected under the ecological constraints of the visual action recognition task. To our knowledge these are the first large human eye tracking datasets to be collected and made publicly available for video, \texttt{vision.imar.ro/eyetracking} (497,107 frames, each viewed by 16 subjects), unique in terms of their \emph{(a) large scale and computer vision relevance, (b) dynamic, video stimuli, (c) task control, as opposed to free-viewing}. Second, we introduce novel \emph{sequential consistency and alignment measures}, which underline the remarkable stability of patterns of visual search among subjects. Third,  we leverage the significant amount of collected data in order to pursue studies and build automatic, end-to-end trainable computer vision systems based on human eye movements.  Our studies not only shed light on the differences between computer vision spatio-temporal interest point image sampling strategies and the human fixations, as well as their impact for visual recognition performance, but also demonstrate that human fixations can be accurately predicted, and when used in an end-to-end \emph{automatic} system, leveraging some of the advanced computer vision practice, can lead to state of the art results.
\end{abstract}
% not want either math or citations to appear in the abstract.

}

\maketitle

% is not selected <OR> if conference mode is selected - because compsoc
% conference papers position the abstract like regular (non-compsoc)
\IEEEdisplaynotcompsoctitleabstractindextext

% \fi
\IEEEpeerreviewmaketitle

\section{Introduction}
% Computer Society journal papers do something a tad strange with the very
% first section heading (almost always called "Introduction"). They place it
% ABOVE the main text! IEEEtran.cls currently does not do this for you.
% However, You can achieve this effect by making LaTeX jump through some
% hoops via something like:
%
%\ifCLASSOPTIONcompsoc
%  \noindent\raisebox{2\baselineskip}[0pt][0pt]%
%  {\parbox{\columnwidth}{\section{Introduction}\label{sec:introduction}%
%  \global\everypar=\everypar}}%
%  \vspace{-1\baselineskip}\vspace{-\parskip}\par
%\else
%  \section{Introduction}\label{sec:introduction}\par
%\fi
%
% Admittedly, this is a hack and may well be fragile, but seems to do the
% trick for me. Note the need to keep any \label that may be used right
% after \section in the above as the hack puts \section within a raised box.


\IEEEPARstart{R}{ecent} progress in computer visual recognition, in particular image classification, object detection and segmentation or action recognition heavily relies on machine learning methods trained on large scale human annotated datasets. The level of annotation varies, spanning a degree of detail from global image or video labels to bounding boxes or precise segmentations of objects\cite{EveringhamEtAl2010}. 
%Such annotations have proven invaluable for performance evaluation and have also supported fundamental progress in designing models and algorithms. 
However, the annotations are often subjectively defined, primarily by the high-level visual recognition tasks generally agreed upon by the computer vision community. While such data has made advances in system design and evaluation possible, it does not necessarily provide insights or constraints into those intermediate levels of computation, or deep structure, that are perceived as ultimately necessary in order to design highly reliable computer vision systems. This is noticeable in the accuracy of state of the art systems trained with such annotations, which still lags significantly behind human performance on similar tasks. Nor does existing data make it immediately possible to exploit insights from an existing working system--the human eye--to potentially derive better features, models or algorithms. 

The divide is well epitomized by the lack of matching large scale datasets that would provide recordings of the workings of the human visual system, in the context of a visual recognition task, at different levels of interpretations including neural systems or eye movements. The human eye movement level, defined by image fixations and saccades, is potentially the less controversial to measure and analyze. It is sufficiently `high-level' or `behavioral' for the computer vision community to rule-out, to some degree at least, open-ended debates on where and what should one record, as could be the case, for instance with neural systems in different brain areas\cite{Rolls2011}. Besides, our goals in this context are pragmatic: fixations provide a sufficiently high-level signal that can be precisely registered with the image stimuli, for testing hypotheses and for training visual feature extractors and recognition models quantitatively. It can potentially foster links with the human vision community, in particular researchers developing biologically plausible models of visual attention, who would be able to test and quantitatively analyze their models on shared large scale datasets\cite{JuddEtAl2009,LarochelleHinton2010,Rolls2011}.

%Apart from linking the human and computer vision communities, human eye movement annotations offer pragmatic potential: fixations provide a sufficiently high-level signal that can be precisely registered with the image stimuli, for testing hypotheses and for training visual feature extractors and recognition models quantitatively. 
Some of the most successful approaches to action recognition employ bag-of words representations based on descriptors computed at spatial-temporal video locations, obtained at the maxima of an interest point operator biased to fire over non-trivial local structure (space-time `corners' or spatial-temporal interest points\cite{Laptev2005}). More sophisticated image representations based on objects and their relations, as well as multiple kernels have been employed with a degree of success\cite{DongEtAl2009}, although it appears still difficult to detect a large variety of useful objects reliably in challenging video footage. Although human pose estimation could greatly disambiguate the interations between actors and manipulated objects, it is a difficult problem even in a controlled setting due to the large number of local minima in the search space\cite{SminchisescuTriggs2005}. The dominant role of sparse spatial-temporal interest point operators as front end in computer vision systems raises the question whether computational insights from a working system like the human visual system can be used to improve performance. The sparse approach to computer visual recognition is not inconsistent to the one of biological systems, but the degree of repeatability and the effect of using human fixations with computer vision algorithms in the context of action recognition have not been yet explored. In this paper we make the following contributions:

\begin{enumerate}
\item We undertake a significant effort of recording and analyzing human eye movements in the context of \emph{dynamic visual action recognition tasks} for two existing computer vision datasets, Hollywood 2\cite{MarszalekEtAl2009} and UCF-Sports\cite{RodriguezEtAl2008}. This dynamic data is made publicly available to the research community at \texttt{vision.imar.ro/eyetracking}.
%\cite{dataset_url}.
\item We introduce novel consistency models and algorithms, as well as relevance evaluation measures adapted for video. Our findings (see \S\ref{s:consistency}) suggest a remarkable degree of sequential consistency--both spatial and sequential--in the fixation patterns of human subjects but also underline a less extensive influence of task on dynamic fixations than previously believed, at least within the class of the datasets and actions we studied.
\item By using our large scale training set of human fixations and by leveraging static and dynamic image features based on color, texture, edge distributions (HoG) or motion boundary histograms (MBH), we introduce novel saliency detectors and show that these can be trained effectively to predict human fixations  as measured under both average precision (AP), and as Kullblack-Leibler spatial comparison measures. See \S\ref{s:saliency_predict} and \tabref{t:judd_measures} for results.
\item We show that training an end-to-end automatic visual action recognition system based on our learned saliency interest operator (point 3), and using advanced computer vision descriptors and fusion methods, leads to state of the art results in the Hollywood-2 and UCF-Sports action datasets. This is, we argue, one of the first demonstrations of a successful symbiosis of computer vision and human vision technology, within the context of a very challenging dynamic visual recognition task. It shows the potential of interest point operators learnt from human fixations for computer vision. Models and experiments appear in \S\ref{s:end_to_end_recognition}, results in \tabref{t:classification}. This paper extends our prior work in \cite{MatheSmi12}.
\end{enumerate}

\begin{figure}[ht]
\begin{center}
\begin{tabular}{@{\vspace{-1mm}}c@{}c@{}c@{}c}
\scalebox{0.126}{\includegraphics[viewport=0.3cm 0.2cm 15.5cm 6.0cm]{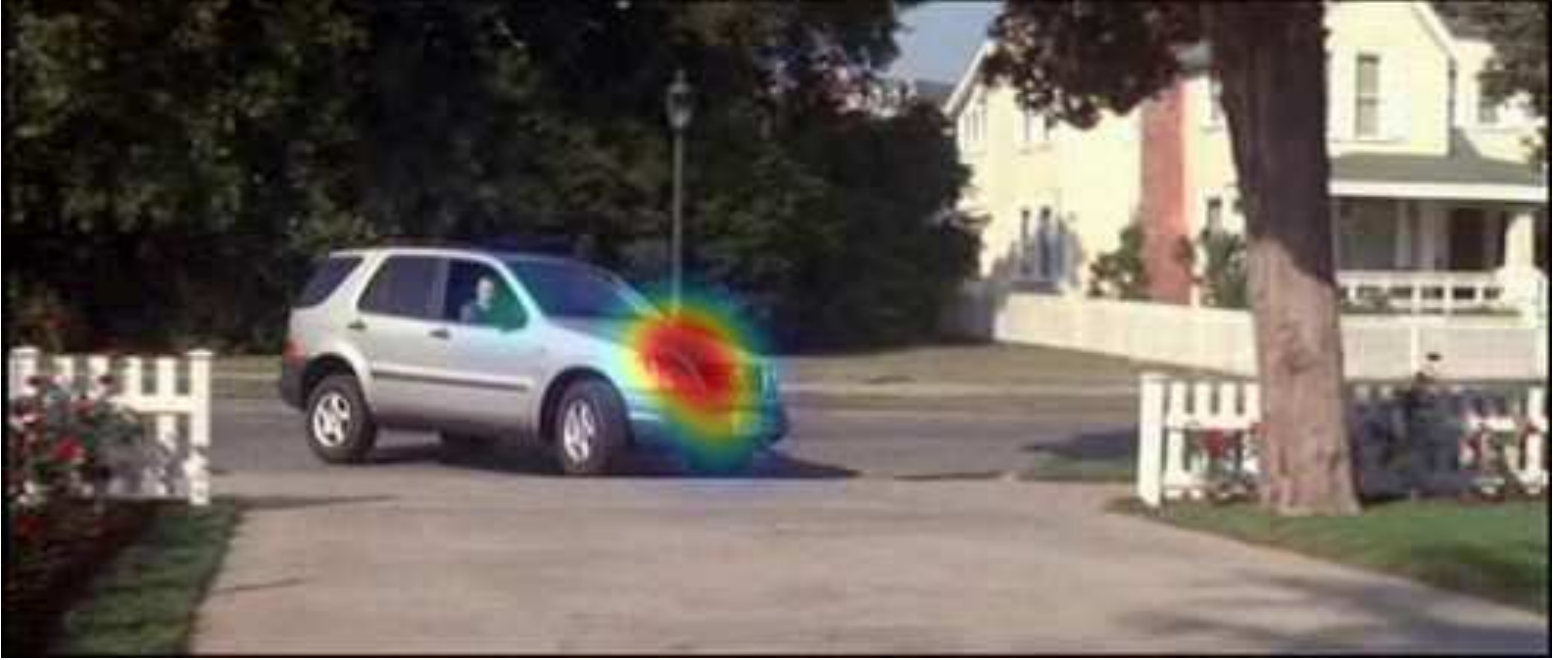}}
&
\hspace{0.4mm}
\scalebox{0.126}{\includegraphics[viewport=0.3cm 0.2cm 15.5cm 6.0cm]{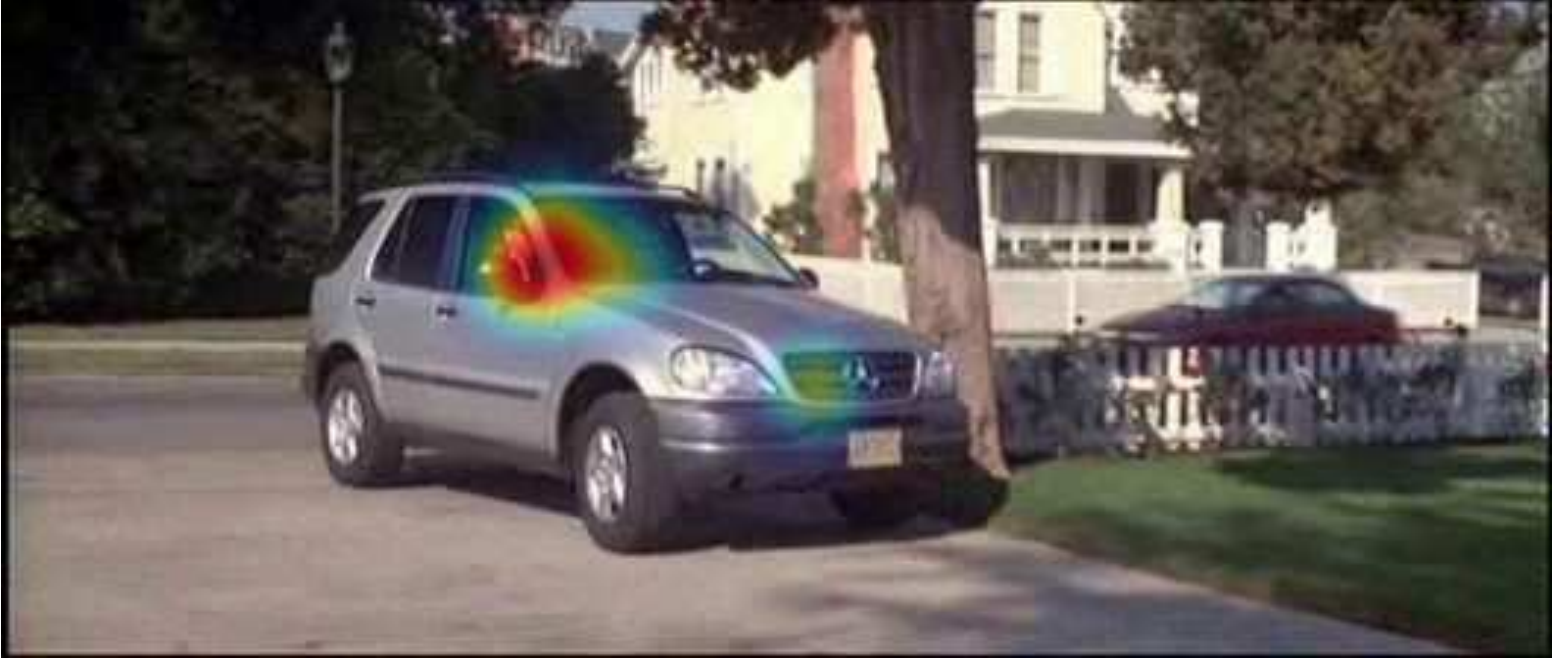}}
&
\hspace{0.4mm}
\scalebox{0.126}{\includegraphics[viewport=0.3cm 0.2cm 15.5cm 6.0cm]{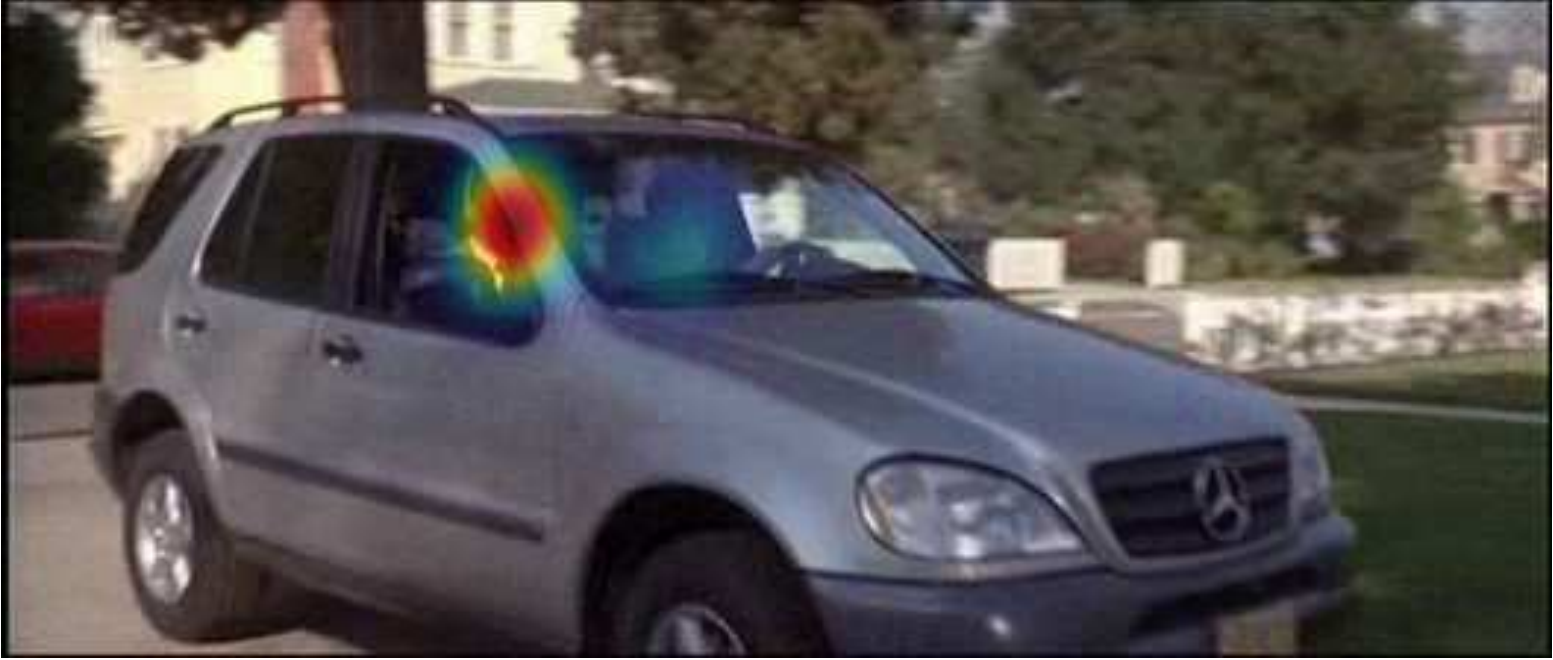}}
&
\hspace{0.4mm}
\scalebox{0.126}{\includegraphics[viewport=0.3cm 0.2cm 15.5cm 6.0cm]{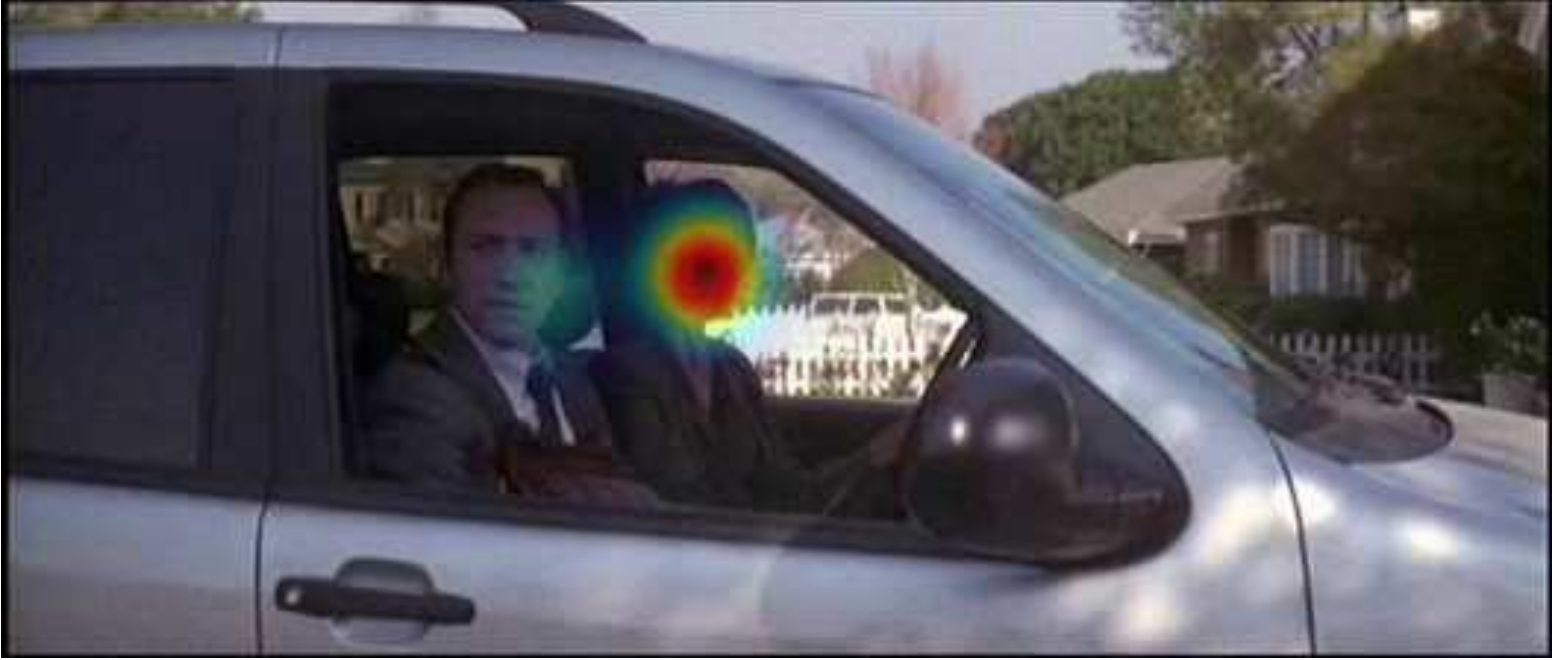}}
\\
\scalebox{0.8}{frame 25}
&
\scalebox{0.8}{frame 50}
&
\scalebox{0.8}{frame 90}
&
\scalebox{0.8}{frame 150}
\\
\scalebox{0.126}{\includegraphics[viewport=0.3cm 0.2cm 15.5cm 10.0cm]{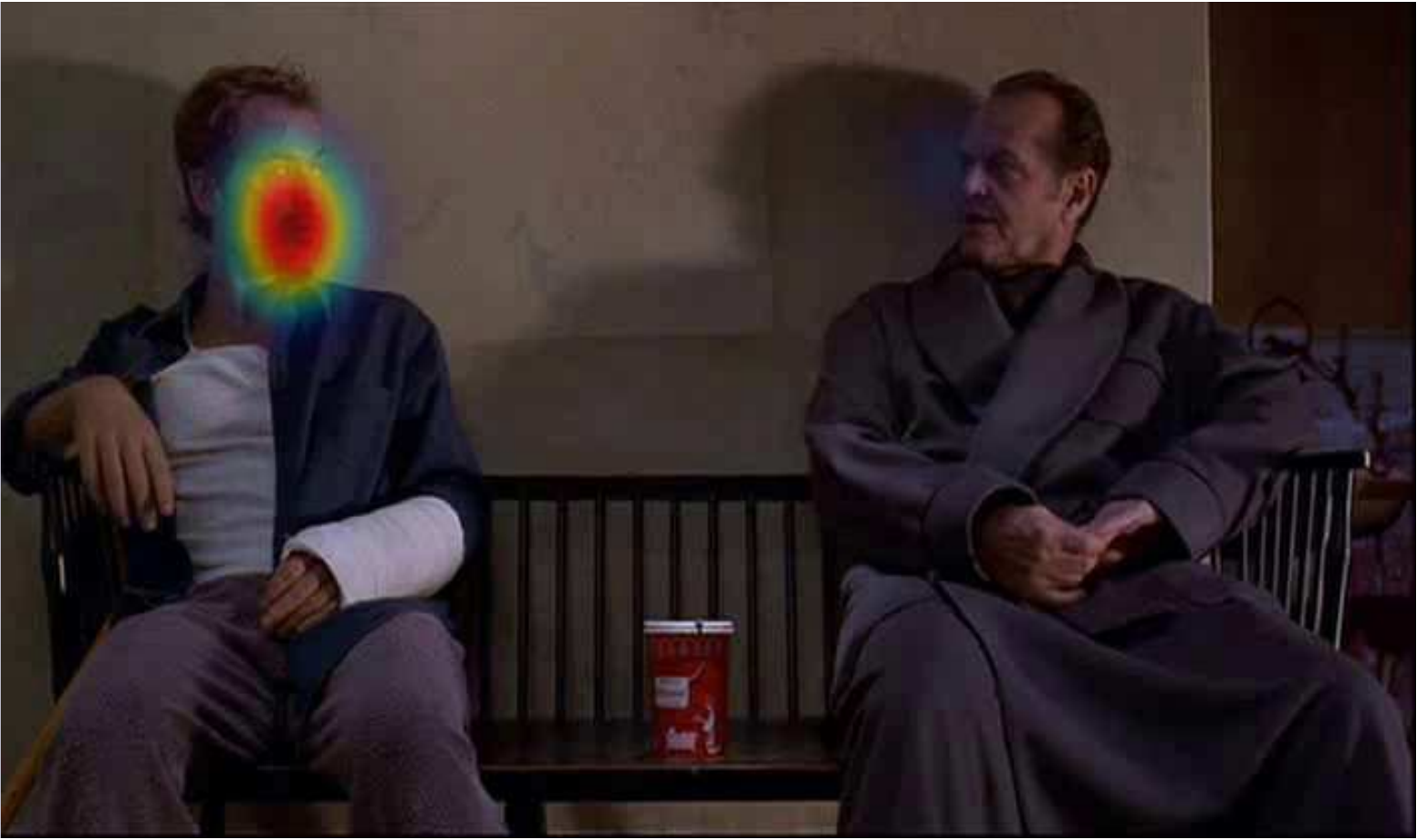}}
&
\hspace{0.4mm}
\scalebox{0.126}{\includegraphics[viewport=0.3cm 0.2cm 15.5cm 10.0cm]{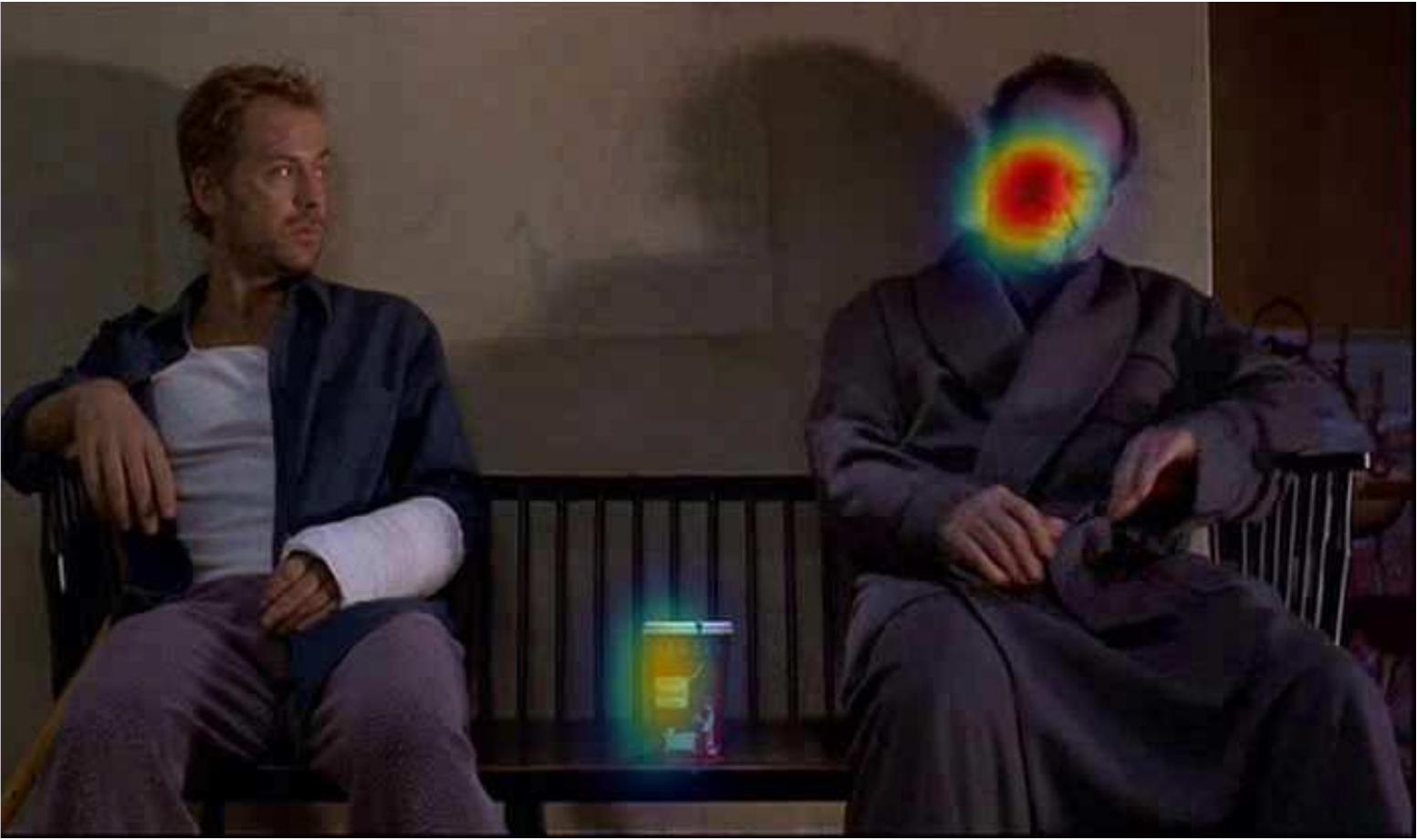}}
&
\hspace{0.4mm}
\scalebox{0.126}{\includegraphics[viewport=0.3cm 0.2cm 15.5cm 10.0cm]{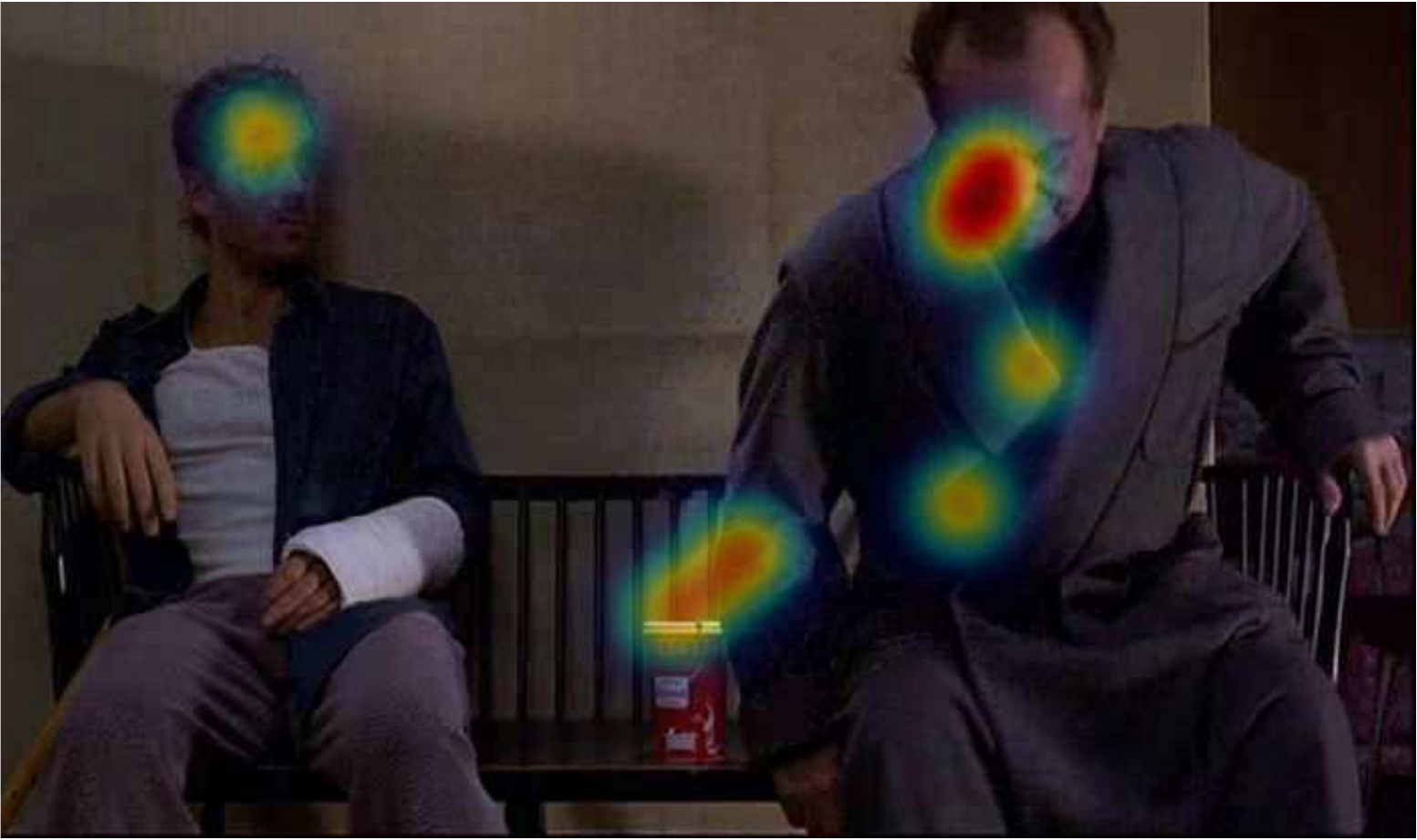}}
&
\hspace{0.4mm}
\scalebox{0.126}{\includegraphics[viewport=0.3cm 0.2cm 15.5cm 10.0cm]{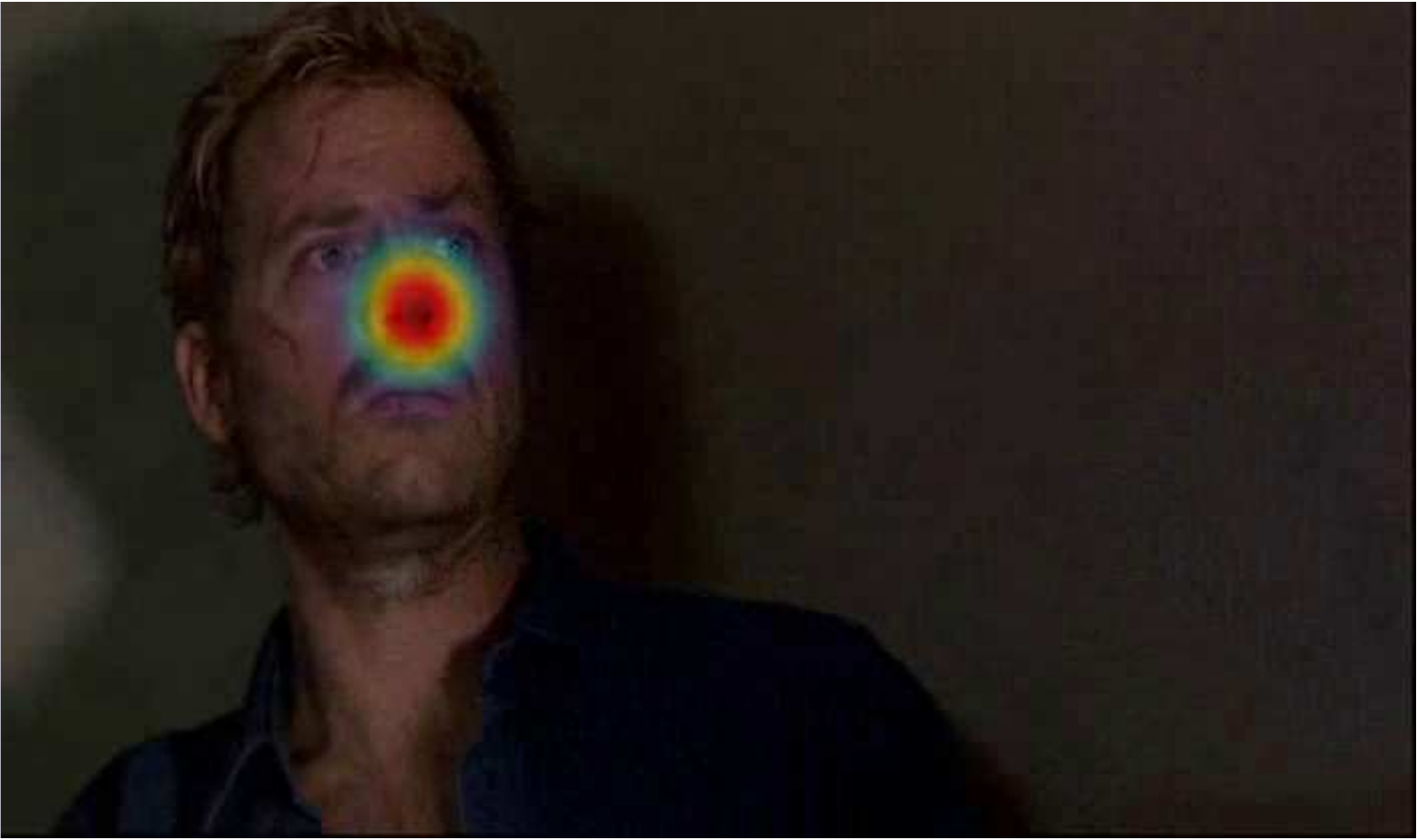}}
\\
\scalebox{0.8}{frame 15}
&
\scalebox{0.8}{frame 40}
&
\scalebox{0.8}{frame 60}
&
\scalebox{0.8}{frame 90}
\\
\scalebox{0.126}{\includegraphics[viewport=0.3cm 0.2cm 15.5cm 9.0cm]{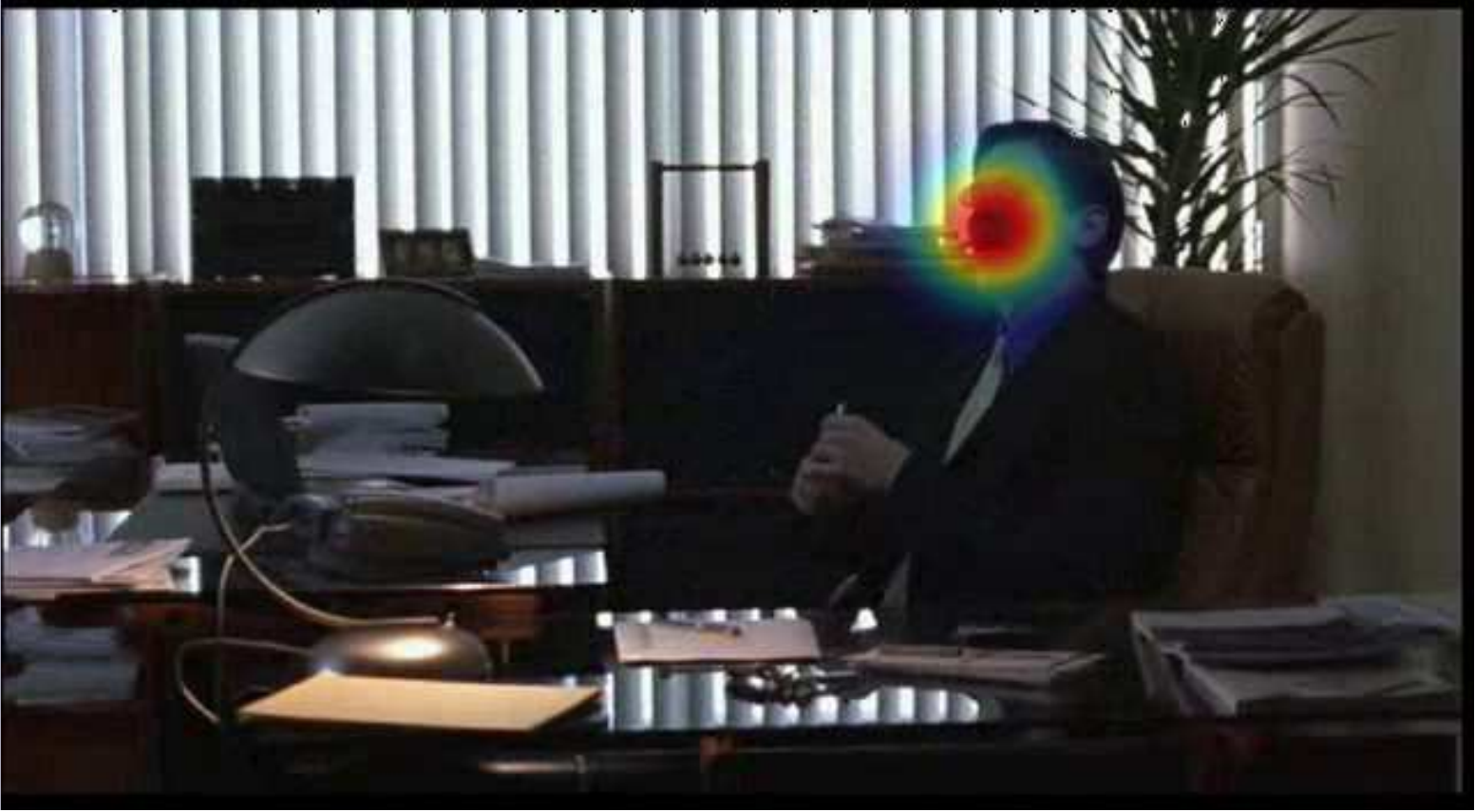}}
&
\hspace{0.4mm}
\scalebox{0.126}{\includegraphics[viewport=0.3cm 0.2cm 15.5cm 9.0cm]{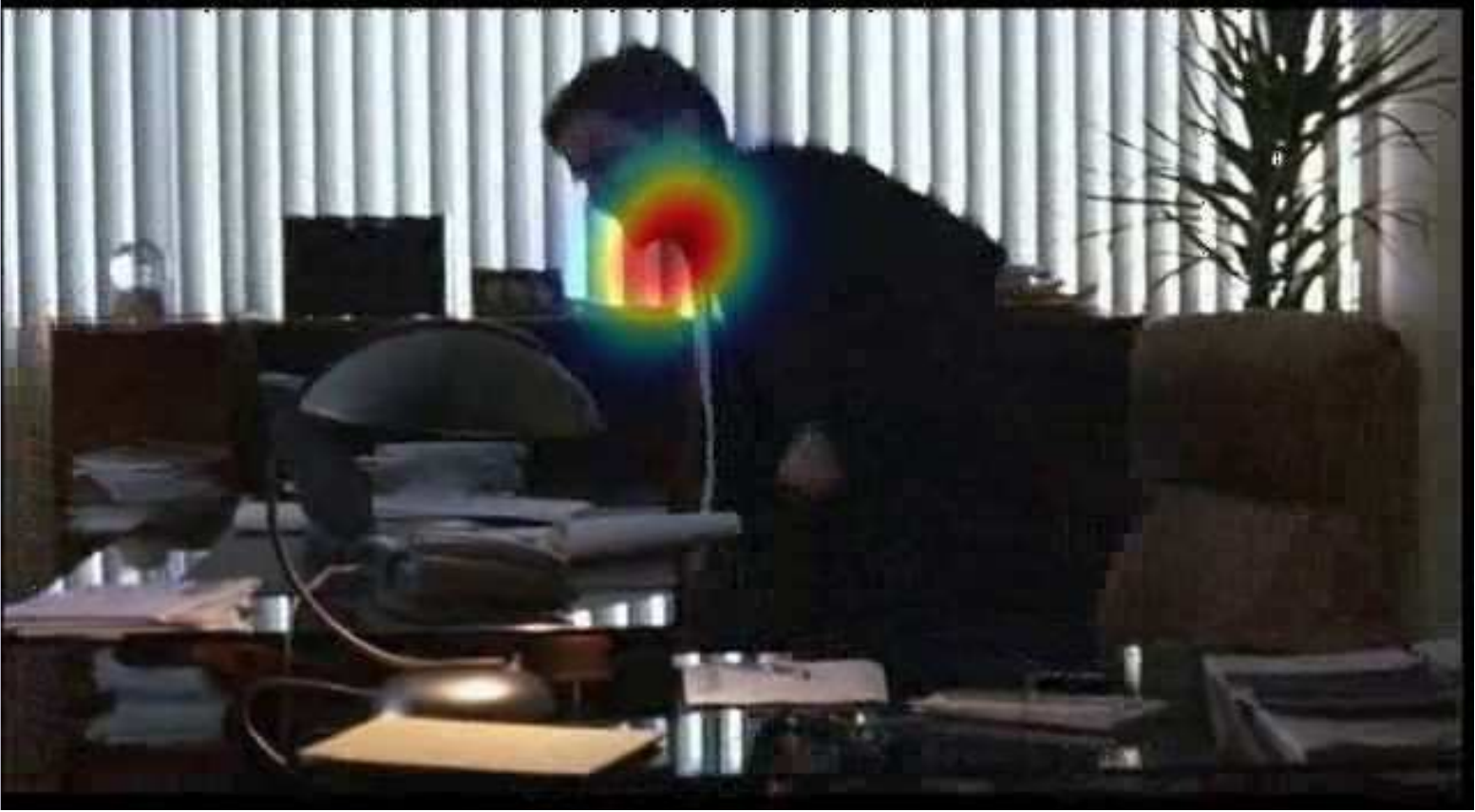}}
&
\hspace{0.4mm}
\scalebox{0.126}{\includegraphics[viewport=0.3cm 0.2cm 15.5cm 9.0cm]{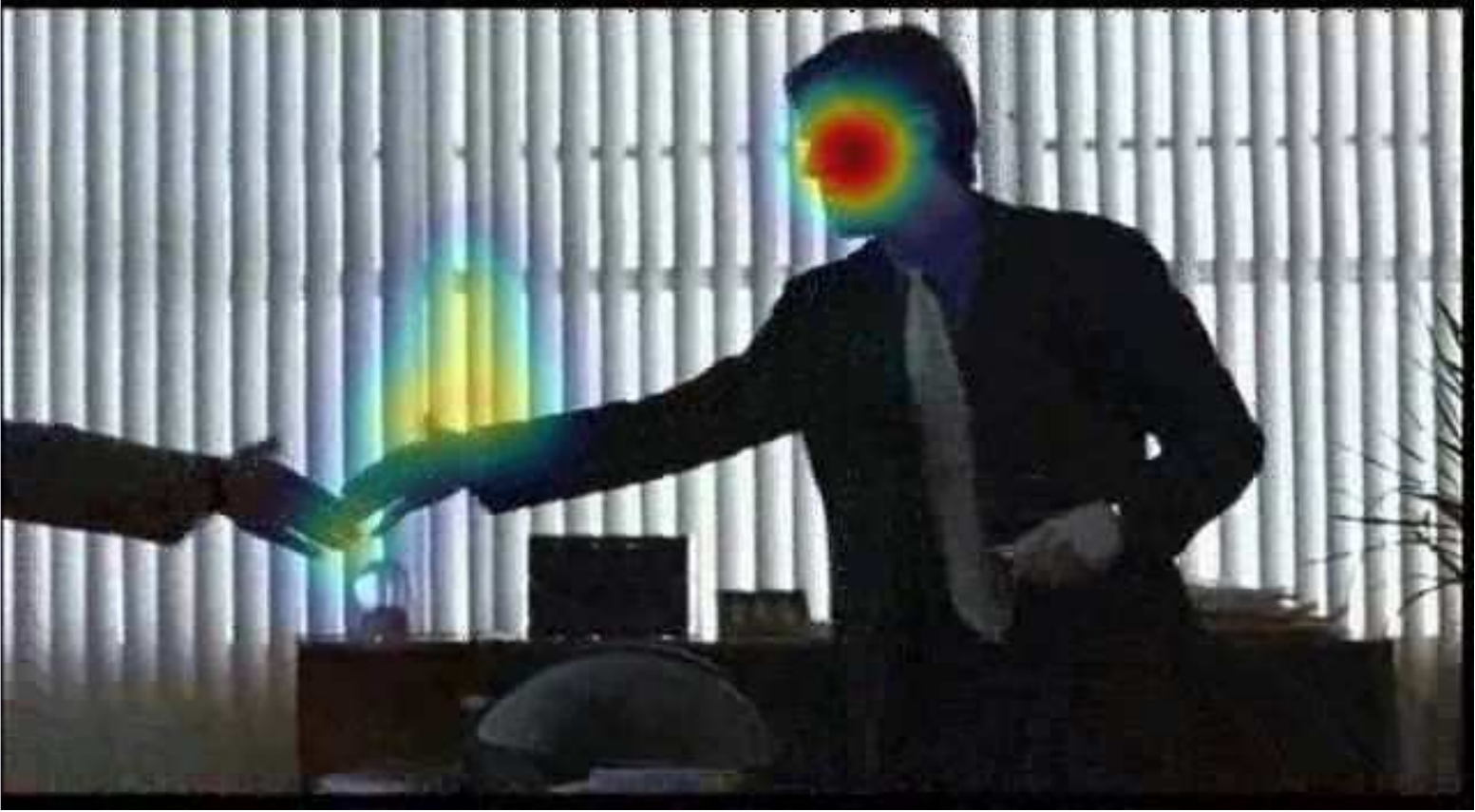}}
&
\hspace{0.4mm}
\scalebox{0.126}{\includegraphics[viewport=0.3cm 0.2cm 15.5cm 9.0cm]{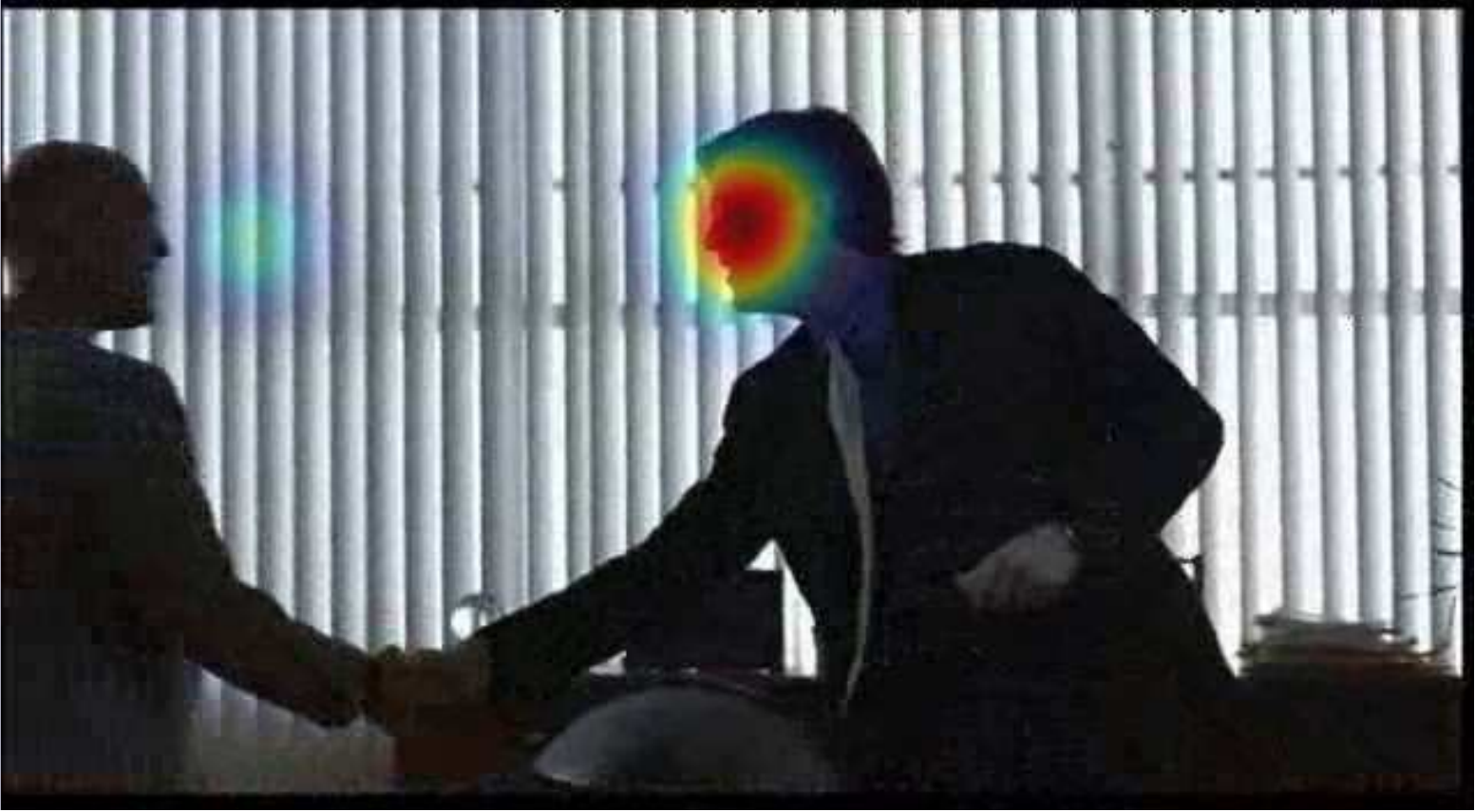}}
\\
\scalebox{0.8}{frame 10}
&
\scalebox{0.8}{frame 35}
&
\scalebox{0.8}{frame 70}
&
\scalebox{0.8}{frame 85}
\\
\scalebox{0.1}{\includegraphics[viewport=0.3cm 0.2cm 20.0cm 14.0cm]{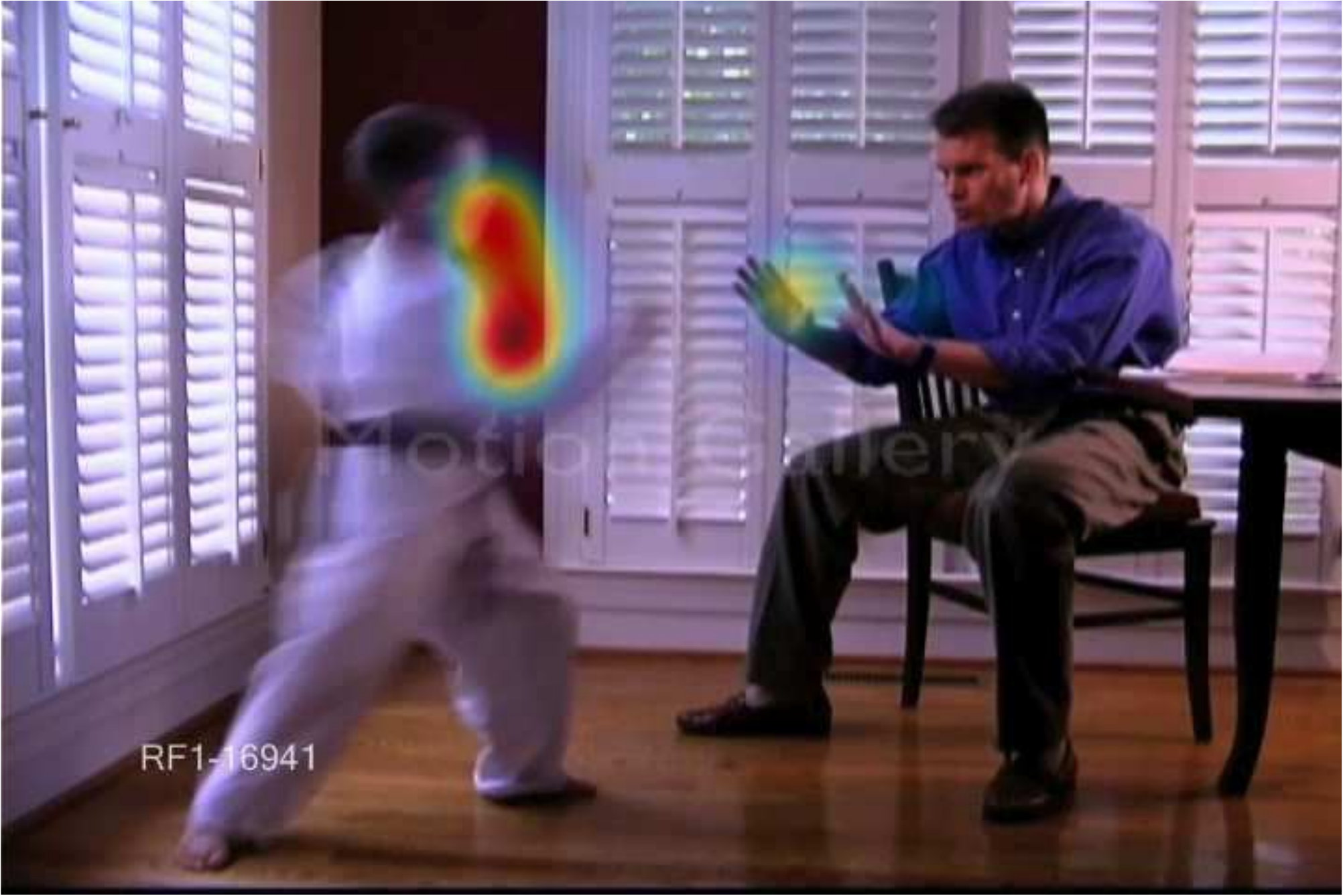}}
&
\hspace{0.4mm}
\scalebox{0.1}{\includegraphics[viewport=0.3cm 0.2cm 20.0cm 14.0cm]{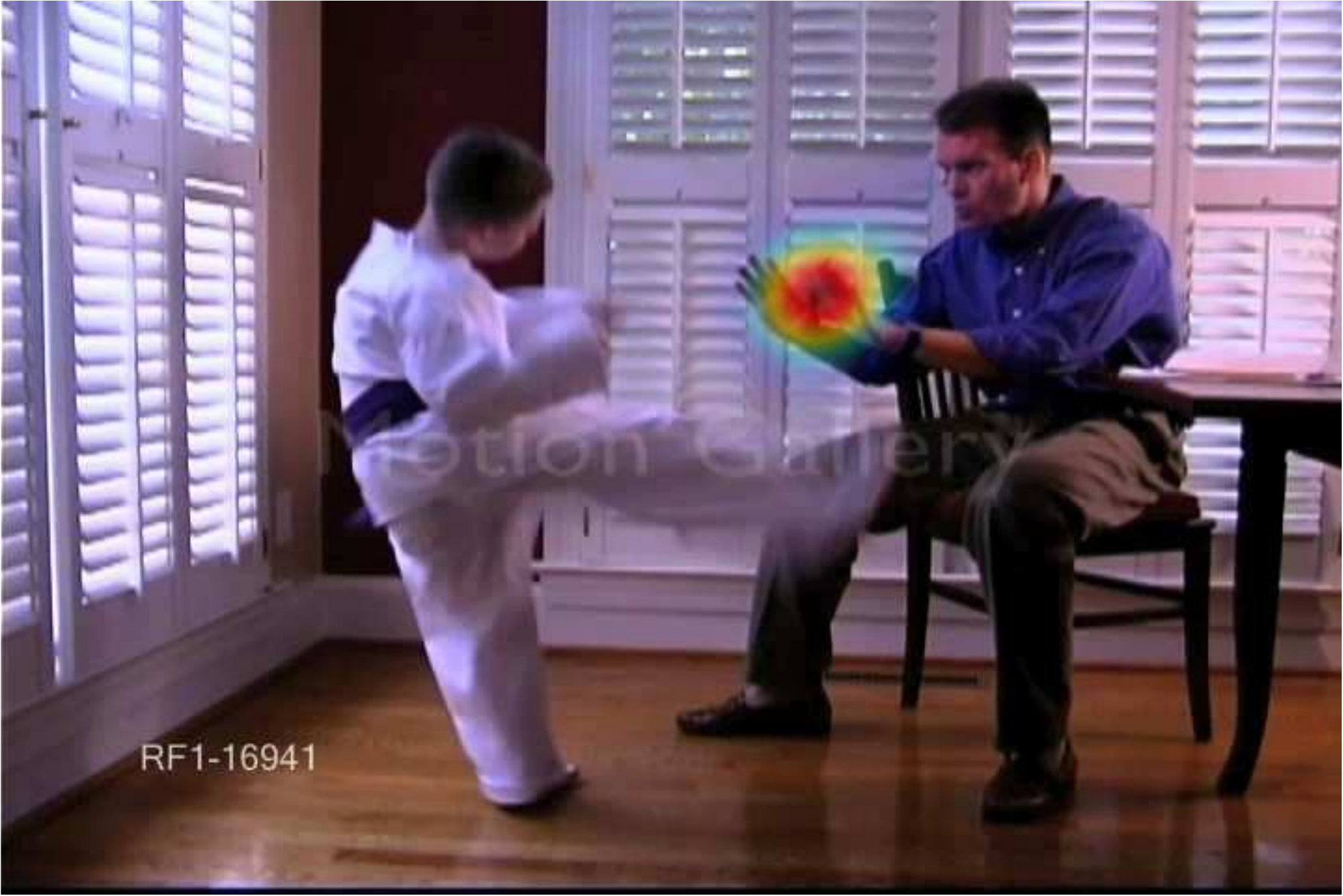}}
&
\hspace{0.4mm}
\scalebox{0.1}{\includegraphics[viewport=0.3cm 0.2cm 20.0cm 14.0cm]{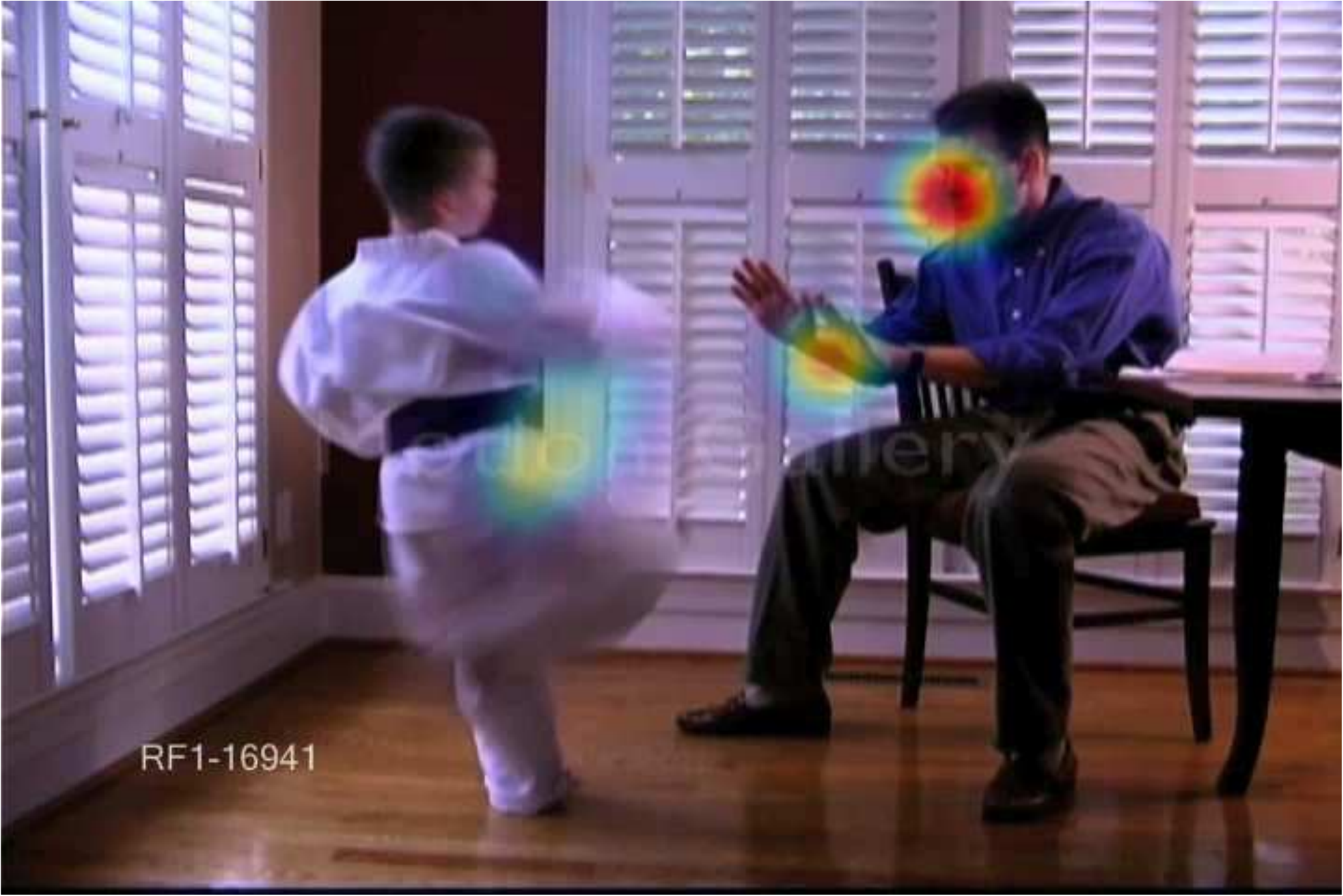}}
&
\hspace{0.4mm}
\scalebox{0.1}{\includegraphics[viewport=0.3cm 0.2cm 20.0cm 14.0cm]{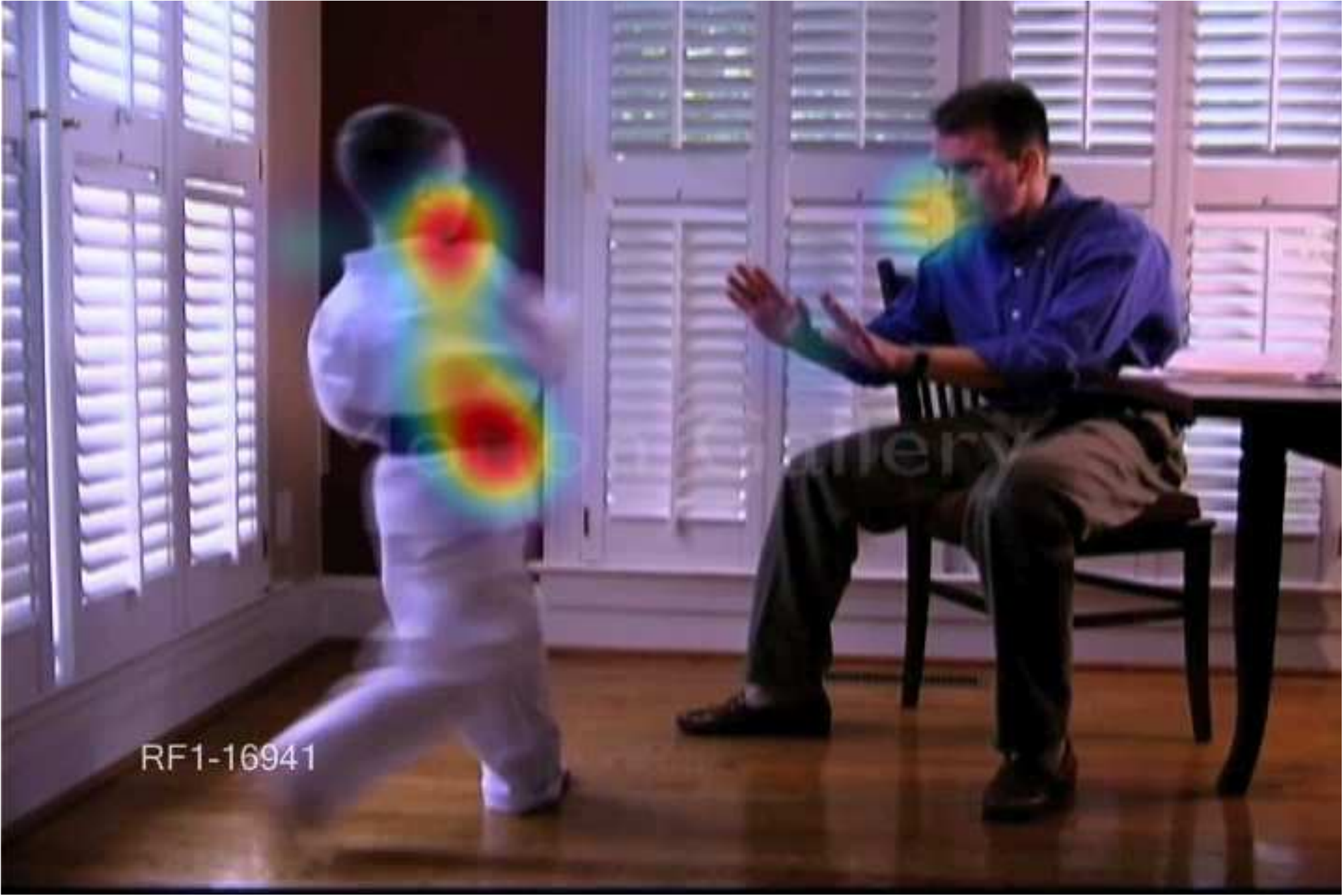}}
\\
\scalebox{0.8}{frame 3}
&
\scalebox{0.8}{frame 10}
&
\scalebox{0.8}{frame 18}
&
\scalebox{0.8}{frame 22}
\\
\scalebox{0.105}{\includegraphics[viewport=0.3cm 0.2cm 19.0cm 16.0cm]{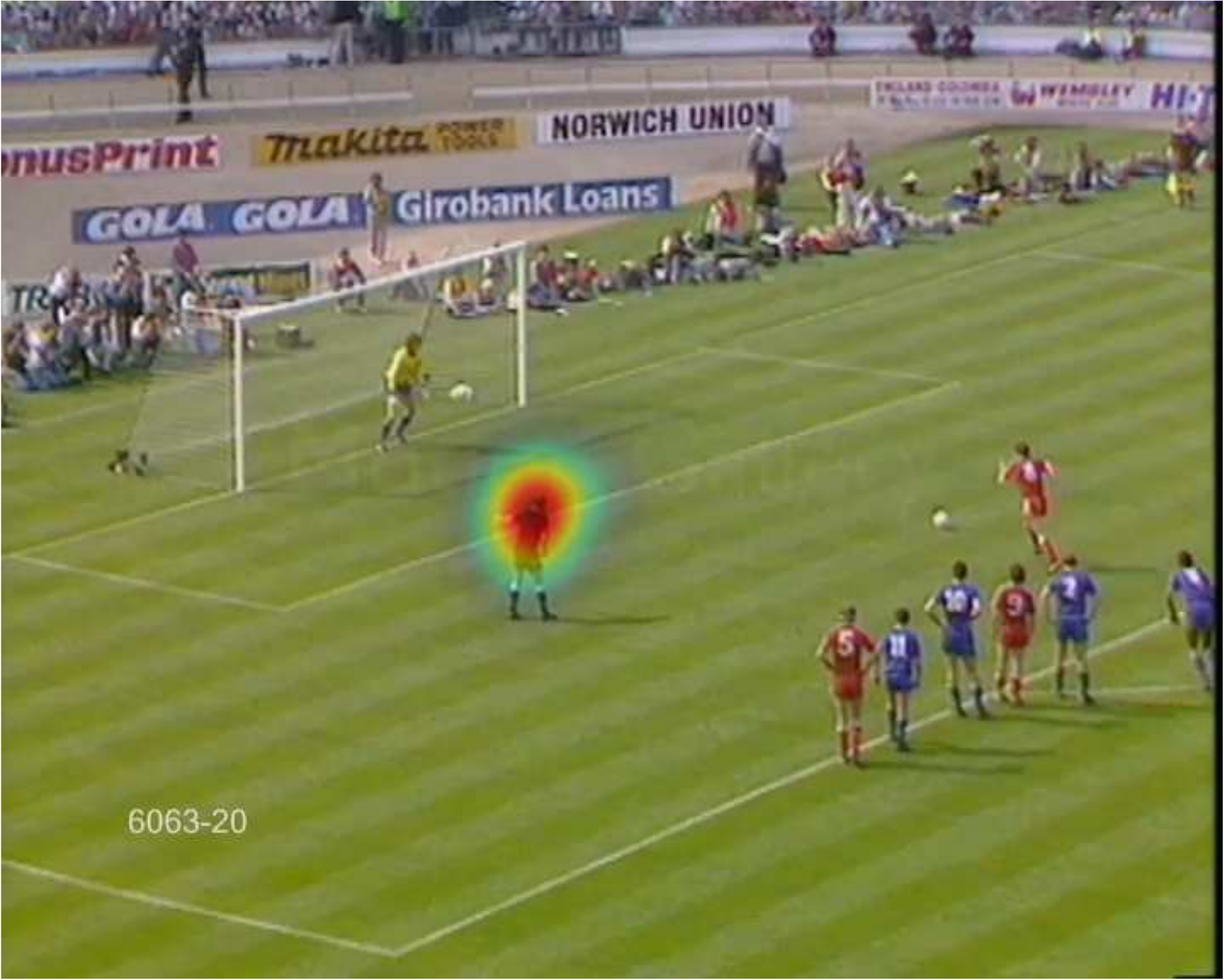}}
&
\hspace{0.4mm}
\scalebox{0.105}{\includegraphics[viewport=0.3cm 0.2cm 19.0cm 16.0cm]{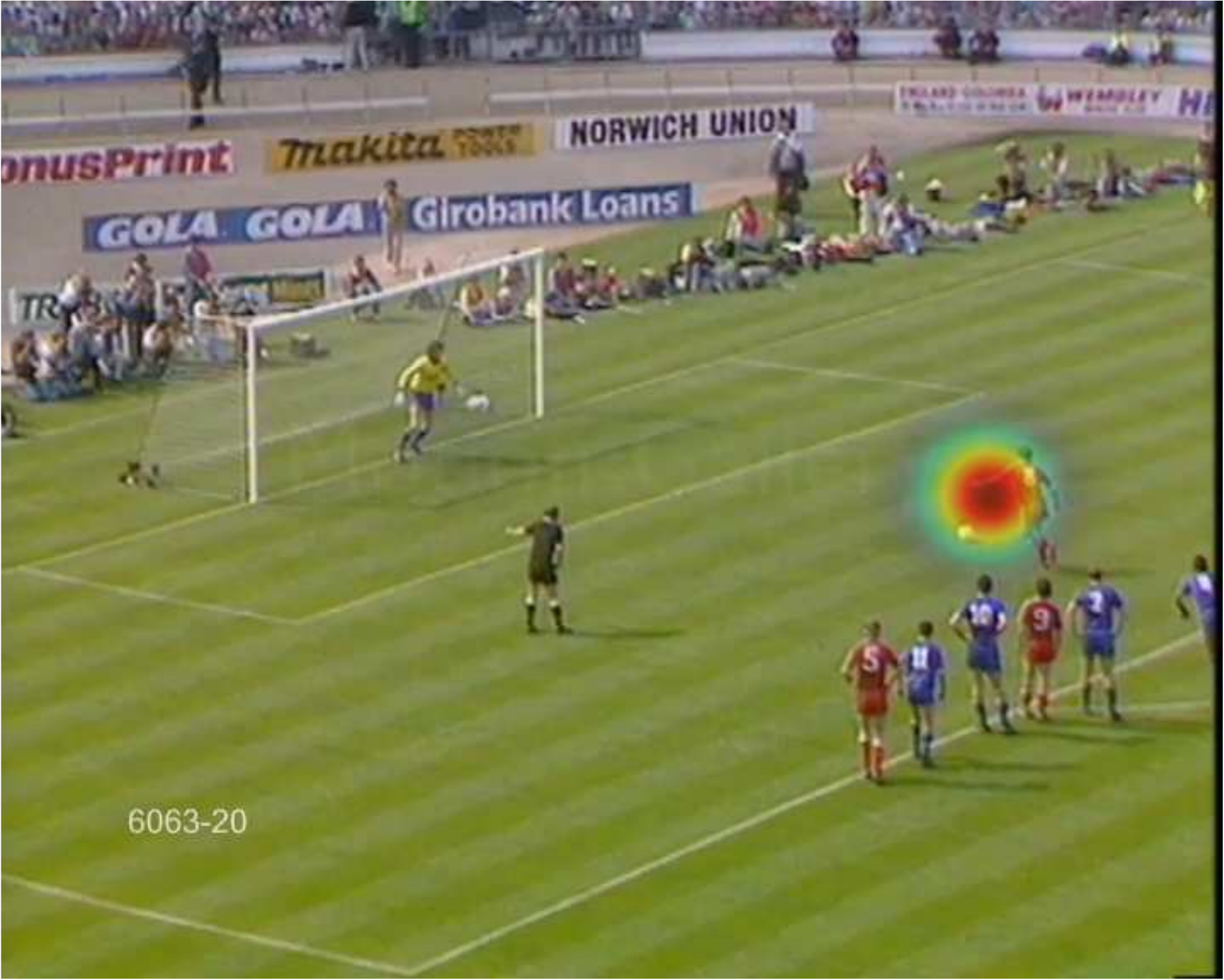}}
&
\hspace{0.4mm}
\scalebox{0.105}{\includegraphics[viewport=0.3cm 0.2cm 19.0cm 16.0cm]{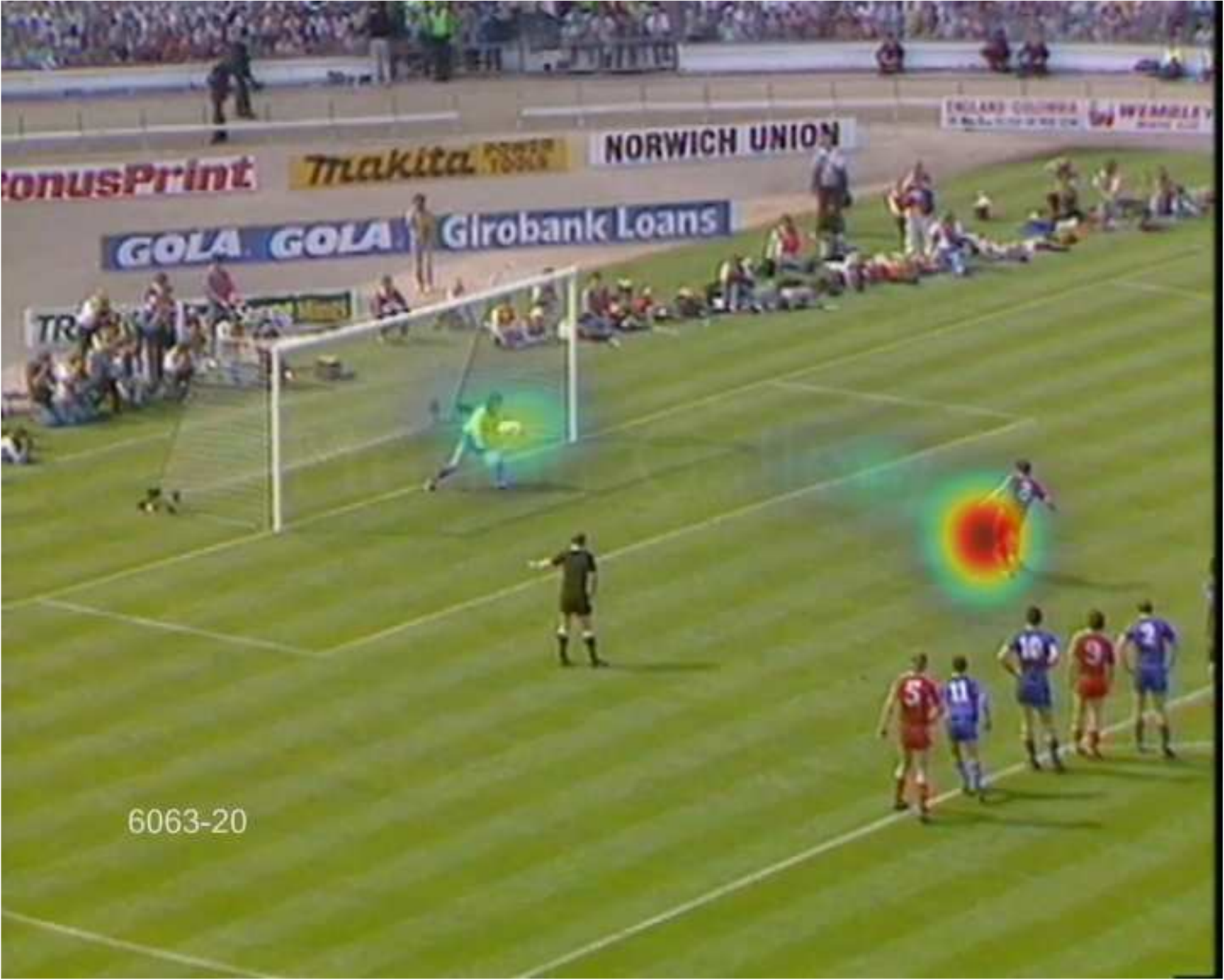}}
&
\hspace{0.4mm}
\scalebox{0.105}{\includegraphics[viewport=0.3cm 0.2cm 19.0cm 16.0cm]{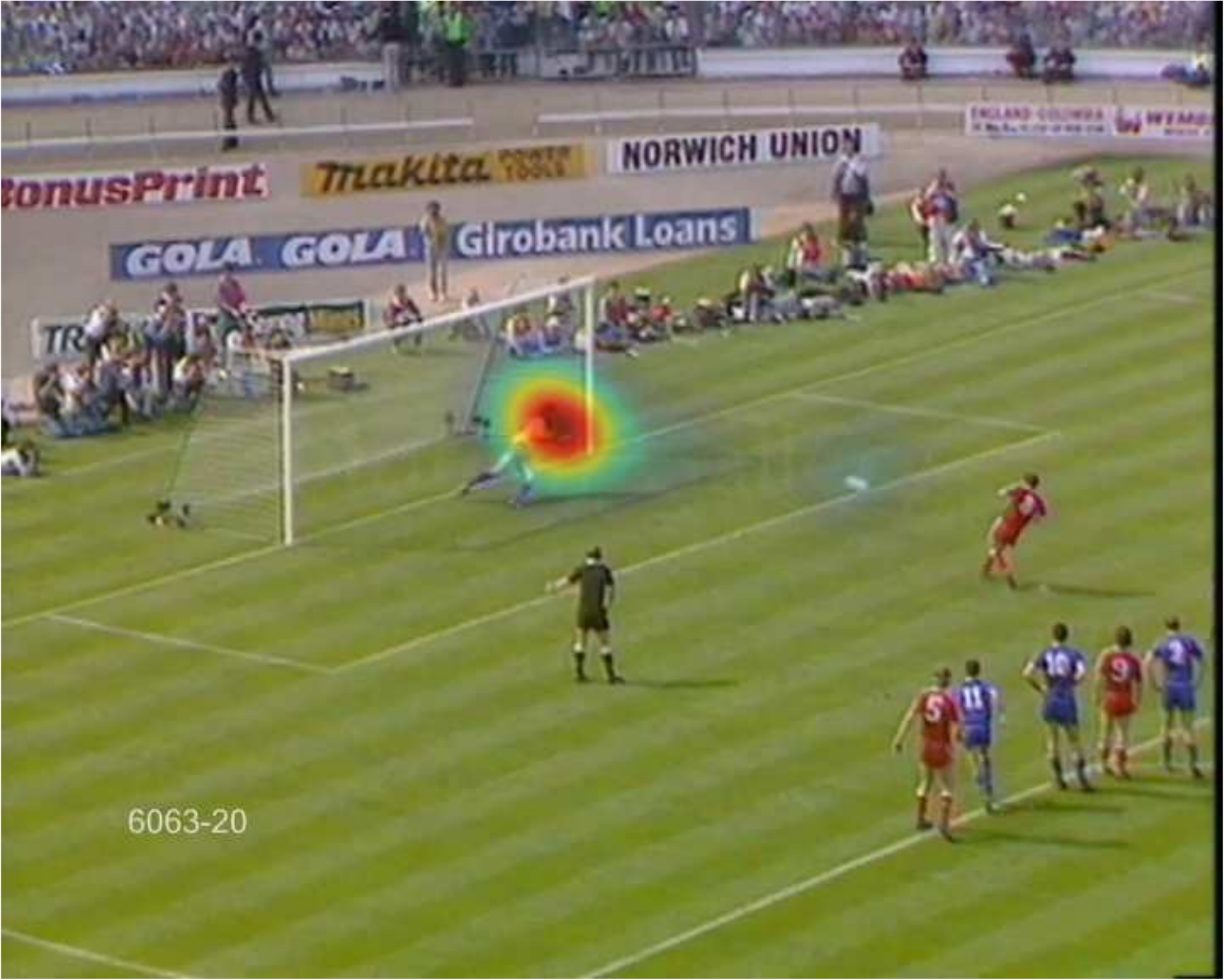}}
\\
\scalebox{0.8}{frame 3}
&
\scalebox{0.8}{frame 7}
&
\scalebox{0.8}{frame 15}
&
\scalebox{0.8}{frame 19}
\\
\scalebox{0.1}{\includegraphics[viewport=0.3cm 0.2cm 20.0cm 14.0cm]{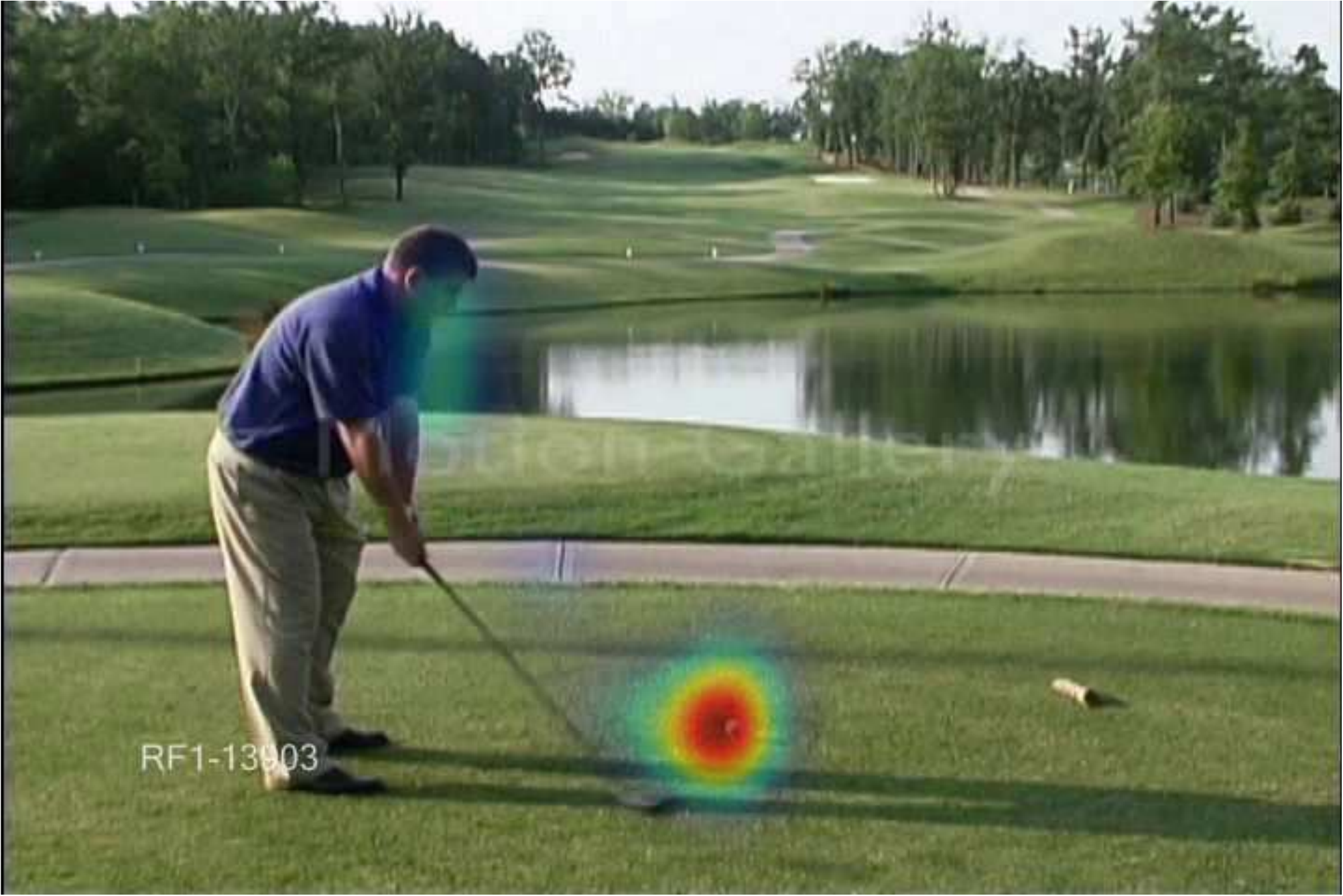}}
&
\hspace{0.4mm}
\scalebox{0.1}{\includegraphics[viewport=0.3cm 0.2cm 20.0cm 14.0cm]{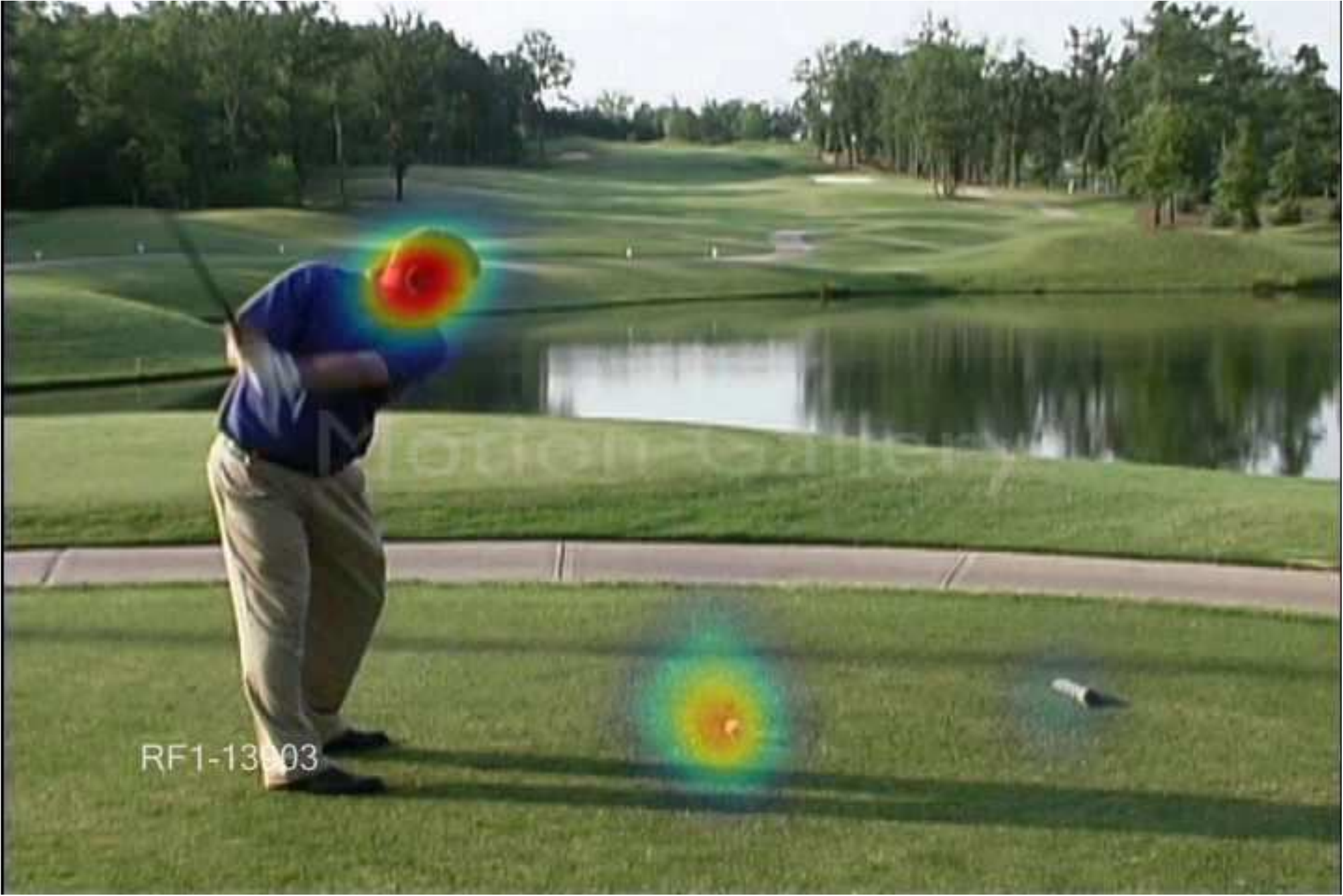}}
&
\hspace{0.4mm}
\scalebox{0.1}{\includegraphics[viewport=0.3cm 0.2cm 20.0cm 14.0cm]{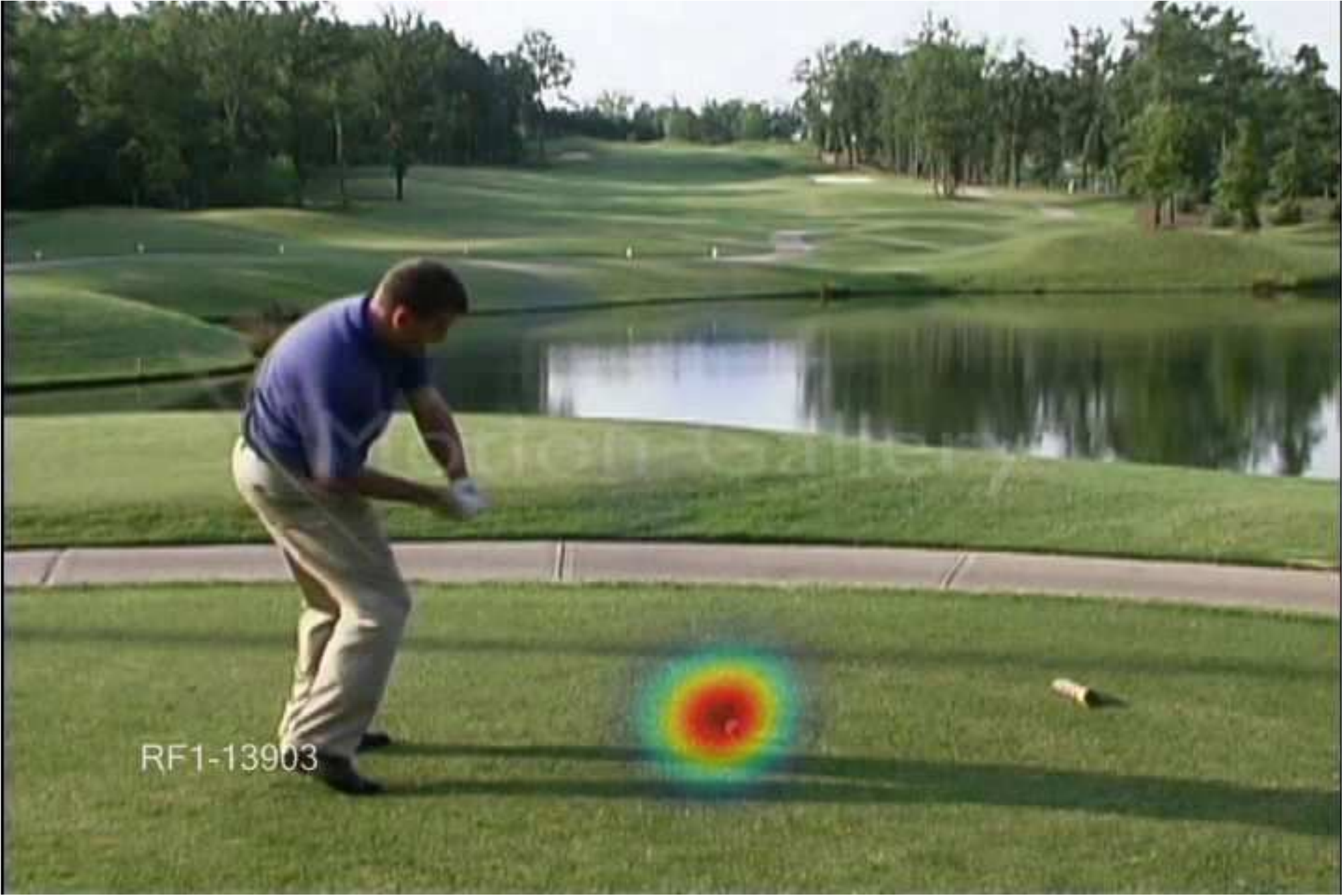}}
&
\hspace{0.4mm}
\scalebox{0.1}{\includegraphics[viewport=0.3cm 0.2cm 20.0cm 14.0cm]{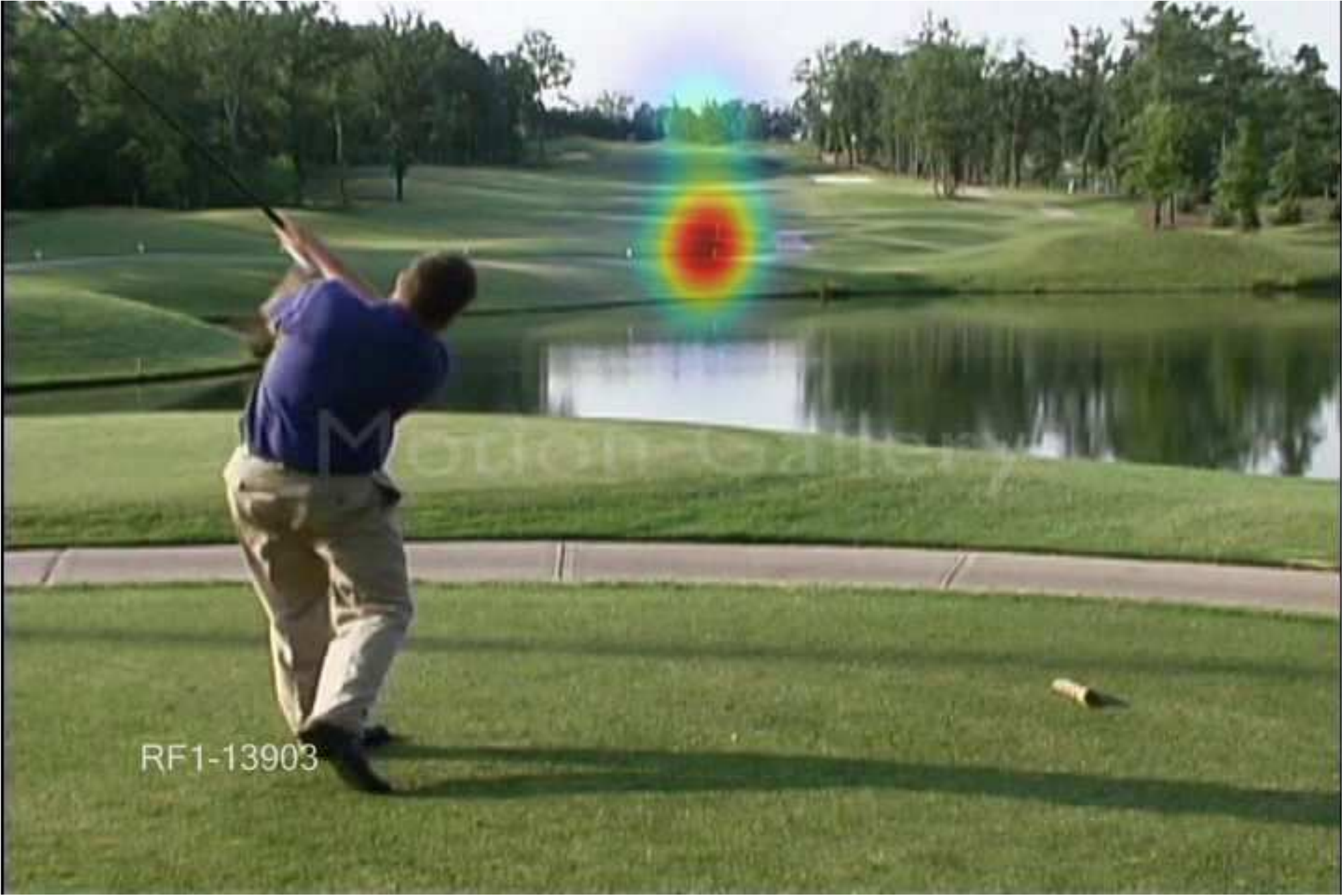}}
\\
\scalebox{0.8}{frame 4}
&
\scalebox{0.8}{frame 12}
&
\scalebox{0.8}{frame 25}
&
\scalebox{0.8}{frame 31}
\end{tabular}
\end{center}
\caption{Heat maps generated from the fixations of 16 human subjects viewing 6 videos selected from the Hollywood-2 and UCF Sports datasets. Fixated locations are generally tightly clustered. This suggests a significant degree of consistency among human subjects in terms of the spatial distribution of their visual attention. See fig. \ref{fig:static_consistency} for quantitative studies.}\label{fig:motivation}
\end{figure}

The paper is organized as follows. In \S\ref{s:relwork} we briefly review existing studies on human visual attention and saliency, as well as the state-of-the art computational models for automatic action recognition from videos. Our dataset and data collection methodology are introduced in \S\ref{s:dataset}. In \S\ref{s:consistency} we analyze inter-subject agreement and introduce two novel metrics for measuring spatial and sequential visual consistency in the video domain.
In addition to showing remarkable visual consistency, human subjects also tend to fixate image structures that are semantically meaningful, which we illustrate in \S\ref{s:visual_vocabularies}. This naturally suggests that human fixations could provide useful information to support automatic action recognition systems. In section \S\ref{s:exp_setup} we introduce our action recognition pipeline which we shall use through the remainder of the paper. Section \S\ref{s:fix_studies} explores the action recognition potential of several interest point operators derived from ground truth human fixations and visual saliency maps. In \S\ref{s:saliency_predict} we turn our attention to the problem of human visual saliency prediction, and introduce a novel spatio-temporal human fixation detector trained using our human gaze dataset. Section \S\ref{s:end_to_end_recognition} illustrates how predicted saliency maps can be integrated into a modern state-of-the-art end-to-end action recognition system. We draw our final conclusions in \S\ref{s:conclusions}.

\section{Related Work}\label{s:relwork}

The study of gaze patterns in humans has long received significant interest by the human vision community \cite{Yarbus1967}. Whether visual attention is driven by purely bottom-up cues \cite{IttiEtAl1998} or a combination of top-down and bottom-up influences \cite{IttiReesTsotsos2005} is still open to debate (see \cite{LandTatler2009,IttiReesTsotsos2005} for comprehensive reviews). One of the oldest theories of visual attention has been that bottom-up features guide vision towards locations of high saliency \cite{LandTatler2009,IttiReesTsotsos2005}. Early computational models of attention \cite{IttiEtAl1998,IttiKoch2000} assume that human fixations are driven by maxima inside a topographicals map that encodes the saliency of each point in the visual field.

Models of saliency can be either pre-specified or learned from eye tracking data. In the former category falls the basic saliency model \cite{IttiKoch2000} that combines information from multiple channels into a single saliency map. Information maximization \cite{BruceTsotsos2005} provides an alternative criterion for building saliency maps. These can be learned from low-level features \cite{KienzleEtAl2006} or from a combination of low, mid and high-level ones \cite{JuddEtAl2009,TorralbaEtAl2006,JuddEtAl2011}. Saliency maps have been used for scene classification \cite{BorjiItti2011}, object localization and recognition \cite{ElazaryItti2010,EhingerEtAl2009,SerreEtAl2007,WaltherEtAl2005,HanVasconcelos2010} and action recognition \cite{SerreEtAlICCV2007,KienzleEtAl2007}. Comparatively little attention has been devoted to computational models of saliency maps for the dynamic domain. Bottom-up saliency models for static images have been extended to videos by incorporating motion and flicker channels \cite{MaratEtAl2007,LeMeurEtAl2007}. All these models are pre-specified. One exception is the work of Kienzle et al. \cite{KienzleEtAl2007}, who train an interest point detector using fixation data collected from human subjects in a free viewing (rather than specific) task.

Datasets containing human gaze pattern annotations of images have emerged from studies carried out by the human vision community, some of which are publicly available \cite{IttiKoch2000,JuddEtAl2009,EhingerEtAl2009,JuddEtAl2011,FathiLiRehg2012} and some that are not \cite{KienzleEtAl2007} (see \cite{WinklerSubramanian2013} for an overview). Most of these datasets have been designed for small  quantitative studies, consisting of at most a few hundred images or videos, usually recorded under free-viewing, in sharp contrast with the data we provide, which is large scale, dynamic, and task controlled. These studies \cite{IttiKoch2000,LarochelleHinton2010,KienzleEtAl2007,EhingerEtAl2009,JuddEtAl2011,FeiFeiEtAl2007,JhuangEtAl2007,IttiReesTsotsos2005} could however benefit from larger scale natural datasets, and from studies that emphasize the task, as we pursue.

The problem of visual attention and the prediction of visual saliency have long been of interest in the human vision community \cite{IttiReesTsotsos2005,IttiKoch2000,LandTatler2009}. Recently there was a growing trend of training visual saliency models based on human fixations, mostly in static images (with the notable exception of \cite{KienzleEtAl2007}), and under subject free-viewing conditions \cite{JuddEtAl2009,TorralbaEtAl2006}. While visual saliency models can be evaluated in isolation under a variety of measures against human fixations, for computer vision, their ultimate test remains the demonstration of relevance within an end-to-end automatic visual recognition pipeline. While such integrated systems are still in their infancy, promising demonstrations have recently emerged for computer vision tasks like scene classification \cite{BorjiItti2011}, verifying correlations with object (pedestrian) detection responses\cite{ElazaryItti2010,EhingerEtAl2009}. An interesting early biologically inspired recognition system was presented by Kienzle et al.\cite{KienzleEtAl2007}, who learn a fixation operator from human eye movements collected under video free-viewing, then learn action classification models for the KTH dataset with promising results. Recently, under the constraint of a first person perspective, Fathi et. al.\cite{FathiLiRehg2012} have shown that human fixations can be predicted and used to enhance action recognition performance.

In contrast, in computer vision, interest point detectors have been successfully used in the bag-of-visual-words framework for action classification and event detection\cite{Laptev2005,MarszalekEtAl2009,DongEtAl2009,WangEtAl2011}, but a variety of other methods exists, including random field models \cite{YamatoEtAl1992,LiEtAl2008} and stuctured output SVMs\cite{HoaiDeLaTorre2012}. Currently the most successful systems remain the ones dominated by complex features extracted at interesting locations, bagged and fused using advanced kernel combination techniques \cite{MarszalekEtAl2009,DongEtAl2009}. This study is driven, primarily, by our computer vision interests, yet leverages data collection and insights from human vision. While in this paper we focus on bag-of-words spatio-temporal computer action recognition pipelines, the scope for study and the structure in the data are broader. We do not see this investigation as a terminus, but rather as a first step in exploring some of the most advanced data and models that human vision and computer vision can offer at the moment.

\section{Large Scale Human Eye Movement Data Collection in Video}\label{s:dataset}

An objective of this work is to introduce additional annotations in the form of eye recordings for two large scale video data sets for action recognition.\\

\vspace{-2.7mm}

\noindent\textbf{The Hollywood-2 Movie Dataset}: Introduced in \cite{MarszalekEtAl2009}, it is one of the largest and most challenging available datasets for real world actions. It contains 12 classes: answering phone, driving a car, eating, fighting, getting out of a car, shaking hands, hugging, kissing, running, 
sitting down, sitting up and standing up. These actions are collected from a set of 69 Hollywood movies. The data set is split into a training set of 823 sequences and a test set of 884 sequences. There is no overlap between the 33 movies in the training set and the 36 movies in the test set. The data set consists of about 487k frames, totaling about 20 hours of video.\\

\vspace{-2.7mm}

\noindent\textbf{The UCF Sports Action Dataset}: This high resolution dataset \cite{RodriguezEtAl2008} was collected
mostly from broadcast television channels. It contains 150 videos covering 9 sports action classes: diving, golf swinging, kicking, lifting, horseback riding, running, skateboarding, swinging and walking.\\

\vspace{-2.7mm}

\noindent\textbf{Human subjects}:  We have collected data from 16 human volunteers (9 male and 7 female) aged between 21 and 41. We split them into an active group, which had to solve an action recognition task, and a free viewing group, which was not required to solve any specific task while being presented with the videos in the two datasets. There were 12 active subjects (7 male and 5 female) and 4 free viewing subjects (2 male and 2 female). None of the free viewers was aware of the task of the active group and none was a cognitive scientist. We chose the two groups such that no pair of subjects from different groups were acquainted with each other, in order to limit biases on the free viewers.\\

\vspace{-2.7mm}

\noindent\textbf{Recording environment}: Eye movements were recorded using an SMI iView X HiSpeed 1250 tower-mounted eye tracker, with a sampling frequency of 500Hz. The head of the subject was placed on a chin-rest located at $60$ cm from the display. Viewing conditions were binocular and gaze data was collected from the right eye of the participant.\footnote{None of our subjects exhibited a strongly dominant left eye, as determined by the two index finger method.} The LCD display had a resolution $1280 \times 1024$ pixels, with a physical screen size of $47.5$ x $29.5$ cm. The calibration procedure was carried out at the beginning of each block. The subject had to follow a target that was placed sequentially at $13$ locations evenly distributed across the screen. Accuracy of the calibration was then validated at $4$ of these calibrated locations. If the error in the estimated position was greater than $0.75^\circ$ of visual angle, the experiment was stopped and calibration restarted. At the end of each block, validation was carried out again, to account for fluctuations in the recording environment. If the validation error exceeded $0.75^\circ$ of visual angle, the data acquired during the block was deemed noisy and discarded from further analysis. Because the resolution varies across the datasets, each video was rescaled to fit the screen, preserving the original aspect ratio. The visual angles subtended by the stimuli were $38.4^\circ$ in the horizontal plane and ranged from $13.81^\circ$ to $26.18^\circ$ in the vertical plane.\\

\vspace{-2.7mm}

\noindent\textbf{Recording protocol}: Before each video sequence was shown, participants in the active group were required to fixate the center of the screen. 
Display would proceed automatically using the trigger area-of-interest feature provided by the iView X software. 
Participants had to identify the actions in each video sequence.
Their multiple choice answers were recorded through a set of check-boxes displayed at the end of each video, which the subject manipulated using a mouse.\footnote{While the representation of actions is an open problem, we relied on the datasets and labeling of the computer vision community, as a first study, and to maximize impact on current research. In the long run, weakly supervised learning could be better suited to map persistent structure to higher level semantic action labels.} Participants in the free viewing group underwent a similar protocol, the only difference being that the questionnaire answering step was skipped. To avoid fatigue, we split the set of stimuli into 20 sessions (for the active group) and 16 sessions (for the free viewing group), each participant undergoing no more than one session per day.
Each session consisted of 4 blocks, each designed to take approximately 8 minutes to complete (calibration excluded), with 5-minute breaks between blocks.
Overall, it took approximately 1 hour for a participant to complete one session. The video sequences were shown to each subject in a different random order.

\section{Spatial and Sequential Consistency}\label{s:consistency}

\subsection{Action Recognition by Humans}

Our goal is to create a data set that captures the gaze patterns of humans solving a recognition task. Therefore, it is important to ensure that our subjects are successful at this task. \Figref{fig:recog} shows the confusion matrix between the answers of human subjects and the ground truth. For Hollywood-2, there can be multiple labels associated with the same video. We show, apart from the 12 action labels, the 4 most common combinations of labels occurring in the ground truth and omit less common ones. The analysis reveals, apart from near-perfect performance, the types of errors humans are prone to make.
The most frequent human errors are omissions of one of the actions co-occurring in a video. False positives are much less frequent. The third type of error of mislabeling a video entirely, almost never happens, and when it does it usually involves semantically related actions, \textit{e.g.} \textit{DriveCar} and \textit{GetOutOfCar} or \textit{Kiss} and \textit{Hug}.

\begin{figure}
\scalebox{0.39}{\includegraphics[viewport=1cm 7cm 21cm 21cm]{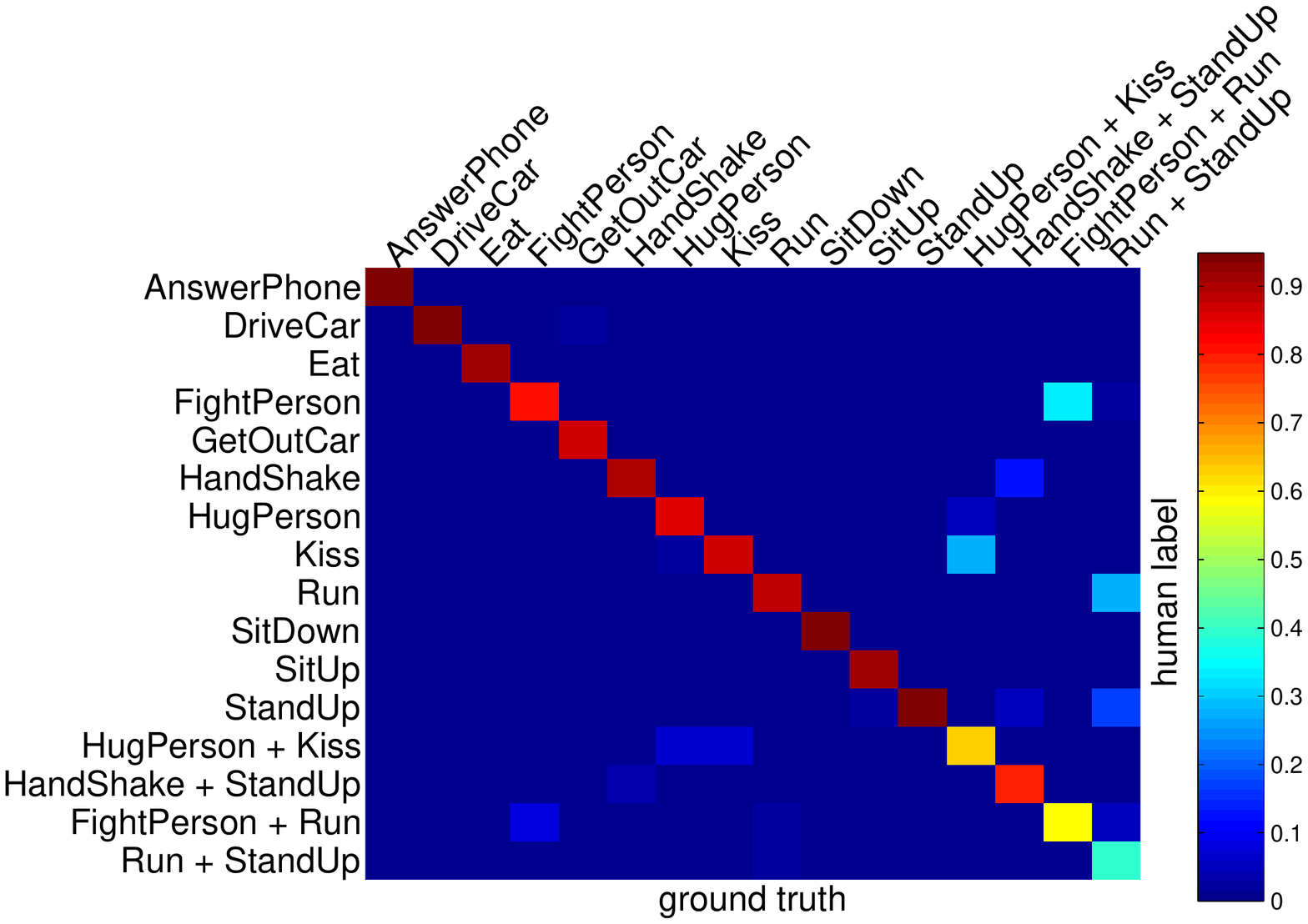}}
\caption{Action recognition perfomance by humans on the Hollywood-2 database. The confusion matrix includes the 12 action labels plus the 4 most frequent combinations of labels in the groundc truth.}\label{fig:human_recognition}
\end{figure}\label{fig:recog}

\subsection{Spatial Consistency Among Subjects}

In this section, we investigate how well the regions fixated by human subjects agree on a frame by frame basis, by generalizing to video data the procedure used by Ehinger et. al. \cite{EhingerEtAl2009} in the case of static stimuli. \\

\vspace{-2.7mm}

\noindent\textbf{Evaluation Protocol}: For the task of locating people in a static image, \cite{EhingerEtAl2009} have evaluated how well one can predict the regions fixated by a particular subject from the regions fixated by the other subjects on the same image. For cross-stimulus control, this measure is however not meaningful in itself, as part of the inter-subject agreement is due to bias in the stimulus itself (\textit{e.g.} photographer's bias) or due to the tendency of humans to fixate more often at the center of the screen\cite{LandTatler2009}. Therefore, one can address this issue by checking how well the fixation of a subject on one simulus can be predicted from those of the other subjects on a different, unrelated, stimulus. Normally, the average precision when predicting fixations on the same stimulus is much greater than on different stimuli.

We generalize this protocol for video, by randomly choosing frames from our videos and checking inter-subject correlation on them. We test each subject in turn with respect to the other (training) subjects. An empirical saliency map is generated by adding a Dirac impulse at each pixel fixated by the training subjects in that frame, followed by the application of a Gaussian blur filter. We then consider this map as a confidence map for predicting the test subject's fixation. There is a label of 1 at the test subjects' fixation and 0 elsewhere. The area under the curve (AUC) score for this classification problem is then computed for the test subject. We average score over all test subjects defines the final consistency metric. This value ranges from 0.5 -- when no consistency or bias effects are present in the data -- and 1 -- when all subjects fixate the same pixel and there is no measurement noise. For cross-stimulus control, we repeat this process for pairs of frames chosen from different videos and attempt to predict the fixation of each test subject on the first frame from the fixations of the other subjects on the other frame.

Unlike in the procedure followed in \cite{EhingerEtAl2009}, who considered several fixations per subject for each exposed image, we only consider the single fixation, if any, that a subject made on that frame. The reason is that our stimulus is dynamic and the spatial positions of future fixations are bound to be altered by changes in the stimulus itself. In our experiment, we set the width of the Gaussian blur kernel to match a visual angle span of $1.5^{\text{o}}$. We draw 1000 samples for both the same-stimulus and different stimulus predictions. We disregard the first 200ms from the beginning of each video to remove bias due to the initial central fixation.\\

\begin{figure}
\begin{center}
\begin{tabular}{cc}
\scalebox{0.23}{\includegraphics[viewport=1.9cm 7cm 20cm 22cm]{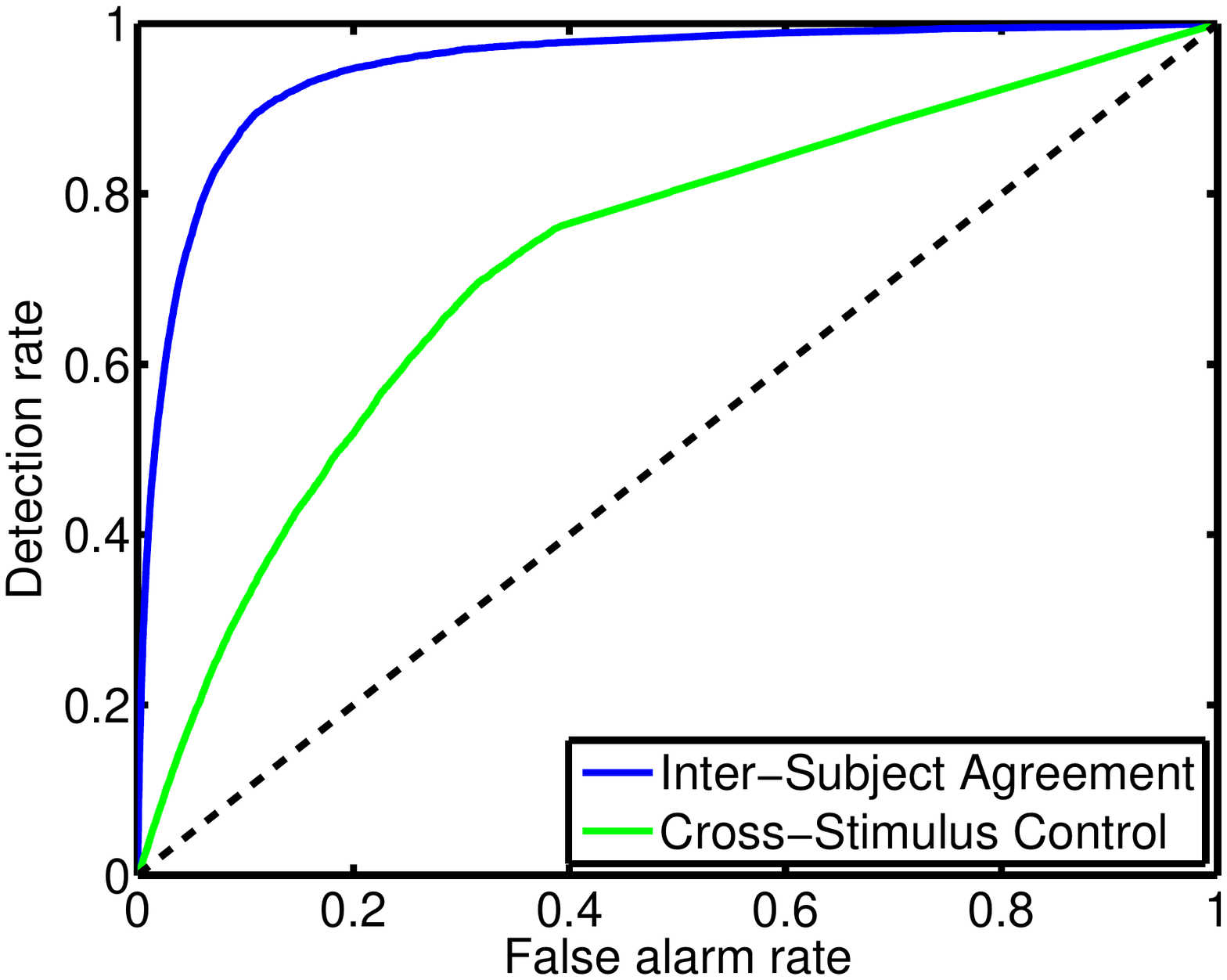}}
&
\scalebox{0.23}{\includegraphics[viewport=1.9cm 7cm 20cm 22cm]{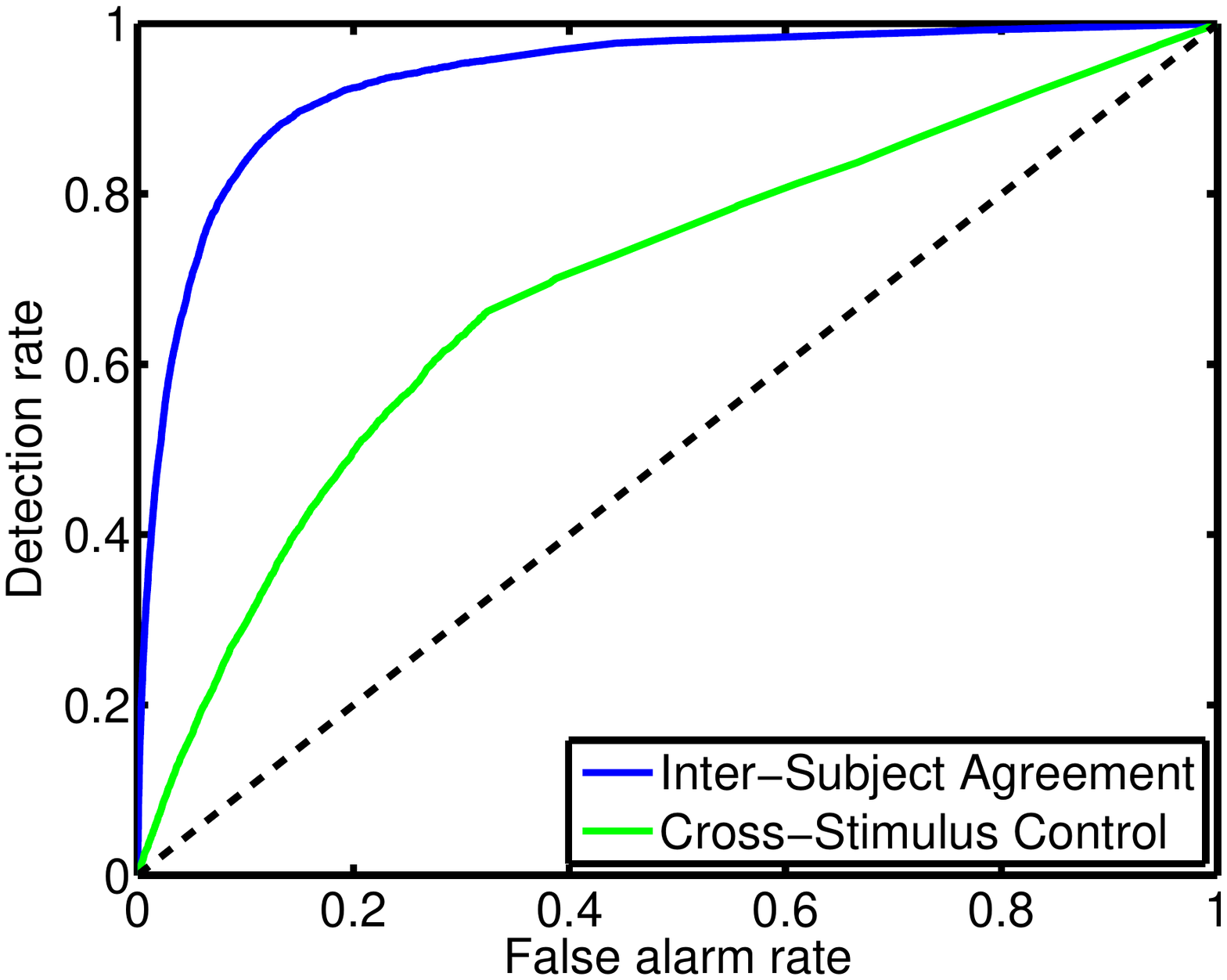}}
\\
Hollywood-2
&
UCF Sports Actions
\end{tabular}
\end{center}
\caption{Spatial inter-subject agreement. The ROC curves correspond to predicting the fixations of one subject from the fixations of the other subjects on the same video frame (blue) or on a different video (green) randomly selected from the dataset.}\label{fig:static_consistency}
\end{figure}

\vspace{-2.7mm}

\noindent\textbf{Findings}: The ROC curves for inter-subject agreement and cross-stimulus control are shown in \figref{fig:static_consistency}. For the Hollywood-2 dataset, the AUC score is $94.8\%$ for inter-subject agreement and $72.3\%$ for cross-stimulus control. For UCF Sports, we obtain values of $93.2\%$ and $69.2\%$. These values are consistent with the results reported for static stimuli by \cite{EhingerEtAl2009}, with slightly higher cross-stimulus control. This suggests that shooter's bias is stronger in artistic datasets (movies) than in natural scenes, a trend that has been observed by \cite{DorrEtAl2010} for human observers free viewing Hollywood movie trailers as opposed to video shoots of outdoor scenes. 

We also analyze inter-subject agreement on subsets of videos corresponding to each action class and for 4 significant labelings considered in  \figref{fig:recog}. On each of these subsets, inter-subject consistency remains strong, as illustrated in Table \ref{table:static_consistency}a,b. Interestingly, there is significant variation in the cross-stimulus control across these classes, especially in the UCF Sports dataset. The setup being identical, we conjecture that part of this variation is due to the different ways in which various categories of scenes are filmed and the way the director aims to present the actions to the viewer. For example, in TV footage for sports events, the actions are typically shot from a frontal pose, the peformer is centered almost perfectly and there is limited background clutter. These factors lead to a high degree of similarity among the stimuli within the class and makes it easier to extrapolate subject fixations across videos, explaining the unusually high values of this metric for the actions like \textit{Dive}, \textit{Lift} and \textit{Swing}.

\subsection{The Influence of Task on Eye Movements}

We next evaluate the impact of the task on eye movements for this data set. For a given video frame, we compute the fixation probability distribution using data from all active subjects. Then, for each free viewer, we compute the $p$-statistic of the fixated location with respect to this distribution. We repeat this process for 1000 randomly sampled frames and compute the average $p$-value for each subject. Somewhat surprisingly, we find that none of our free viewers exhibits a fixation pattern that deviates significantly from that of our active group (see Table \ref{t:consistency_active_control}). Since in the Hollywood-2 dataset several actions can be present in a video, either simultaneously or sequentially, this rules out initial habituation effects and further neglect (free viewing) to some degree.\footnote{Our findings do not assume or imply that free-viewing subjects may not be recognizing actions. However we did not ask them to perform a task, nor where they aware of the purpose of the experiment, or the interface presented to subjects given a task. While this is one approach to analyze task influence, it is not the only possible. For instance, subjects may be asked to focus on different tasks (e.g. actions versus general scene recognition), although this setting may induce biases due to habituation with stimuli presented at least twice.}

Similar lack of discriminability between tasks has been remarked by \textit{Greene et al.}\cite{GreeneEtAl2012} for static stimuli. Their study shows that classifiers trained on human scanpaths can successufully discriminate observers or stimuli, but cannot classify the tasks of the observers. Although our results may seem to support the generalization of such observations to the video domain, we mention here two reasons why we believe no definite conclusion can be drawn at this point. First of all, we compare the action recognition task to a no-task condition, rather than discriminate between   several specific task conditions. Second, in \cite{GreeneEtAl2012} the free viewers have a relatively large amount of time (10s) to solve the task. Although the authors analyze whether the first 2 seconds of viewing are discriminative for the task -- and find it not to be the case -- it is still unclear to what extent the somewhat lax time constraints will direct the observer to exhibit task-specific eye movements or to focus such behavior early on during exposure. On the other hand, the video domain has an intrinsic and variable exposure time induced by the changing scene, which may, in principle, evoke task-specific behavior during episodes of intense dynamic activity.

\begin{table*}
\caption{Spatial and Sequential Consistency Analysis}\label{table:static_consistency}
\begin{center}
\scalebox{0.95}{
\begin{tabular}{|C{3cm}|C{2cm}|C{2cm}|c|C{2cm}|C{2cm}|c|C{2cm}|C{2cm}|}
\hline
                    & \multicolumn{2}{c|}{\textbf{spatial consistency}} & & \multicolumn{2}{c|}{\textbf{temporal AOI alignment}} & & \multicolumn{2}{c|}{\textbf{AOI Markov Dynamics}} \\
\cline{2-9}
\textbf{Label}      & \textbf{inter-subject} & \textbf{cross-stimulus} & & \textbf{inter-subject}    & \textbf{random}          & & \textbf{inter-subject}    & \textbf{random}          \\
                    & \textbf{agreement}     & \textbf{control}        & & \textbf{agreement}        & \textbf{control}         & & \textbf{agreement}        & \textbf{control}        \\
                    & \textbf{(a)}           & \textbf{(b)}            & & \textbf{(c)}              & \textbf{(d)}             & & \textbf{(e)}              & \textbf{(f)}             \\
\hline
AnswerPhone         & 94.9\%                 & 69.5\%                  & & 71.1\%                    & 51.2\%                   & & 65.7\%                    & 15.3\% \\
\hline
DriveCar            & 94.3\%                 & 69.9\%                  & & 70.4\%                    & 53.6\%                   & & 78.5\%                    & 8.7\% \\
\hline
Eat                 & 93.0\%                 & 69.7\%                  & & 73.2\%                    & 50.2\%                   & & 79.9\%                    & 7.9\% \\
\hline
FightPerson         & 96.1\%                 & 74.9\%                  & & 66.2\%                    & 48.3\%                   & & 78.5\%                    & 9.8\% \\
\hline
GetOutCar           & 94.5\%                 & 69.3\%                  & & 72.5\%                    & 52.0\%                   & & 68.1\%                    & 13.8\% \\
\hline
HandShake           & 93.7\%                 & 72.6\%                  & & 69.3\%                    & 50.6\%                   & & 68.4\%                    & 14.3\% \\
\hline
HugPerson           & 95.7\%                 & 75.4\%                  & & 71.7\%                    & 53.2\%                   & & 71.7\%                    & 12.9\% \\
\hline
Kiss                & 95.6\%                 & 72.1\%                  & & 68.8\%                    & 49.9\%                   & & 71.5\%                    & 12.7\% \\
\hline
Run                 & 95.9\%                 & 72.3\%                  & & 76.9\%                    & 54.9\%                   & & 69.6\%                    & 13.8\% \\
\hline
SitDown             & 93.5\%                 & 68.4\%                  & & 68.3\%                    & 49.7\%                   & & 69.7\%                    & 12.7\% \\
\hline
SitUp               & 95.4\%                 & 72.0\%                  & & 71.9\%                    & 54.1\%                   & & 64.0\%                    & 16.1\% \\
\hline
StandUp             & 94.3\%                 & 69.2\%                  & & 67.3\%                    & 53.9\%                   & & 65.1\%                    & 15.1\% \\
\hline
HugPerson + Kiss    & 92.8\%                 & 70.6\%                  & & 68.2\%                    & 52.1\%                   & & 71.2\%                    & 13.1\% \\
\hline
HandShake + StandUp & 91.3\%                 & 60.9\%                  & & 70.5\%                    & 51.0\%                   & & 65.7\%                    & 15.0\% \\
\hline
FightPerson + Run   & 96.1\%                 & 66.6\%                  & & 73.4\%                    & 52.3\%                   & & 72.1\%                    & 12.5\% \\
\hline
Run + StandUp       & 92.8\%                 & 66.4\%                  & & 68.3\%                    & 52.4\%                   & & 67.4\%                    & 14.4\% \\
\hline
\textbf{Any}        & \textbf{94.8\%}        & \textbf{72.3\%}         & & \textbf{70.8\%}           & \textbf{51.8\%}          & & \textbf{70.2\%}          & \textbf{12.7\%} \\
\hline
\end{tabular}
}

\vspace{1mm}

Hollywood-2

\end{center}

\vspace{1mm}

Spatial and sequential consistency analysis results measured both globally and for each action class. Columns (a-b) represent the areas under the ROC curves for spatial inter-subject consistencies and the corresponding cross-stimulus control. The classes marked by $^{*}$ show significant shooter's bias due to the very similar filming conditions of all the videos within them (same shooting angle and position, no clutter). Good sequential consistency is revealed by the match scores obtained by temporal alignment (c-d) and Markov dynamics (e-f).
\end{table*}

\begin{table}
\caption{Spatial Consistency Between Active and Free-Viewing Subjects}\label{t:consistency_active_control}
\begin{center}
\vspace{-0.3cm}
\scalebox{0.85}{
	\begin{tabular}{|c|c|c|}
	\hline
	\textbf{free-viewing} & \textbf{p-value}     & \textbf{p-value} \\
	\textbf{subject}      & \textbf{Hollywood-2} & \textbf{UCF Sports} \\
	\hline
	1 & 0.67 & 0.55 \\
	\hline
	2 & 0.67 & 0.55 \\
	\hline
	3 & 0.60 & 0.60 \\
	\hline
	4 & 0.65 & 0.64 \\
	\hline
  \textbf{Mean} & \textbf{0.65} & \textbf{0.58} \\
	\hline
	\end{tabular}
}

\vspace{2mm}
(a)

\end{center}
\begin{minipage}{0.45\linewidth}
\begin{center}
\scalebox{0.85}{
\begin{tabular}{|c|c|c|}
\hline
\textbf{Action Class} & \textbf{p-value} \\
\hline
AnswerPhone           & 0.66 \\
\hline
DriveCar              & 0.65 \\
\hline
Eat                   & 0.65 \\
\hline
FightPerson           & 0.65 \\
\hline
GetOutCar             & 0.65 \\
\hline
HandShake             & 0.68 \\
\hline
HugPerson             & 0.63 \\
\hline
Kiss                  & 0.63 \\
\hline
Run                   & 0.65 \\
\hline
SitDown               & 0.61 \\
\hline
SitUp                 & 0.65 \\
\hline
StandUp               & 0.64 \\
\hline
\textbf{Mean}         & \textbf{0.65} \\
\hline
\end{tabular}
}
\vspace{1mm}

Hollywood-2

\vspace{3mm}
\end{center}
\end{minipage}
\begin{minipage}{0.45\linewidth}
\begin{center}
\scalebox{0.85}{
\begin{tabular}{|c|c|c|}
\hline
\textbf{Action Class} & \textbf{$p$-value} \\
\hline
Dive                  & 0.68 \\
\hline
GolfSwing            & 0.67 \\
\hline
Kick                  & 0.58 \\
\hline
Lift                  & 0.57 \\
\hline
RideHorse            & 0.51 \\
\hline
Run                   & 0.54 \\
\hline
Skateboard            & 0.60 \\
\hline                
Swing                 & 0.59 \\
\hline                
Walk                  & 0.51 \\
\hline
\textbf{Mean}         & \textbf{0.58} \\
\hline
\end{tabular}
}
\vspace{1mm}

UCF Sports Actions

\vspace{3mm}
\end{center}
\end{minipage}
\begin{center}
\vspace{-4mm}
(b)
\end{center}
We measure consistency as the $p$-value associated with predicting each free-viewer from the saliency map derived from active subjects, averaged over 1000 randomly chosen video frames. We find that none of the scanpaths belonging to our free-viewing subjects (a) deviates significanly from ground truth data. Likewise, no significant differences are found when restricting the analysis to videos belonging to specific action classes (b).
\end{table}

\subsection{Sequential Consistency Among Subjects}
Our spatial inter-subject agreement analysis shows that the spatial distribution of fixations in video is highly consistent across subjects. It does not however reveal whether there is significant consistency in \emph{the order} in which subjects fixate among these locations. To our knowledge, there are no agreed upon sequential consistency measures in the community at the moment\cite{LeMeurBaccino2012}. In this section, we propose two metrics that are sensitive to the temporal ordering among fixations and evaluate consistency under them. We first model the scanpath made by each subject as a sequence of discrete symbols and show how this representation can be produced \emph{automatically}. We then define two metrics, \emph{AOI Markov dynamics} and \emph{temporal AOI alignment}, and show how they can be computed for this representation. After we define a baseline for our evaluation we conclude with a discussion of the results.

\begin{figure*}
\begin{minipage}{0.6\linewidth}
\hspace{-0.3cm}
\center
\begin{tabular}{ccc}
\scalebox{0.16}{\includegraphics[viewport=0.3cm 10.2cm 21.0cm 18.5cm]{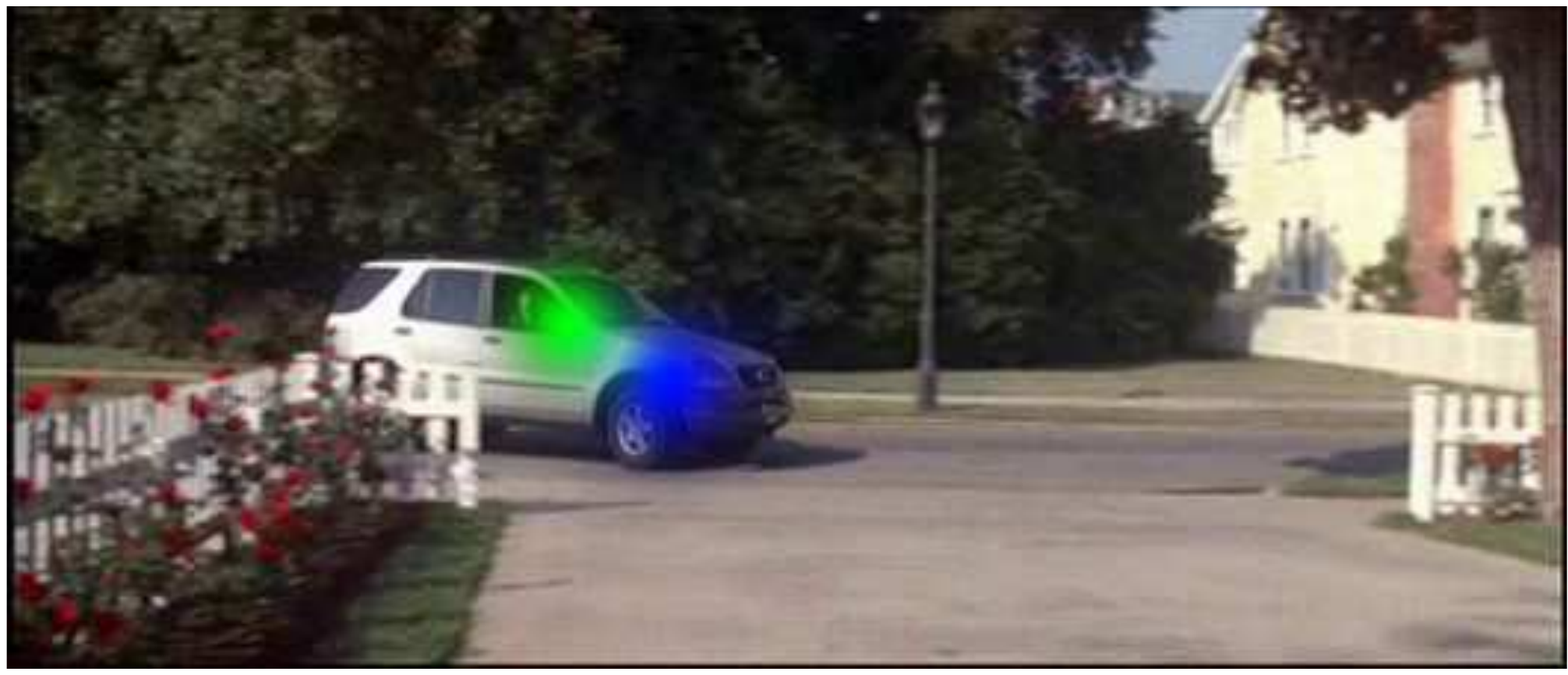}}
&
\scalebox{0.16}{\includegraphics[viewport=0.3cm 10.2cm 21.0cm 18.5cm]{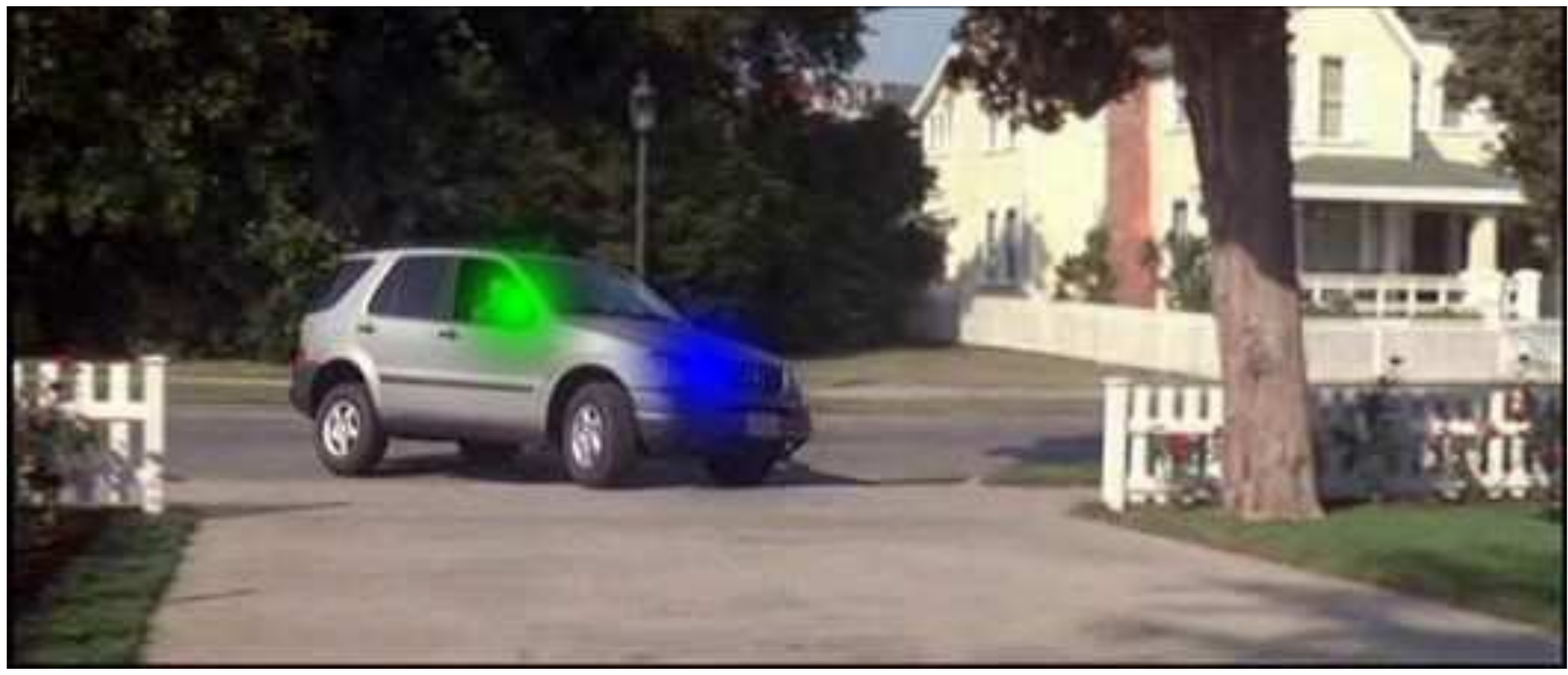}}
&
\scalebox{0.16}{\includegraphics[viewport=0.3cm 10.2cm 21.0cm 18.5cm]{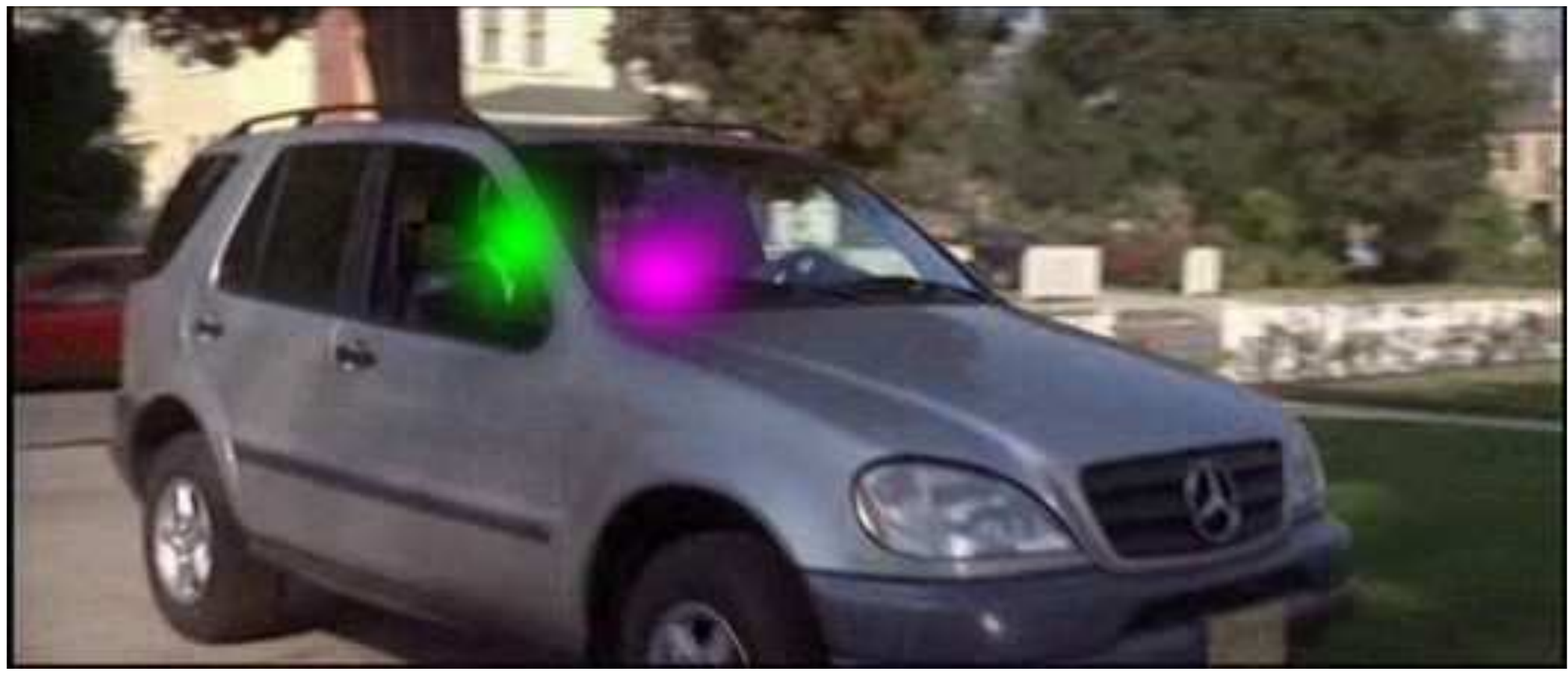}}
\\
\scalebox{0.8}{frame 10}
&
\scalebox{0.8}{frame 30}
&
\scalebox{0.8}{frame 90}
\\
\scalebox{0.16}{\includegraphics[viewport=0.3cm 10.2cm 21.0cm 18.5cm]{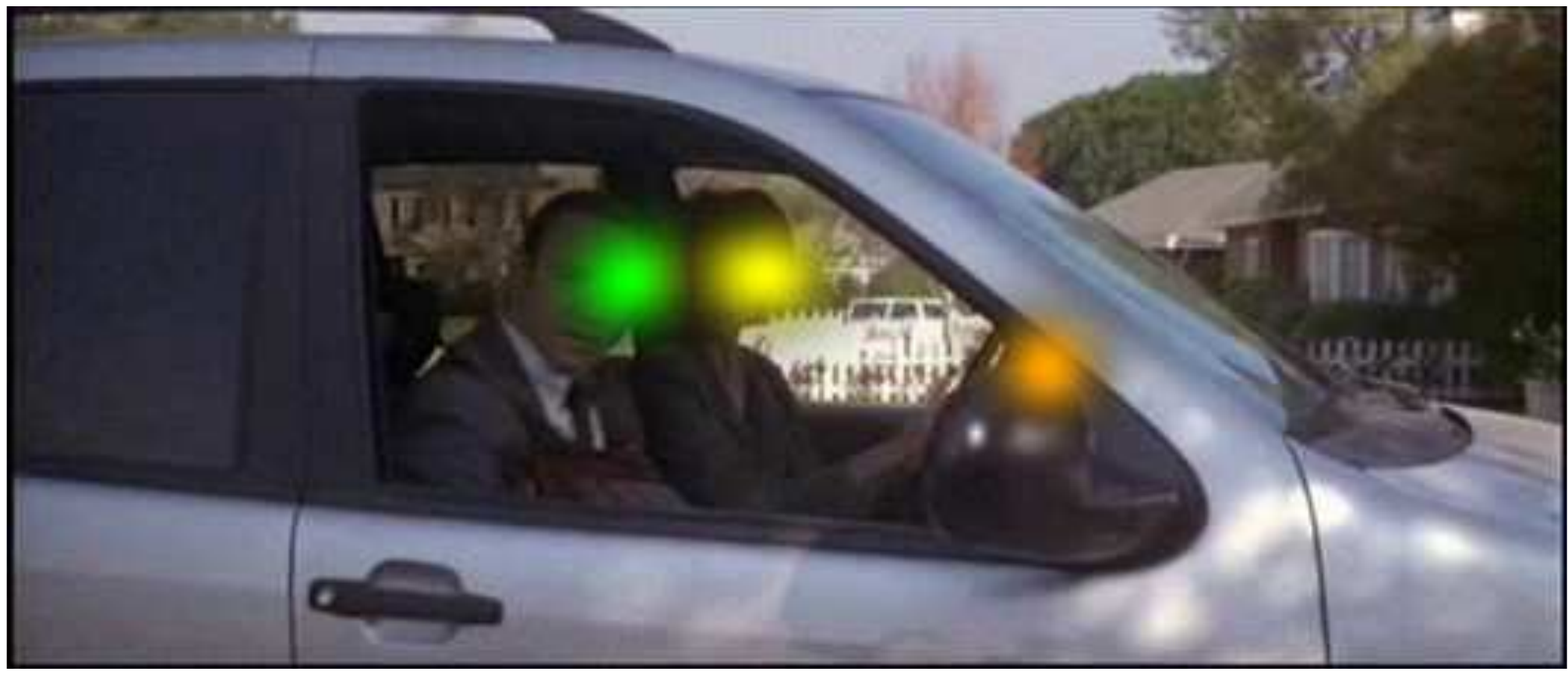}}
&
\scalebox{0.16}{\includegraphics[viewport=0.3cm 10.2cm 21.0cm 18.5cm]{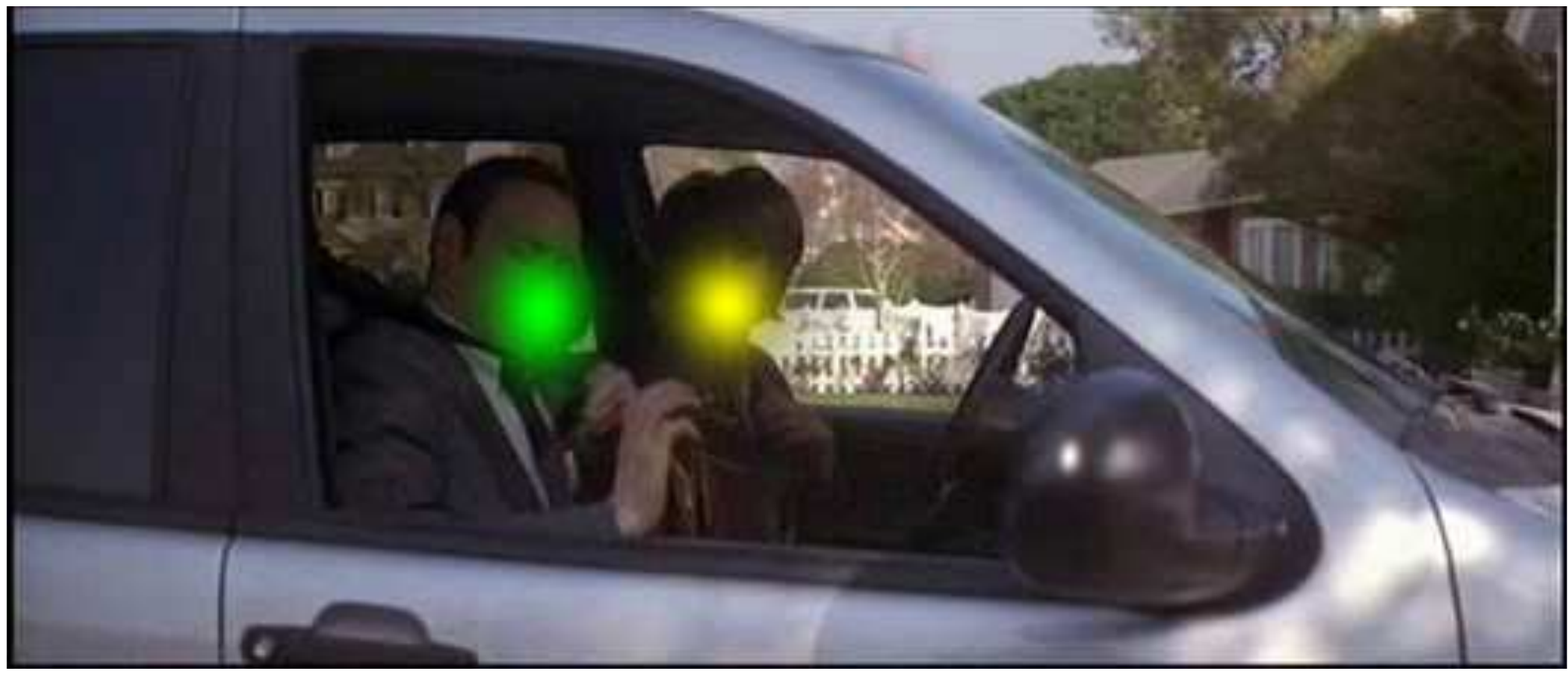}}
&
\scalebox{0.16}{\includegraphics[viewport=0.3cm 10.2cm 21.0cm 18.5cm]{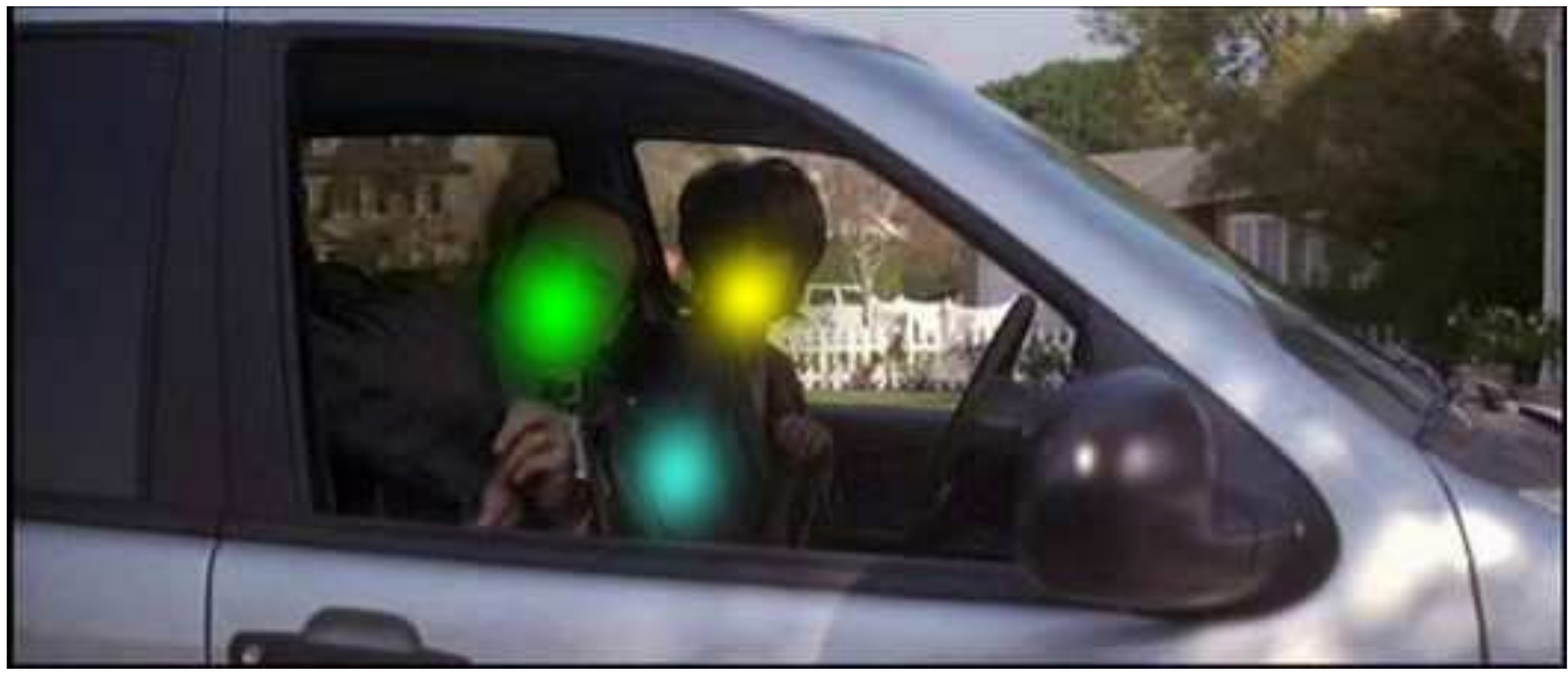}}
\\
\scalebox{0.8}{frame 145}
&
\scalebox{0.8}{frame 215}
&
\scalebox{0.8}{frame 230}
%&
%frame 230
\end{tabular}
\end{minipage}
\begin{minipage}{0.4\linewidth}
\vspace{5mm}
\center{\scalebox{0.35}{\includegraphics[viewport=2.0cm 15.5cm 19cm 27cm]{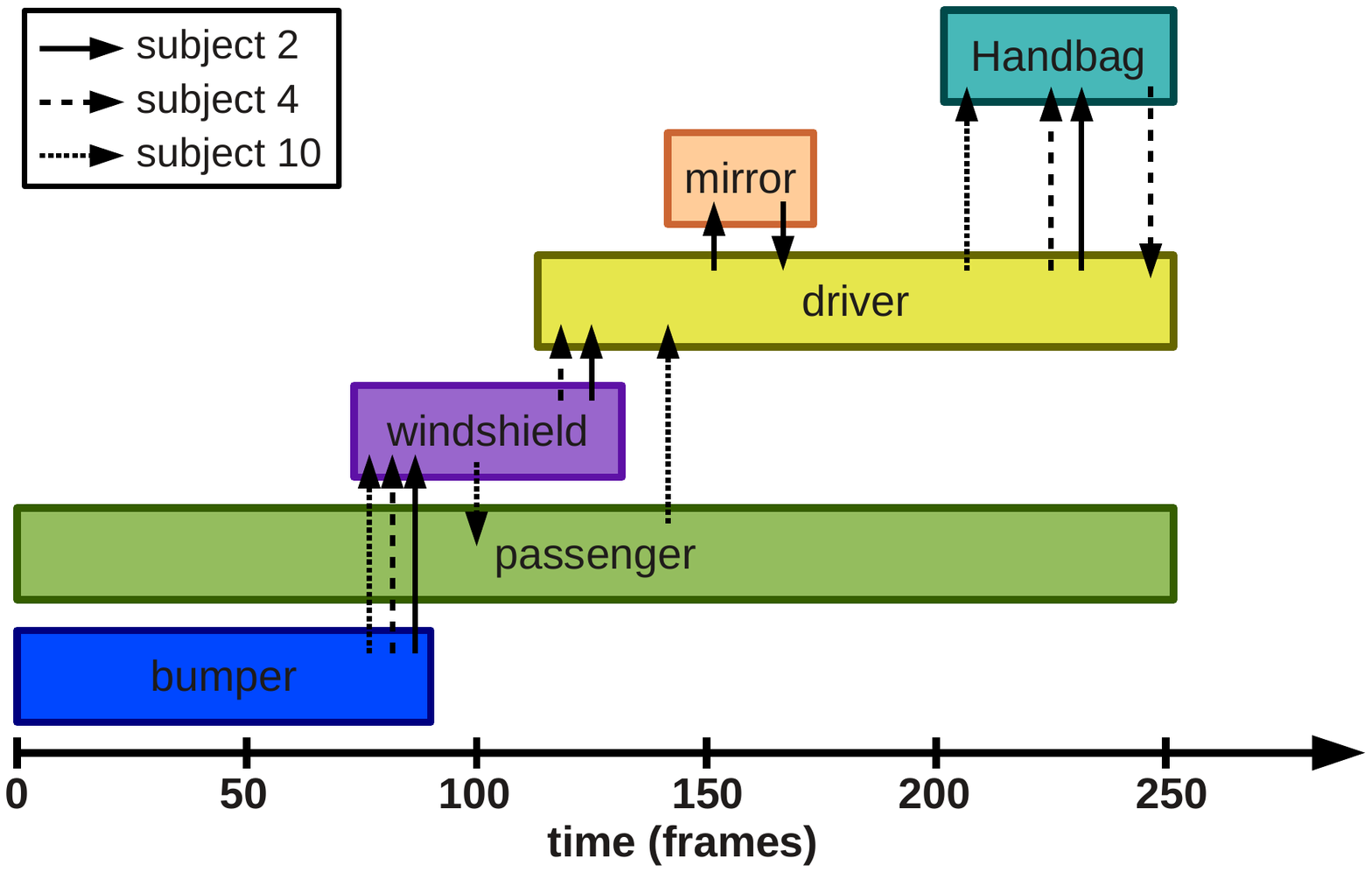}}}
\end{minipage}
\caption{Areas of interest are obtained \textit{automatically} by clustering the fixations of subjects.
\textbf{Left:} Heat maps illustrating the assignments of fixations to AOIs. The colored blobs
have been generated by pooling together all fixations belonging to the same AOI.
\textbf{Right:} Scan path through automatically generated AOIs (colored boxes) for three subjects.
Arrows illustrate saccades. Semantic labels have been manually assigned and illustrates the existance of cognitive routines centered at semantically meaningful objects.
}\label{fig:aoi}
\end{figure*}

\noindent \textbf{Scanpath Representation}: Human fixations tend to be tightly clustered spatially at one or more locations in the image. Assuming that such regions, called \emph{areas of interest} (AOIs), can be identified, the sequence of fixations belonging to a subject can be represented discretely by assigning each fixation to the closest AOI. For example, from the video  depicted in \figref{fig:aoi}-left, we identify six AOIs: the bumper of the car, its windshield, the passenger and the handbag he carries, the driver and the side mirror. We then trace the scan path of each subject through the AOIs based on spatial proximity, as shown in \figref{fig:aoi}-right. Each fixation gets assigned a label. For subject 2 shown in the example, this results in the sequence [\textit{bumper}, \textit{windshield}, \textit{driver}, \textit{mirror}, \textit{driver}, \textit{handbag}]. Notice that AOIs are semantically meaningful and tend to correspond to physical objects. Interestingly, this supports recent computer vision strategies based on object detectors for action recognition\cite{DongEtAl2009,YaoFei2010,PrestEtAl2011,HwangEtAl2011}.\\

\vspace{-2.7mm}

\noindent \textbf{Automatically Finding AOIs}: Defining areas of interest manually is labour intensive, especially in the video domain. Therefore, we introduce an \emph{automatic} method for determining their locations based on clustering the fixations of all subjects in a frame. We start by running k-means with 1 cluster and we successively increase their number until the sum of squared errors drops below a threshold. We then link centroids from successive frames into tracks, as long as they are closely located spatially. For robustness, we allow for a temporal gap during the track building process. Each resulting track becomes an AOI, and each fixation is assigned to the closest AOI at the time of its creation.\\

\vspace{-2.7mm}

\noindent \textbf{AOI Markov Dynamics}: In order to capture the dynamics of eye movements, and due to data sparsity, we represent the transitions of human visual attention between AOIs by means of a first-order Markov process. 
%We assume that the human visual system transitions to the next AOI conditioned on the previously visited AOIs. 
%Due to data sparsity, we restrict our analysis to a first order Markov process. 
Given a set of human fixation strings $\mathbf{f}_i$, where the $j^{\text{th}}$ fixation of subject $i$ is encoded by the index $f_i^j \in \overline{1,A}$ of the corresponding AOI, we estimate the probability $p(s_t=b \mid s_{t-1}=a)$ of transitioning to AOI $b$ at time $t$ given that AOI $a$ was fixated at time $t-1$ by counting transition frequencies. We regularize the model using Laplace smoothing to account for data sparsity. The probability of a novel fixation sequence $g$ under this model is $\prod_{j}p(s_t=g^{j}\mid s_{t-1}=g^{j-1})$ assuming the first state in the model, the central fixation, has probability 1. We measure the consistency among a set of subjects by considering each subject in turn, computing the probability of his scanpath with respect to the model trained from the fixations of the other subjects and normalizing by the number of fixations in his scanpath. The final consistency score is the average probability over all subjects.\\

\vspace{-2.7mm}

\noindent \textbf{Temporal AOI Alignment}: Another approach to evaluate sequential consistency is by measuring how pairs of AOI strings corresponding to different subjects can be globally aligned, based on their content. Although not modeling transitions explicitly, a sequence alignment has the advantage of being able to handle gaps and missing elements. An efficient algorithm having these properties due to Needleman-Wunsch\cite{NeedleManEtAl1970} uses dynamic programming to find the optimal match between two sequences $f^{1:n}$ and $g^{1:m}$, by allowing for the insertion of gaps in either sequence. It recursively computes the alignment score $h_{i,j}$ between subsequences $f^{1:i}$ and $g^{1:j}$ by considering the alternative costs of a match between $f^i$ and $g^j$ versus the insertion of a gap into either sequence. The final consistency metric is the average alignment score over all pairs of distinct subjects, normalized by the length of the longest sequence in each pair. We set the similarity metric to 1 for matching AOI symbols and to $0$ otherwise, and assume no penalty is incurred for inserting gaps in either sequence. This setting gives the score a semantic meaning: it is the average percentage of symbols that can be matched when determinig the longest common subsequence of fixations among pairs of subjects.\\

\vspace{-2.7mm}

\noindent \textbf{Baselines}: In order to provide a reference for our consistency evaluation, we generate 10 random AOI strings per video and compute the consistency on these strings under our metrics. We note however that the dynamics of the stimulus places constraints on the sampling process. First, a random string must obey the time ordering relations among AOIs (\textit{e.g.} the passenger is not visible until the second half of the video in \figref{fig:aoi}). Second, our automatic AOIs are derived from subject fixations and are biased by their gaze preference. The lifespan of an AOI will not be initiated until at least one subject has fixated it, even if the corresponding object is already visible. To remove some of the resulting bias from our evaluation, we extend each AOI both forward and backwards in time, until the image patch at its center has undergone significant appearance changes, and use these extended AOIs when generating our random baselines. \\

\vspace{-2.7mm}

\noindent \textbf{Findings}: For the Hollywod-2 dataset, we find that the average transition probability of each subject's fixations under AOI Markov dynamics is $70\%$, compared to $13\%$ for the random baseline (\tabref{table:static_consistency}). We also find that, across all videos, $71\%$ of the AOI symbols are successfully aligned, compared to only $51\%$ for the random baseline. We notice similar gaps in the UCF Sports dataset. These results indicate a high degree of consistency in human eye movement dynamics across the two datasets. Alignment scores vary to some extent across classes.
%The small variation in alignment scores across classes is also noticeable. 
%This is supportive of the conclusion that part of the correlation between the subjects is due to the dynamic nature of the stimulus, whereas the extent to which this consistency is induced by the task is still an open problem.

%\section{Learnt Saliency Models for Visual Action Recognition}
%
%In this section, we show that it is possible to train an effective human fixation detector on our dataset and present a state-of-the-art end-to-end automatic visual action recognition system based on the saliency maps generated by our detector.
\begin{figure}[ht]
\begin{center}
\begin{tabular}{cc}

\scalebox{0.24}{\includegraphics[viewport=3.0cm 7.0cm 18.5cm 21.7cm]{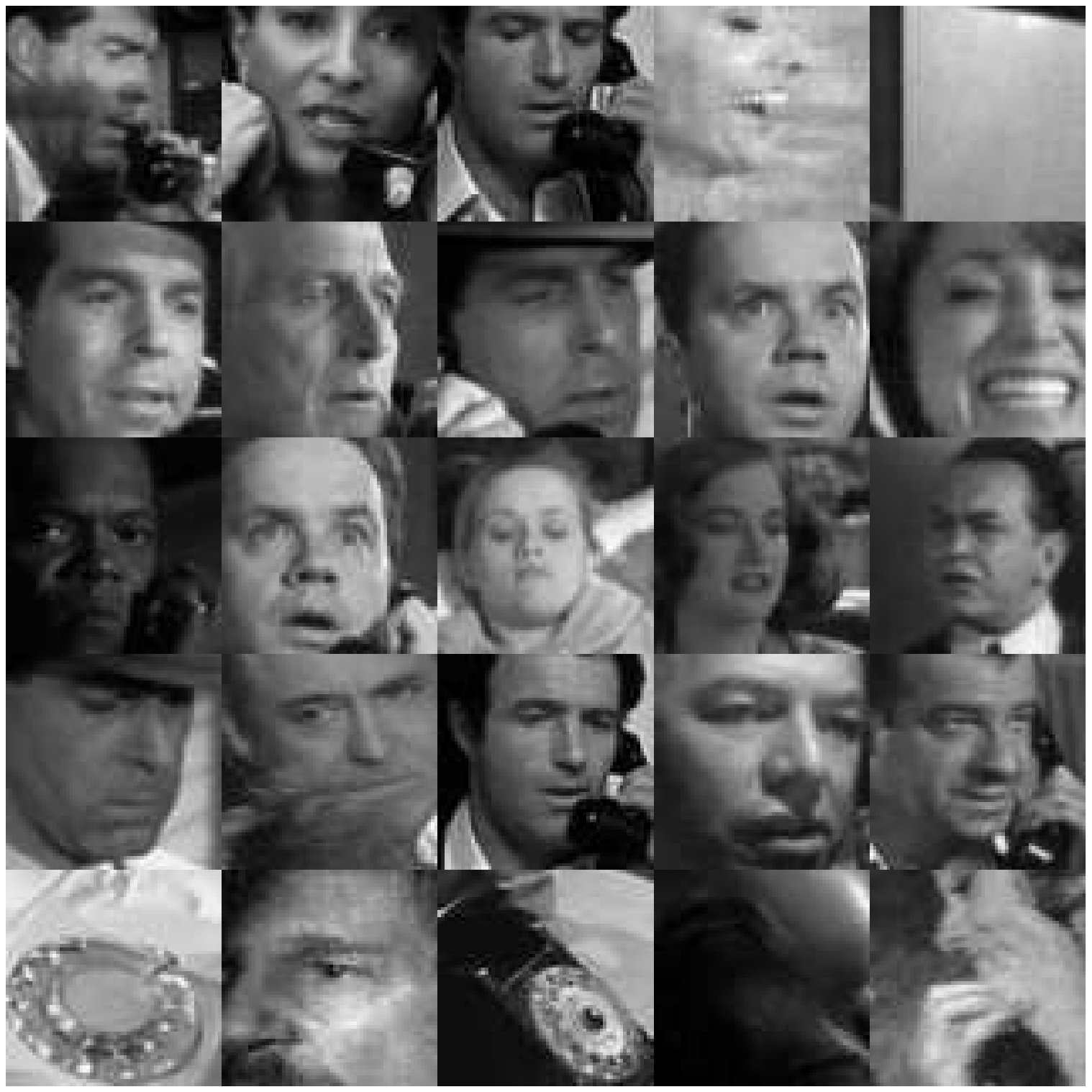}}

\vspace{0.5mm}

&

\scalebox{0.24}{\includegraphics[viewport=3.0cm 7.0cm 18.5cm 21.7cm]{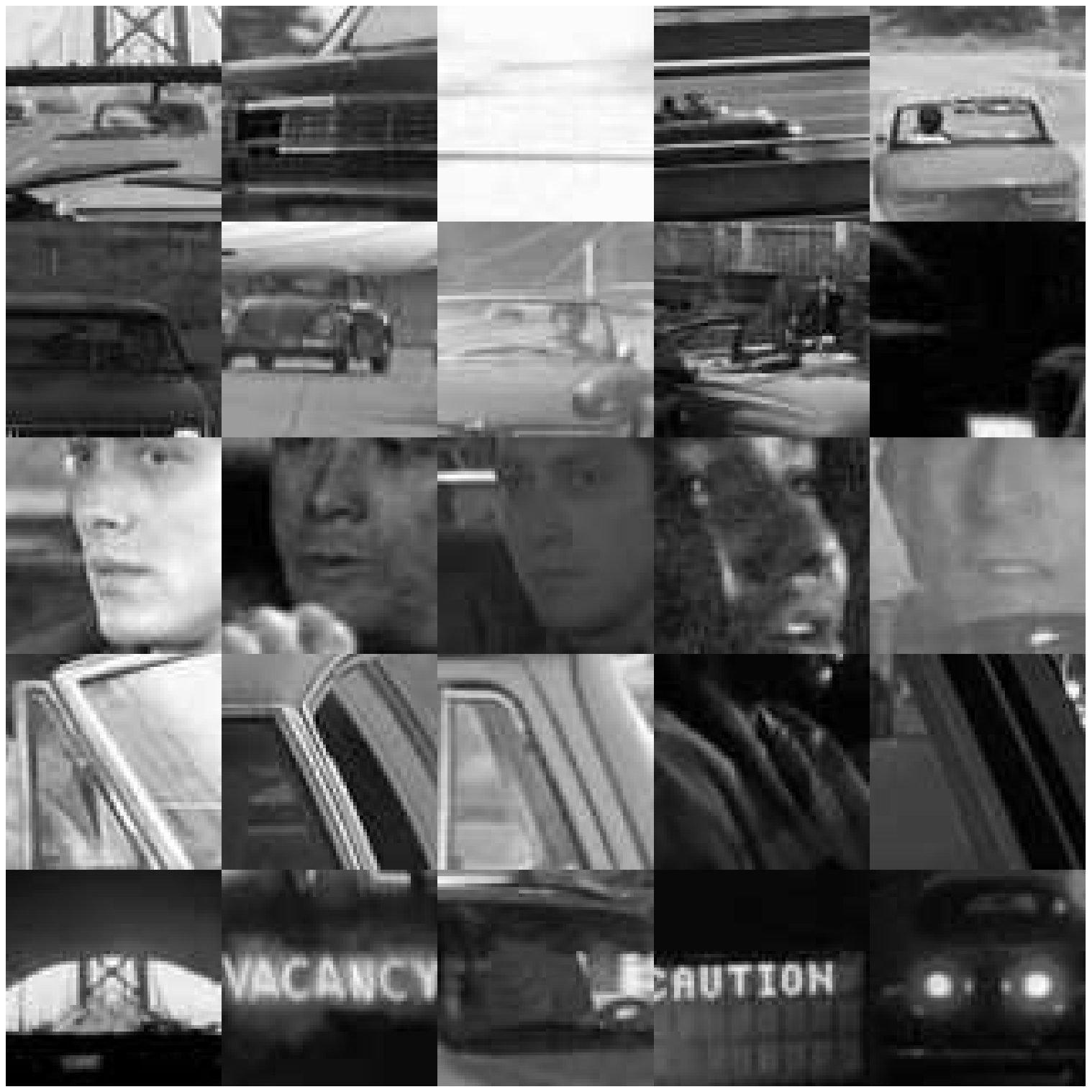}}

\vspace{0.5mm}

\\

\textbf{Answerphone}

\vspace{1.0mm}

&

\textbf{DriveCar}

\vspace{1.0mm}

\\

\scalebox{0.24}{\includegraphics[viewport=3.0cm 7.0cm 18.5cm 21.7cm]{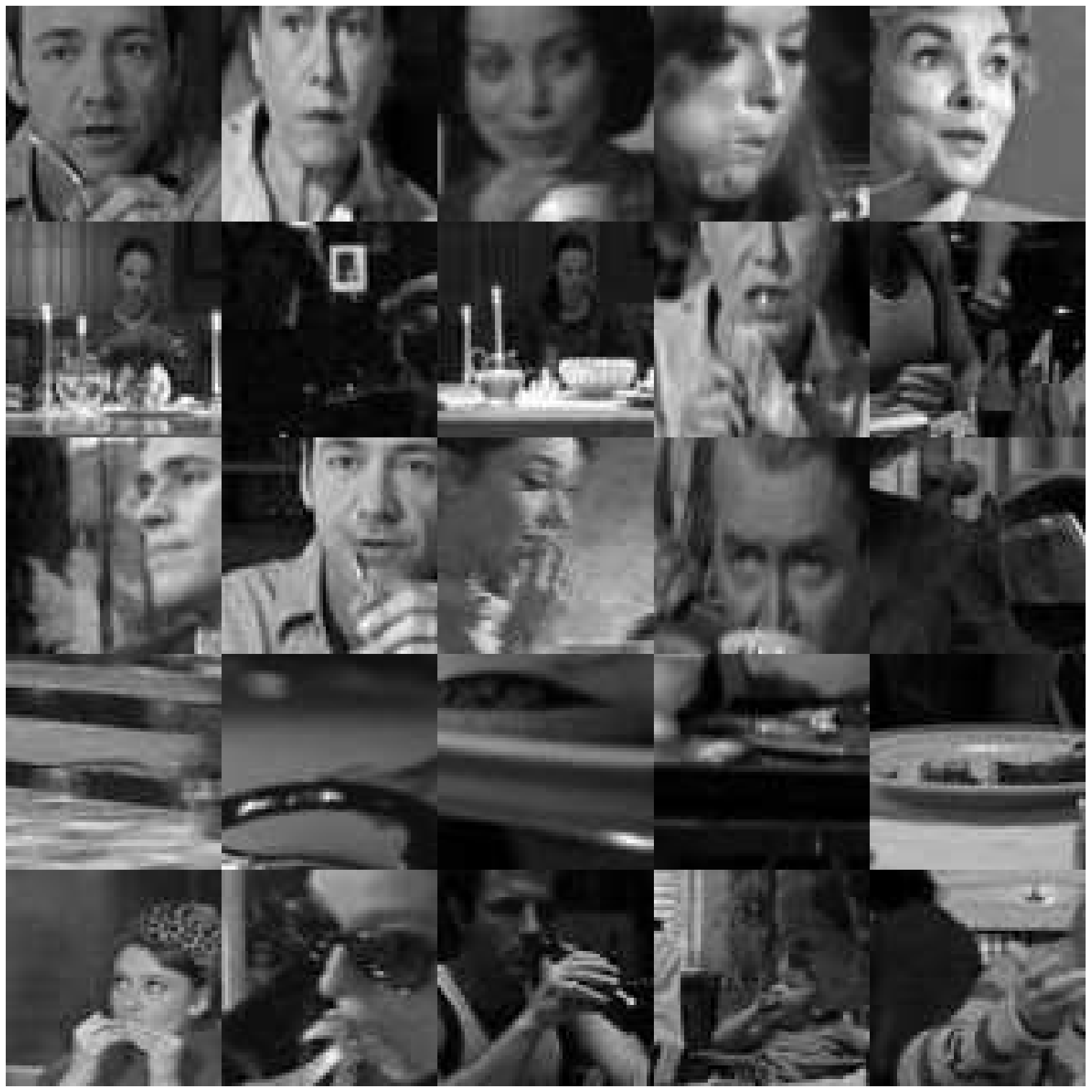}}

\vspace{0.5mm}

&

\scalebox{0.24}{\includegraphics[viewport=3.0cm 7.0cm 18.5cm 21.7cm]{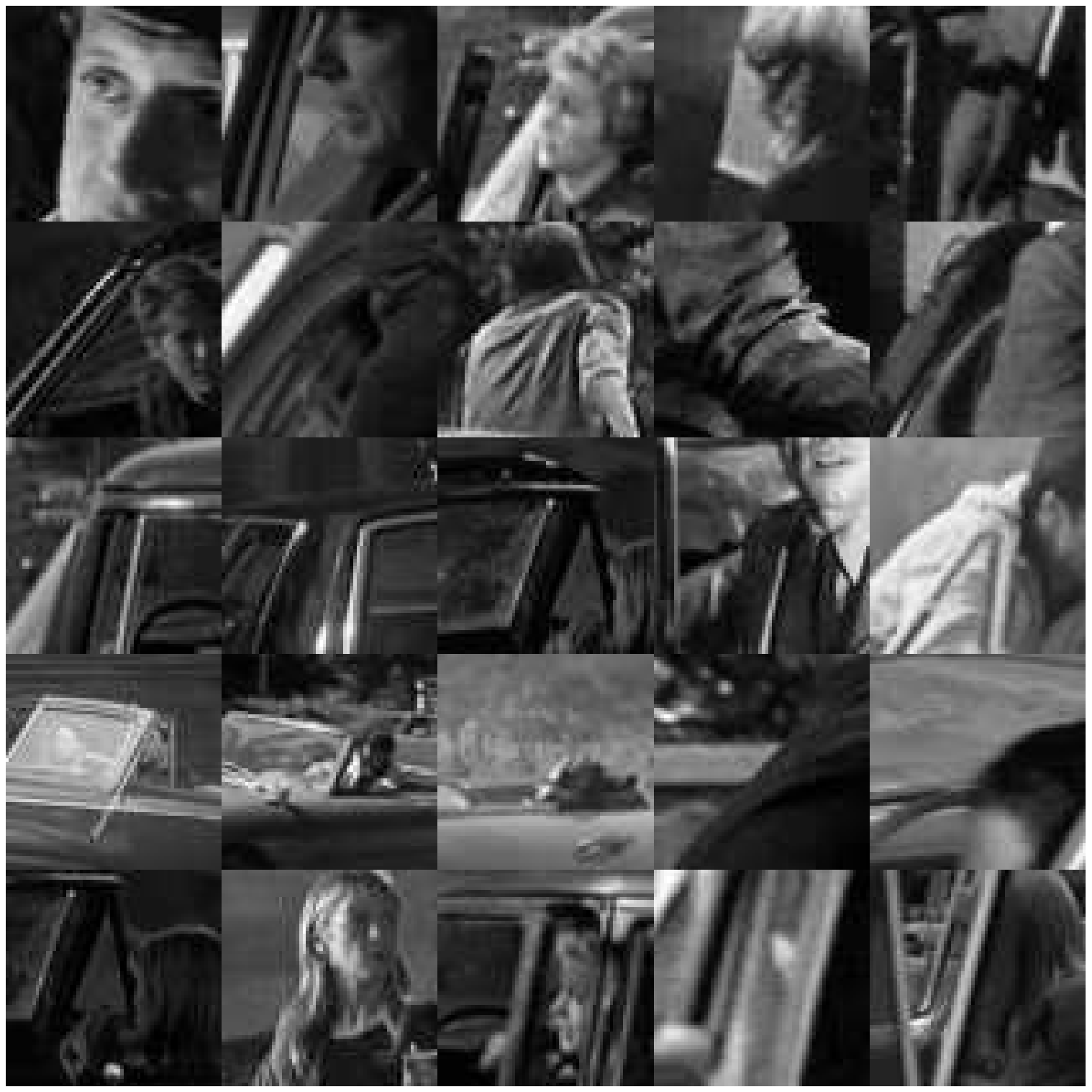}}

\vspace{0.5mm}

\\

\textbf{Eat}

\vspace{1.0mm}

&

\textbf{GetOutCar}

\vspace{1.0mm}

\\

\scalebox{0.24}{\includegraphics[viewport=3.0cm 7.0cm 18.5cm 21.7cm]{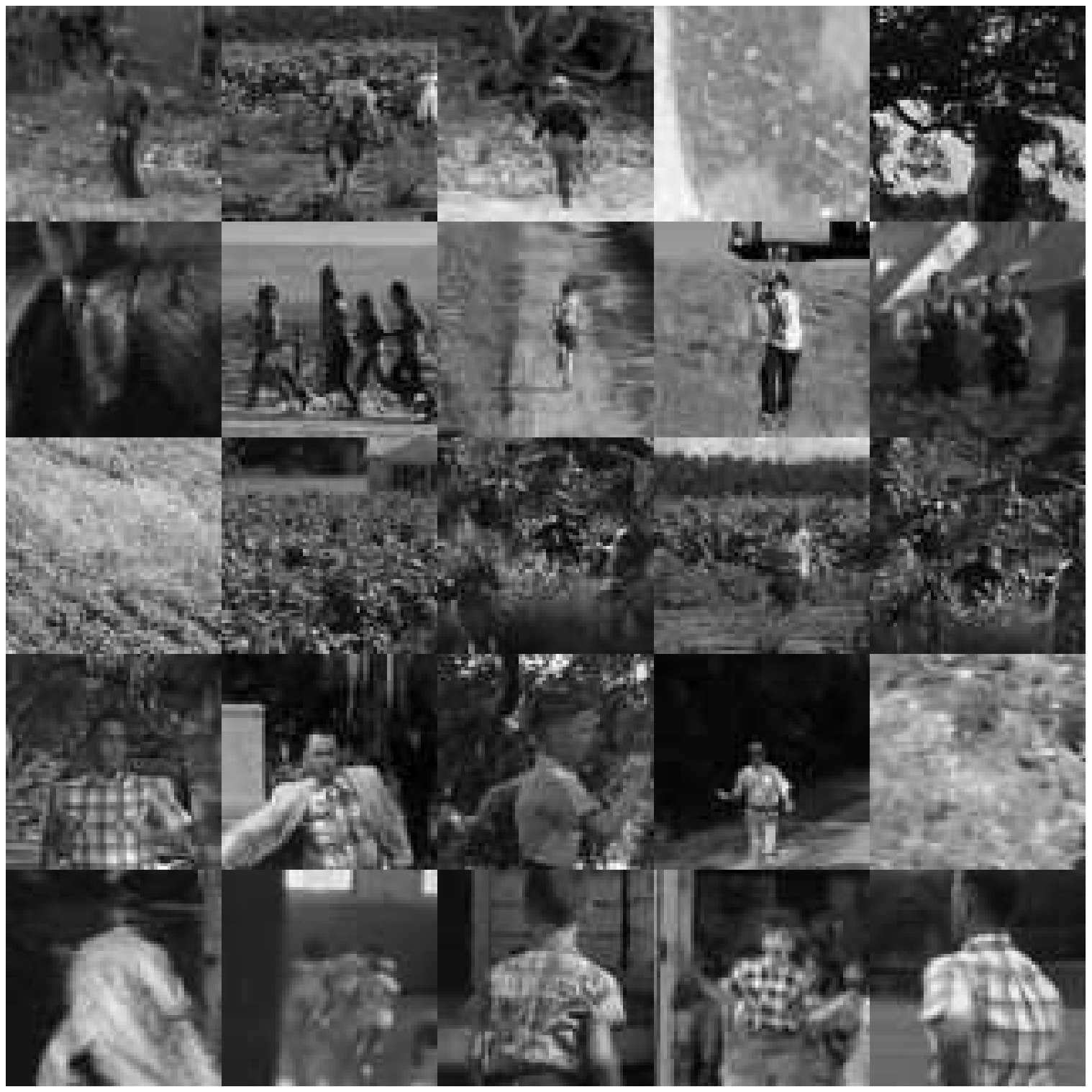}}

\vspace{0.5mm}

&

\scalebox{0.24}{\includegraphics[viewport=3.0cm 7.0cm 18.5cm 21.7cm]{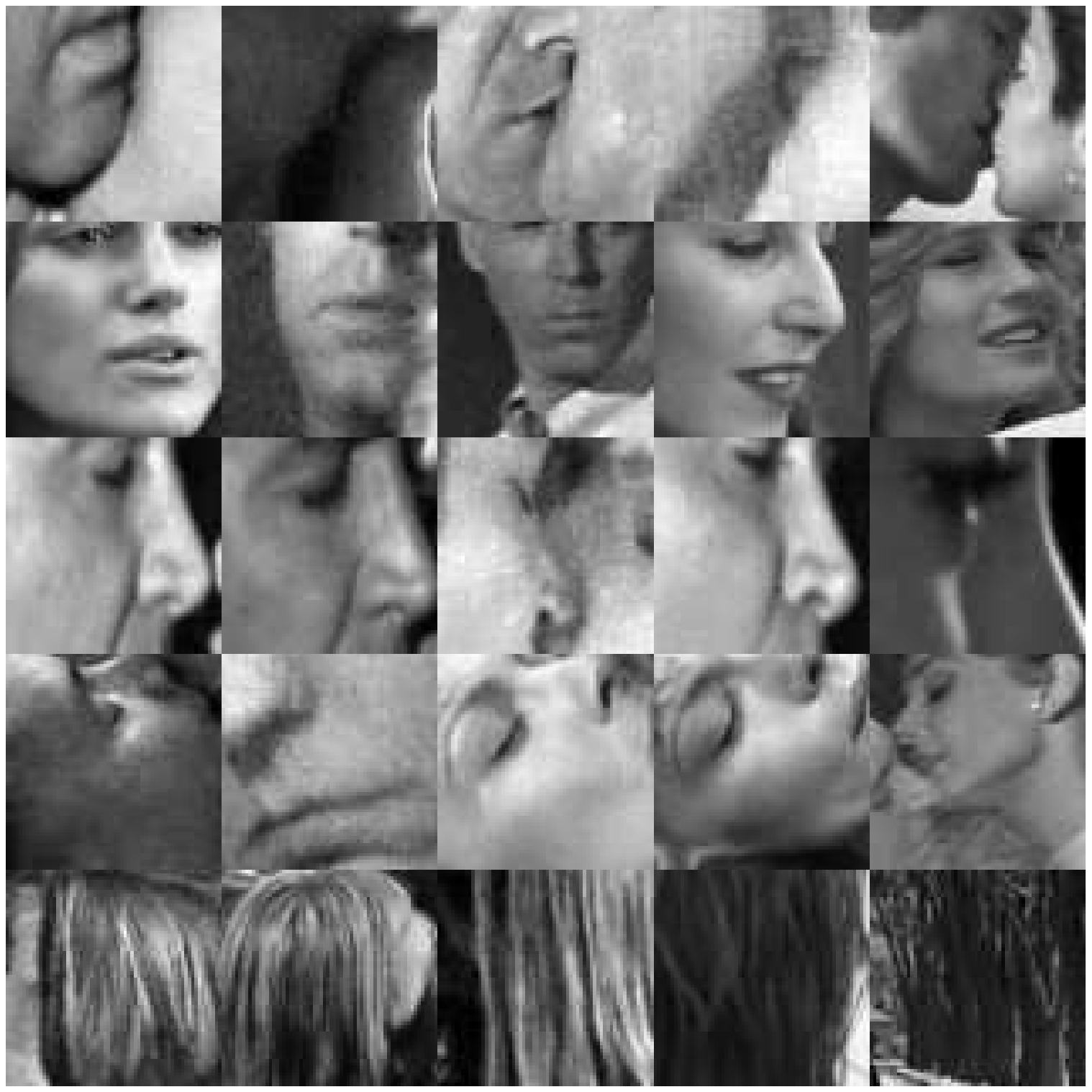}}

\vspace{0.5mm}

\\

\textbf{Run}

\vspace{1.0mm}

&

\textbf{Kiss}

\vspace{1.0mm}

\end{tabular}

\end{center}
\vspace{-5mm}
\caption{Sampled entries from visual vocabularies obtained by clustering fixated image regions in the space of HoG descriptors, for several action classes from the Hollywood-2 dataset.}\label{fig:vocabularies}
\end{figure}

\section{Semantics of Fixated Stimulus Patterns}\label{s:visual_vocabularies}

Having shown that visual attention is consistently drawn to particular locations in the stimulus, a natural question is whether the patterns at these locations are repeatable and have semantic structure. In this section, we investigate this by building vocabularies over image patches collected from the locations fixated by our subjects when viewing videos of the various actions in the Hollywood-2 data set.\\

\vspace{-2.7mm}

\noindent\textbf{Protocol}: Each patch spans $1^{\text{o}}$ away from the center of fixation to simulate the extent of high foveal accuity. We then represent each patch using HoG descriptors with a spatial grid resolution of $3 \times 3$ cells and 9 orientation bins. We cluster the resulting descriptors using k-means into 500 clusters.\footnote{We have found that using 500 clusters provides a good tradeoff between the semantic under and over-segmentation of the image patch space.}\\

\vspace{-2.7mm}

\noindent\textbf{Findings}: In Fig. \ref{fig:vocabularies} we illustrate image patches that have been assigned high probability by the mixture of Gaussians model underlying k-means. Each row contains the top 5 most probable patches from a cluster (in decreasing order of their probability). The patches in a row are restricted to come from different videos, \textit{i.e.} we remove any patches for which there is a higher probability patch coming from the same video. Note that clusters are not entirely semantically homogeneous, in part due to the limited descriptive power of HoG features (\textit{e.g.} a driving wheel is clustered together with a plate, see row 4 of the vocabulary for class \textit{Eat}).  Nevertheless, fixated patches do include semantically meaningful image regions of the scene, including actors in various poses (people eating or running or getting out of vehicles), objects being manipulated or related to the action (dishes, telephones, car doors) and, to a lesser extent, context of the surroundings (vegetation, street signs). Fixations fall almost always on objects or parts of objects and almost never on unstructured parts of the image. Our analysis also suggests that subjects generally avoid focusing on object boundaries unless an interaction is being shown (\textit{e.g.} kissing), otherwise preferring to center the object, or one of its features, onto their fovea. Overall, the vocabularies seem to capture semantically relevant aspects of the action classes. This suggests that human fixations provide a degree of object and person repeatability that could be used to boost the performance of computer-based action-recognition methods, a problem we address in the following sections.

\section{Evaluation Pipeline}\label{s:exp_setup}

For all computer action recognition models, we use the same processing pipeline, consisting of an interest point operator (computer vision or biologically derived), descriptor extraction, bag of visual words quantization and an action classifier.\\

\vspace{-2.7mm}

\noindent\textbf{Interest Point Operator:} We experiment with various interest point operators, both computer vision based (see \S\ref{s:fix_ips}) and biologically derived (see \S\ref{s:fix_ips}, \S\ref{s:end_to_end_recognition}). Each interest point operator takes as input a video and generates a set of spatio-temporal coordinates, with associated spatial and, with the exception of the 2D fixation operator presented in \S\ref{s:fix_ips}, temporal scales.\\

\vspace{-2.7mm}

\noindent\textbf{Descriptors:} We obtain features by extracting descriptors at the spatiotemporal locations returned by of our interest point operator. For this purpose, we use the spacetime generalization of the HoG descriptor as described in \cite{LaptevLindeberg2003} as well as the MBH descriptor \cite{DalalEtAl2006} computed from optical flow. We consider 7 grid configurations and extract both types of descriptors for each configuration, and end up with 14 features for classification. We use the same grid configurations in all our experiments. We have selected them from a wider candidate set based on their individual performance for action recognition. Each HoG and MBH block is normalized using a L2-Hys normalization scheme with the recommended threshold of $0.7$\cite{DalalTriggs2005}.\\

\vspace{-2.7mm}

\noindent\textbf{Visual Dictionaries/Second Order Pooling:} We cluster the resulting descriptors using k-means into visual vocabularies of $4000$ visual words.
For computational reasons, we only use $500,000$ randomly sampled descriptors as input to the clustering step, and represent each video by the $L_1$ normalized histogram of its visual words.
Alternatively, in section \S\ref{s:end_to_end_recognition} we also experiment with a different encoding scheme based on second order pooling\cite{CarreiraEtAl2012}, which provides a tradeoff between computational cost and classification accuracy. \\

\vspace{-2.7mm}

\noindent\textbf{Classifiers:} From histograms we compute kernel matrices using the RBF-$\chi^{2}$ kernel and combine the 14 resulting kernel matrices by means of a Multiple Kernel Learning (MKL) framework \cite{HariharanEtAl2010}. We train one classifier for each action label in a one-vs-all fashion. We determine the kernel scaling parameter, the SVM loss parameter $C$ and the $\sigma$ regularization parameter of the MKL framework by cross-validation.
We perform grid search over the range $[10^{-4},10^{4}] \times [10^{-4},10^{4}] \times [10^{-2},10^{-4}]$, with a multiplicative step of $10^{0.5}$. We draw $20$ folds and we run $3$ cross-validation trials at each grid point and select the parameter setting giving the best cross validation average precision to train our final action classifier.
We report the average precision on the test set for each action class, which has become the standard evaluation metric for action recognition on the Hollywood-2 data set\cite{MarszalekEtAl2009,WangEtAl2011}.

\section{Human Fixation Studies}\label{s:fix_studies}

In this section we explore the action recognition potential of computer vision operators derived from ground truth human fixations, using the evaluation pipeline introduced in \S\ref{s:exp_setup}.

\subsection{Human vs. Computer Vision Operators}\label{s:fix_ips}

The visual information found in the fixated regions could potentially aid automated recognition of human actions. One way to capture this information is to extract descriptors from these regions, which is equivalent to using them as interest points. Following this line of thought, we evaluate the degree to which human fixations are correlated with the widely used Harris spatiotemporal cornerness operator\cite{LaptevLindeberg2003}. Then, under the assumption that fixations are available at testing time, we define two interest point operators that fire at the centers of fixation, one spatially on a frame by frame basis and one at a spatio-temporal scale. We compare the performance of these two operators to that of the Harris operator for computer based action classification.\\

\vspace{-2.7mm}

\noindent\textbf{Experimental Protocol}: We start our experiment by running the spatio-temporal Harris corner detector over each video in the dataset. Assuming an angular radius of $1.5^{\text{o}}$ for the human fovea (mapped to a foveated image disc, by considering the geometry of the eye-capturing setting, \textit{e.g.} distance from screen, its size, average human eye structure), we estimate the probability that a corner will be fixated by the fraction of interest points that fall onto the fovea of at least one human observer. We then define two operators based on ground truth human fixations. The first operator generates, for each human fixation, one 2D interest point at the foveated position during the lifetime of the fixation. The second operator, generates one 3D interest point for each fixation, with the temporal scale proportional to the length of the fixation. We run the Harris operator and the two fixation operators though our classification pipeline (see \S\ref{s:exp_setup}). We determine the optimal spatial scale for the human fixation operators -- which can be interpreted as an assumed fovea radius -- by optimizing the cross-validation average precision over the range $[0.75,4]^{\text{o}}$ with a step of $1.5^{\text{o}}$.\\

\vspace{-2.7mm}

\noindent\textbf{Findings}: We find \emph{low correlation} between the locations at which classical interest point detectors fire and the human fixations.  The probability that a spatio-temporal Harris corner will be fixated by a subject is approximately $6\%$, with little variability across actions (Table \ref{t:classification_ips}a). This result is in agreement with our findings that human subjects do not generally fixate on object boundaries (\S\ref{s:visual_vocabularies}). In addition, our results show that none of our fixation-derived interest point operators improve recognition performance compared to the Harris-based interest point operator. 

\begin{table}
\centering
\scalebox{0.87}{
\begin{tabular}{|c|c|c|c|c|c|c|}
\hline
                   & \textbf{percent of human} & & \multicolumn{3}{|c|}{\textbf{recognition average precision}} \\
\cline{4-6}
 \textbf{action}   & \textbf{fixated spacetime} & & \textbf{spacetime} & \textbf{spacetime} & \textbf{spatial} \\
                   & \textbf{Harris corners}  & &  \textbf{Harris}   & \textbf{fixations} & \textbf{fixations} \\
                   & (a)                      & & (b)                & (c)                &     (d)            \\
\hline
AnswerPhone        & 6.2\%                    & & \textbf{16.4\%}    &         16.0\%     &         14.6\%     \\
\hline
DriveCar           & 5.8\%                    & &         85.4\%     &         79.4\%     & \textbf{85.9\%}    \\
\hline
Eat                & 6.4\%                    & & \textbf{59.1\%}    &         54.1\%     &         55.8\%     \\
\hline
FightPerson        & 4.6\%                    & &         71.1\%     &         66.5\%     & \textbf{73.9\%}    \\
\hline
GetOutCar          & 6.1\%                    & & \textbf{36.1\%}    &         31.7\%     &         35.4\%     \\
\hline
HandShake          & 6.3\%                    & & \textbf{18.2\%}    &         14.9\%     &         17.5\%     \\
\hline
HugPerson          & 4.6\%                    & &         33.8\%     & \textbf{35.1\%}    &         28.2\%     \\
\hline
Kiss               & 4.8\%                    & &         58.3\%     &         61.3\%     & \textbf{64.3\%}    \\
\hline
Run                & 6.0\%                    & &         73.2\%     & \textbf{78.5\%}    &         78.0\%     \\
\hline
SitDown            & 6.2\%                    & & \textbf{54.0\%}    &         41.9\%     &         51.8\%     \\
\hline
SitUp              & 6.3\%                    & & \textbf{26.1\%}    &         16.3\%     &         22.7\%     \\
\hline
StandUp            & 6.0\%                    & & \textbf{57.0\%}    &         50.4\%     &         46.3\%     \\
\hline
\hline
\textbf{Mean}      & 5.8\%                    & & \textbf{49.1\%}    &         45.5\%     &         47.9\%     \\
\hline
\end{tabular}
}
%\end{center}
%\captionsetup{justification=raggedright, singlelinecheck=false}
\caption{Harris Spacetime Corners vs. Fixations. Only a small percentage of spatio-temporal Harris corners are fixated by subjects across the videos in each action class (a). Classification average precision with various interest point operators: (b) spacetime Harris corner detectors, (c) human fixations, one interest point per fixated frame, and (d) human fixations, one interest point per fixation.  Using fixations as interest point operators does not lead to improvements in recognition performance, compared to the spatio-temporal Harris operator.}\label{t:classification_ips}
\end{table}

\subsection{Impact of Human Saliency Maps for Computer Visual Action Recognition}\label{s:saliency}

Although our findings suggest that the entire foveated area is not informative, this does not rule out the hypothesis that relevant information for action recognition might lie in a subregion of this area. Research of human attention suggests that humans are capable of attending to only a subregion of the fovea at a time, to which the mental processing resources are directed, the so called \emph{covert attention}\cite{LandTatler2009}. Along these lines, we design an experiment in which we generate finer-scaled interest points in the area fixated by the subjects. Given enough samples, we expect to also represent the area to which the covert attention of a human subject was directed at any particular moment through the fixation. To drive this sampling process, we derive a saliency map from human fixations. The map estimates the probability for each spatio-temporal location in the video to be foveated by a human subject. We then define an interest point operator that randomly samples spatio-temporal locations from this probability distribution and compare its performance for action recognition with two baselines: the spatio-temporal Harris operator\cite{LaptevLindeberg2003} and an interest point operator that fires randomly with equal probability over the entire spatio-temporal volume of the video. If subsets of the foveated regions are indeed informative, we expect our saliency-based interest point operator to have superior performance to both baselines.\\

\vspace{-2.7mm}

\noindent\textbf{Experimental Protocol}: Let $I_f$ denote a frame belonging to some video $v$. We label each pixel $\mathbf{x} \in I_f$ with the number of human fixations centered at that pixel. We then blur the resulting label image by convolution with an isotropic Gaussian filter having a standard deviation $\sigma$ equal to the assumed radius of the human fovea. We obtain a map $m_{f}$ over the frame $f$. The probabilitity for each spatio temporal pixel to be fixated is $p_{\text{fix}}(\mathbf{x},f)=\frac{1}{T}\cdot \frac{m_f(\mathbf{x})}
	{
		\sum_{\mathbf{x} \in I_f}
		m_f(\mathbf{x})
	}$ where $T$ is the total number of frames in the video. To obtain the saliency distribution, we regularize this probability to account for observation sparsity, by adding a uniform distribution over the video frame, weighted by a parameter $\alpha$, $p_{\text{sal}}=(1-\alpha)p_{\text{fix}}+\alpha p_{\text{unif}}$. 
%\begin{equation}\label{eq:saliency}
%p_{\text{saliency}}(\mathbf{x},f)=\alpha \cdot p_{\text{uniform}}(\mathbf{x},f)+(1-\alpha)\cdot p_{\text{fixation}}(\mathbf{x},f)
%\end{equation}
%\noindent where $p_{\text{uniform}}$ is the uniform probability distribution over the video frame and $\alpha$ a regularization parameter.
We now define an interest point operator that randomly samples spatio-temporal locations from the ground truth probability distribution $p_{\text{sal}}$, both at training and testing time, and associates them with random spatio-temporal scales uniformly distributed in the range $[2,8]$.
%\footnote{The biological motivation for this choice is that the area over which covert attention integrates information is not well known.}
We train classifiers for each action by feeding the output of this operator through our pipeline (see \S\ref{s:exp_setup}). By doing so, we build vocabularies from descriptors sampled from saliency maps derived from ground truth human fixations. We determine the optimal values for the $\alpha$ regularization parameter and the fovea radius $\sigma$ by cross-validation. We also run two baselines: the spatio-temporal Harris operator (see \S \ref{s:fix_ips}) and the operator that samples locations from the uniform probability distribution, which can be obtained by setting $\alpha=1$ in $p_{\text{sal}}$. In order to make the comparison meaningful, we set the number of interest points sampled by our saliency-based and the uniform random operators in each frame to match the firing rate of the Harris corner detector.\\
%($\approx 55$ interest points per frame).

\vspace{-2.7mm}

\noindent\textbf{Findings}: We find that ground truth saliency sampling (Table \ref{t:classification}e) outperforms both the Harris and the uniform sampling operators significantly, at equal interest point sparsity rates. Our results indicate that saliency maps encoding only the weak surface structure of fixations (no time ordering is used), can be used to boost the accuracy of contemporary methods and descriptors used for computer action recognition. Up to this point, we have relied on the availability of ground truth saliency maps at test time. A natual question is whether it is possible to reliably predict saliency maps, to a degree that still preserves the benefits of action classification accuracy. This will be the focus of the next section. 

\begin{table*}
\caption{Action Recognition Performance on the \textbf{Hollywood-2} Data Set}\label{t:classification}
\begin{center}
\scalebox{1.00}{
\begin{tabular}{|c|c|c|c|c|c|c|c|c|c|c|}
\hline
                   & \multicolumn{5}{|c|}{\textbf{interest points}}                                                          & &                       & \multicolumn{3}{|c|}{\textbf{trajectories + interest points}}
\\
\cline{2-6}\cline{9-11}
                   &                    &                    & \textbf{central}   & \textbf{predicted} & \textbf{ground truth} & &                       &                   & \textbf{predicted} & \textbf{ground truth} \\
\textbf{action}    & \textbf{Harris}    & \textbf{uniform}   & \textbf{bias}      & \textbf{saliency}  & \textbf{saliency}     & & \textbf{trajectories} & \textbf{uniform}   & \textbf{saliency}  & \textbf{saliency}     \\
                   & \textbf{corners}   & \textbf{sampling}  & \textbf{sampling}  & \textbf{sampling}  & \textbf{sampling}     & & \textbf{only}         & \textbf{sampling}    & \textbf{sampling}  & \textbf{sampling}     \\    
                   & (a)                & (b)                & (c)                & (d)                & (e)                   & & (f)                   & (g)               & (h)                & (i)                   \\
\hline
AnswerPhone        & 16.4\%           &        21.3\%      & 23.3\%             &        23.7\%      & \textbf{28.1\%}       & & \textbf{32.6\%}       &         24.5\%    &         25.0\%     &         32.5\%  \\
\hline
DriveCar           & 85.4\%           &        92.2\%      & 92.4\%             &        92.8\%      & \textbf{57.9\%}       & &         88.0\%        &         93.6\%    &         93.6\%     & \textbf{96.2\%} \\
\hline
Eat                & 59.1\%           &        59.8\%      & 58.6\%             & \textbf{70.0\%}    &         67.3\%        & &         65.2\%        &         69.8\%    & \textbf{75.0\%}    &         73.6\%  \\
\hline
FightPerson        & 71.1\%           &        74.3\%      & 76.3\%             &         76.1\%     & \textbf{80.6\%}       & &         81.4\%        &         79.2\%    &         78.7\%     & \textbf{83.0\%} \\
\hline
GetOutCar          & 36.1\%           &        47.4\%      & 49.6\%             &         54.9\%     & \textbf{55.1\%}       & &         52.7\%        &         55.2\%    & \textbf{60.7\%}    &         59.3\%  \\
\hline
HandShake          & 18.2\%           &        25.7\%      & 26.5\%             & \textbf{27.9\%}    &         27.6\%        & & \textbf{29.6\%}       &         29.3\%    &         28.3\%     &         26.6\%  \\
\hline
HugPerson          & 33.8\%           &        33.3\%      & 34.6\%             & \textbf{39.5\%}    &         37.8\%        & & \textbf{54.2\%}       &         44.7\%    &         45.3\%     &         46.1\%  \\
\hline
Kiss               & 58.3\%           &        61.2\%      & 62.1\%             &         61.3\%     & \textbf{66.4\%}       & &         65.8\%        &         66.2\%    &         66.4\%     & \textbf{69.5\%} \\
\hline
Run                & 73.2\%           &        76.0\%      & 77.8\%             &         82.2\%     & \textbf{85.7\%}       & &         82.1\%        &         82.1\%    &         84.2\%     & \textbf{87.2\%} \\
\hline
SitDown            & 54.0\%           &        59.3\%      & 62.1\%             & \textbf{69.0\%}    &         62.5\%        & &         62.5\%        &         67.2\%    & \textbf{70.4\%}    &         68.1\%  \\
\hline
SitUp              & 26.1\%           &        20.7\%      & 20.9\%             &         29.7\%     & \textbf{30.7\%}       & &         20.0\%        &         23.8\%    & \textbf{34.1\%}    &         32.9\%  \\
\hline
StandUp            & 57.0\%           &        59.8\%      & 61.3\%             & \textbf{63.9\%}    &         58.2\%        & &         65.2\%        &         64.9\%    & \textbf{69.5\%}    &         66.0\%  \\
\hline
\hline
	\textbf{Mean}      & 49.1\%           &        52.6\%      & 53.7\%             &         57.6\%     &         57.9\%        & & 58.3\%                &         58.4\%    &          61.0\%    & \textbf{61.7\%} \\
\hline
\end{tabular}
}
\end{center}
\textbf{Columns a-e}: Action recognition performance on the \textbf{Hollywood-2} data set when interest points are sampled randomly across the spatio-temporal volumes of the videos
from various distributions (b-e), with the Harris corner detector as baseline (a). Average precision is shown for the uniform (b), central bias (c) and ground truth (e) distributions, and for the output (d) of our HoG-MBH detector. All pipelines use the same number of interest points per frame as generated by the Harris spatio-temporal corner detector (a).
\textbf{Columns f-i}: Significant improvement over the state of the art\cite{WangEtAl2011} (f) can be achieved by augmenting the method with the channels derived from interest points sampled from the predicted saliency map (g) and ground truth saliency (h),
but not when using the classical uniform sampling scheme (e).
\end{table*}

\definecolor{myred}{rgb}{0.6,0,0}

\begin{table*}
\caption{Action Recognition Performance on the \textbf{UCF Sports Actions} Data Set}\label{t:classification_ucfsa}
\begin{minipage}{0.43\linewidth}
\begin{center}
\scalebox{0.9}{
\begin{tabular}{|c|c|c|}
\hline \textbf{method}  & \textbf{distribution} & \textbf{accuracy}              \\
%\cline{3-4}
\hline                                   & Harris corners (a)                  & 86.6\% \\
\cline{2-3}                              & uniform sampling (b)                & 84.6\% \\                  
\cline{2-3} interest points              & central bias sampling (c)           & 84.8\% \\                     
\cline{2-3}                              & predicted saliency sampling (d)     & 91.3\% \\                     
\cline{2-3}                              & ground truth saliency sampling (e)  & 90.9\% \\                     
\hline\multicolumn{2}{|c|}{trajectories\cite{WangEtAl2011} only (f)}           & 88.2\% \\                     
\hline trajectories +                    & \multirow{2}{*}{predicted saliency sampling (h)}  & \textbf{\multirow{2}{*}{91.5\%}}   \\ 
       interest points                   &                                     &                                                  \\
\hline
\end{tabular}

%\begin{tabular}{|c|c|c|c|}
%\hline \multirow{2}{*}{\textbf{method}}  & \multirow{2}{*}{\textbf{distribution}} & \multicolumn{2}{|c|}{\textbf{accuracy}}              \\
%\cline{3-4}
%                                         &                                     & \textbf{mean}              & \textbf{stdev}             \\
%\hline                                   & Harris corners (a)                  & 86.6\%                     & 0.00\%                     \\
%\cline{2-4}                              & uniform sampling (b)                & 84.6\%                     & 0.35\%                     \\
%\cline{2-4} interest points              & central bias sampling (c)           & 84.8\%                     & 0.35\%                     \\
%\cline{2-4}                              & predicted saliency sampling (d)     & 91.3\%                     & 0.32\%                     \\
%\cline{2-4}                              & ground truth saliency sampling (e)  & 90.9\%                     & 0.39\%                     \\
%\hline\multicolumn{2}{|c|}{trajectories\cite{WangEtAl2011} only (f)}           & 88.2\%                     & 0.00\%                     \\
%\hline trajectories +                    & \multirow{2}{*}{predicted saliency sampling (h)}  & \textbf{\multirow{2}{*}{91.5\%}}    & \multirow{2}{*}{0.29\%}    \\
%       interest points                   &                                     &                                                   &                            \\
%\hline
%\end{tabular}
}
\end{center}
\end{minipage}
\begin{minipage}{0.50\linewidth}
\begin{tabular}{cc}
\scalebox{0.31}{\includegraphics[viewport=5.2cm 7.5cm 20.0cm 19.0cm]{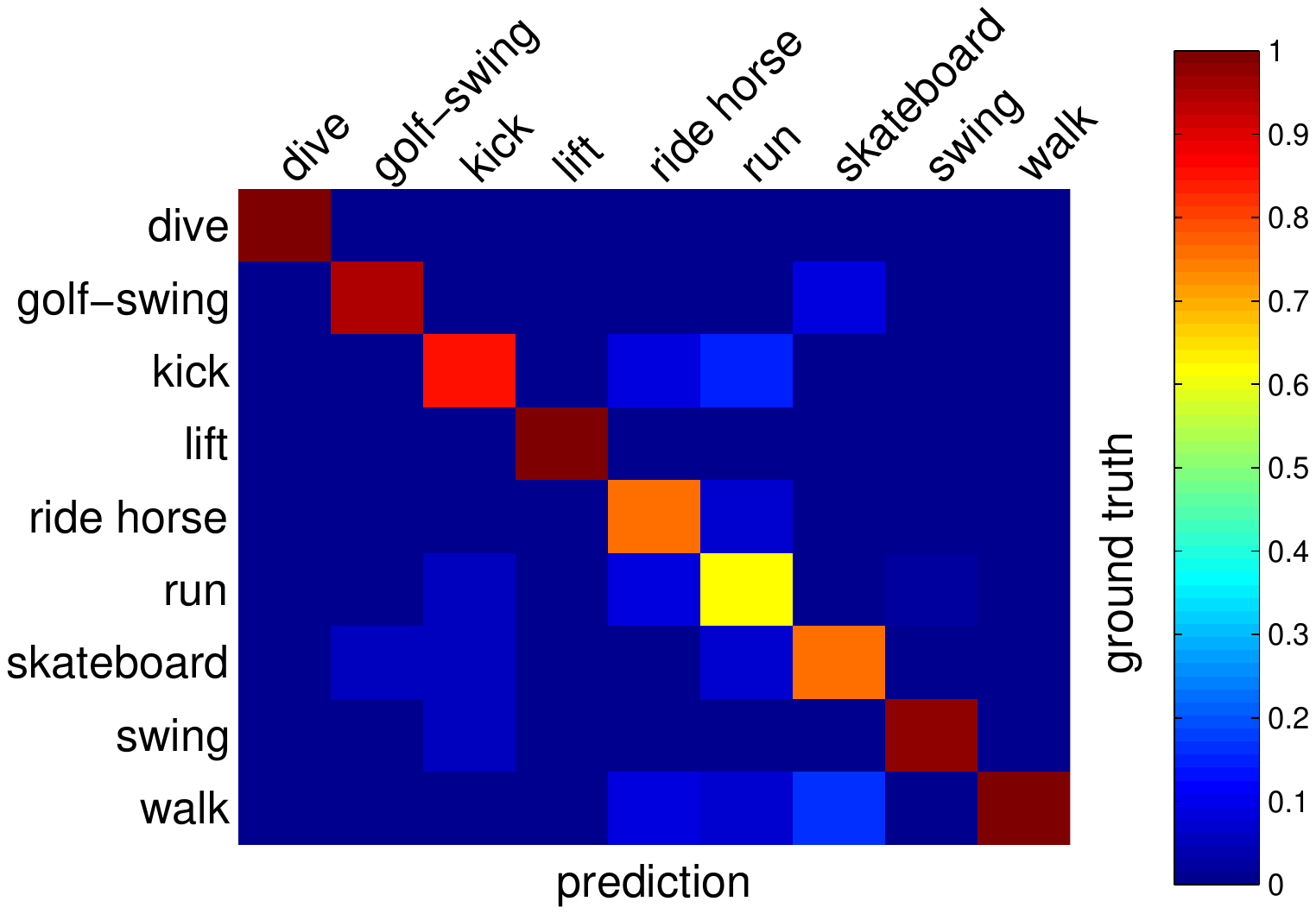}}
&
\scalebox{0.31}{\includegraphics[viewport=5.2cm 7.5cm 20.0cm 19.0cm]{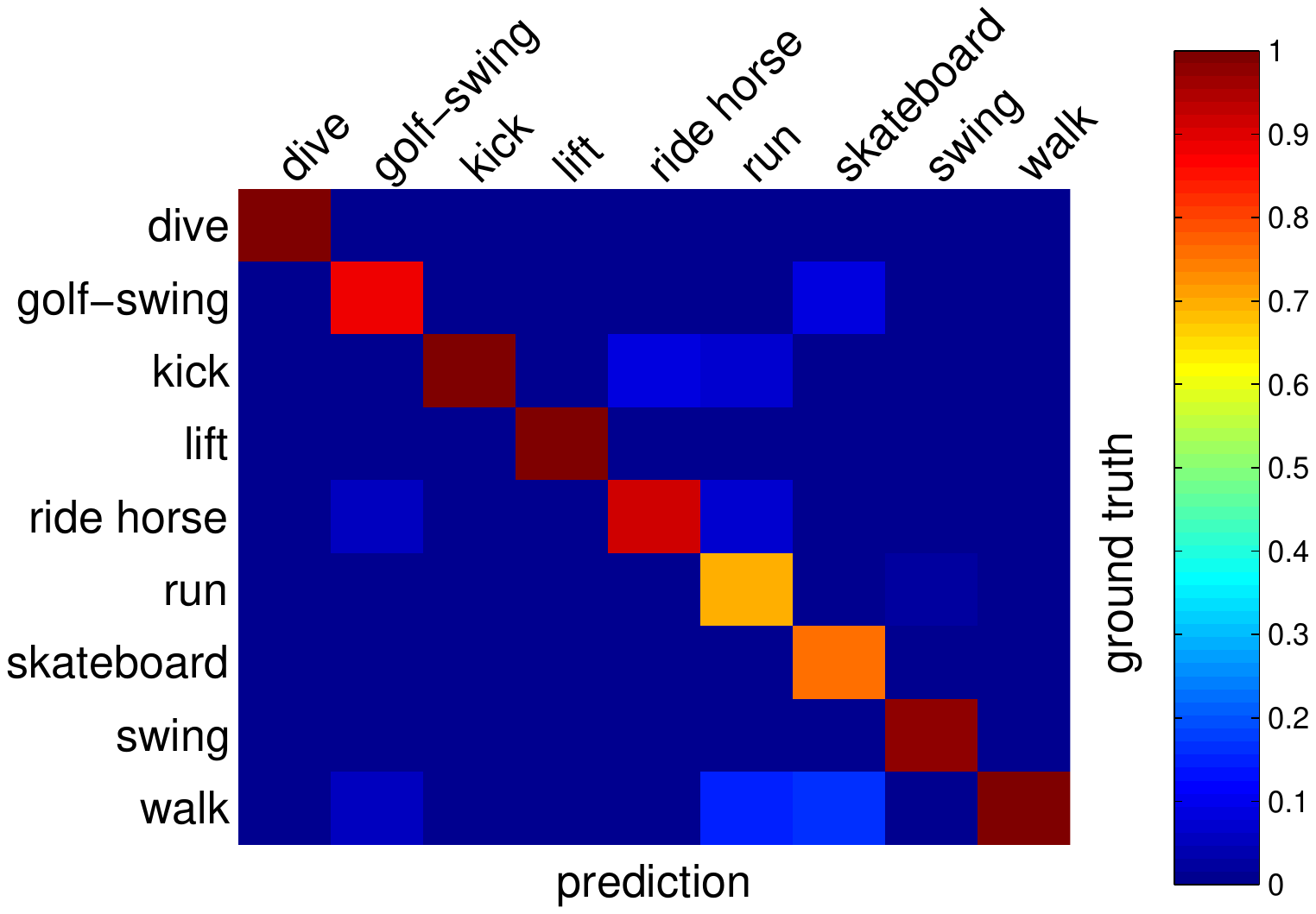}}
\\
trajectories
&
predicted saliency map sampling
\\
&

\end{tabular}
\end{minipage}

\textbf{Left:} Performance comparison among several classification methods (see table \ref{t:classification} for description) on the UCF Sports Dataset.
%Average multiple-class prediction accuracy is reported by lea%ve-one-out cross-validation fashion.
\textbf{Right:} Confusion matrices obtained using dense trajectories \cite{WangEtAl2011} and interest points sampled sparsely from saliency maps predicted by our HoG-MBH detector.
\end{table*}

%\multirow{5}{*}{interest points}

\section{Saliency Map Prediction}\label{s:saliency_predict}

Motivated by the findings presented in the previous section, we now show that we can effectively predict saliency maps. We start by introducing two evaluation measures for saliency prediction. The first is the area-under-the-curve (AUC), which is widely used in the human vision community. The second measure is inspired by our application of saliency maps to action recognition. In the pipeline we proposed in \S\ref{s:saliency}, ground truth saliency maps drive the random sampling process of our interest point operator. We will use the spatial Kullback-Leibler (KL) divergence measure to compare the predicted and the ground truth saliencies. We also propose and study several features for saliency map prediction, both static and motion based. Our analysis includes features derived directly from low, mid and high level image information. In addition, we train a HoG-MBH detector that fires preferentially at fixated locations, using the vast amount of eye movement data available in the dataset. We evaluate all these features and their combinations on our dataset, and find that our detector gives the best performance under the KL divergence measure.\\

\vspace{-2.7mm}

\noindent\textbf{Saliency Map Comparison Measures}: The most commonly used measure for evaluating saliency maps in the image domain\cite{JuddEtAl2009,EhingerEtAl2009,JuddEtAl2012}, the AUC measure, interprets saliency maps as predictors for separating fixated pixels from the rest. The ROC curve is computed for each image and the average area the area under the curve over the whole set of testing images gives the final score. This measure emphasizes the capacity of a saliency map to rank pixels higher when they are fixated then when they are not. This does not imply, however, that the normalized probability distribution associated with the saliency map is close to the ground truth saliency map for the image. A better suited way to compare probability distributions is the spatial Kullback-Leibler (KL) divergence, which we propose as our second evaluation measure, defined as:
%$D_{\text{KL}}(p,s)=\sum_{x \in I} p(x) \log{\frac{p(x)}{s(x)}}$, %
\begin{equation}\label{eq:kl}
D_{\text{KL}}(p,s)=\sum_{x \in I} p(x) \log{\frac{p(x)}{s(x)}}
\end{equation}
\noindent 
where $p(x)$ is the value of the normalized saliency prediction at pixel $x$, and $s$ is the ground truth saliency map. A small value of this metric implies that random samples drawn from the predicted saliency map $p$ are likely to approximate well the ground truth saliency $s$.\\

\vspace{-2.7mm}

\noindent\textbf{Saliency Predictors}: Having established evaluation criteria, we now run several saliency map predictors on our dataset, which we describe below.\\

\vspace{-2.7mm}

\noindent\underline{\textit{Baselines}}: We also provide three baselines for saliency map comparison. The first one is the uniform saliency map, that assigns the same fixation probability to each pixel of the video frame. Second, we consider the center bias (CB) feature, which assigns each pixel with the distance to the center of the frame, regardless of its visual contents. This feature can capture both the tendency of human subjects to fixate near the center of the screen and the preference of the photographer to center objects into the field of view. At the other end of the spectrum lies the human saliency baseline, which derives a saliency map from half of our human subjects and is evaluated with respect to fixations of the remaining ones. \\ 
%, as opposed to the entire subject group.\\

\vspace{-2.7mm}

\noindent\underline{\textit{Static Features (SF)}}: We also include features used by the human vision community for saliency prediction in the image domain\cite{JuddEtAl2009}, which can be classified into three categories: low, mid and high-level. The four low level features used are color information, steerable pyramid subbands, the feature maps used as input by Itti\&Koch's model\cite{IttiKoch2000} and the output of the saliency model described by Oliva and Torralba \cite{OlivaTorralba2001} and Rosenholtz\cite{Rosenholtz1999}. We run a Horizon detector\cite{OlivaTorralba2001} as our mid-level feature. Object detectors are used as high level features, which comprise faces\cite{ViolaJones2001}, persons and cars\cite{FelzenswalbEtAl2008}.\\

\vspace{-2.7mm}

\noindent\underline{\textit{Motion Features (MF)}}: We augment our set of predictors with five novel (in the context of saliency prediction) feature maps, derived from motion or space-time information.\\

\vspace{-2.7mm}

\noindent\textit{Flow:} We extract optical flow from each frame using a state of the art method\cite{SunEtAl2010} and compute the magnitude of the flow at each location. Using this feature, we wish to investigate whether regions with significant optical changes attract human gaze.\\

\vspace{-2.7mm}

\noindent\textit{Pb with flow:} We run the PB edge detector\cite{ArbelaezEtAl2011} with both image intensity and the flow field as inputs. This detector fires both at intensity and motion boundaries.\\ 
%, where both image and flow discontinuities arise.

\vspace{-2.7mm}

\noindent\textit{Flow bimodality:} We wish to investigate how often people fixate on motion edges, where the flow field typically has a bimodal distribution. To do this,
for a neighbourhood centered at a given pixel $\mathbf{x}$, we run K-Means, first with 1 and then with 2 modes, obtaining sum-of-squared-error values of $s_1(\mathbf{x})$ and $s_2(\mathbf{x})$
respectively. We weight the distance between the centers of the two modes by a factor inversely proportional to $\exp\left(1-\frac{s_1(\mathbf{x})}{s_2({\mathbf{x})}}\right)$, to enforce a high response at positions where the optical flow distribution is strongly bimodal and its mode centers are far apart from each other.\\

\vspace{-2.7mm}

%The final feature is computed as:

%\begin{equation}
%f_{\text{fb}}(\mathbf{x})=\frac{2}{1+\exp\left(1-\frac{s_1(\mathbf{x})}{s_2({\mathbf{x})}}\right)} \cdot \left\|\mathbf{c}_1 - \mathbf{c}_2\right\|
%\end{equation}

%\noindent where $\mathbf{c}_1$ and $\mathbf{c}_2$ are the cluster centers. This metric is high when there is a significant improvement in the sum-of-squared-errors
%metric when two cluster centers are used instead of just one, and, at the same time, the distance between their centers is high.

\noindent\textit{Harris:} This feature encodes the spatio-temporal Harris cornerness measure as defined in \cite{Laptev2005}.\\

\vspace{-2.7mm}

\noindent\textit{HoG-MBH detector:} The saliency models we have considered so far access higher level image structure by means of pre-trained object detectors. This approach does not prove effective on our dataset, due to the high variability in pose and illumination. On the other hand, our dataset provides a rich set of human fixations. We observe that fixated image regions are often semantically meaningful, sometimes corresponding to objects or object parts. Inspired by this insight, we aim to exploit the structure present at these locations and train a detector for human fixations. Our detector uses both static (HoG) and motion (MBH) descriptors centered at fixations. We run our detector in a sliding
window fashion across the entire video and obtain a saliency map.\\
%We build a saliency map by computing the confidence of our detector when run over the entire video in sliding window fashion.

\vspace{-2.7mm}

\noindent\underline{\textit{Feature combinations}:} We linearly combine various subsets of our feature maps for better saliency prediction. We investigate the predictive power of static features and motion features alone and in combination, with and without central bias. \\

\vspace{-2.7mm}

\noindent\textbf{Experimental Protocol:} We use $10^6$ examples to train our detector, half of which are positive and half of which are negative. At each of these locations, we extract spatio-temporal HoG and MBH descriptors. We opt for 3 grid configurations, namely 1x1x1, 2x2x1 and 3x3x1 cells. We have experimented with higher temporal grid resolutions, but found only modest improvements in detector performance at a high increase in computational cost. We concatenate all 6 descriptors and lift the resulting vector into a higher dimensional space by employing an order 3 $\chi^2$ kernel approximation using the approach of \cite{VedaldiZisserman2011}. We train an SVM using the LibLinear package\cite{LibLinear} to obtain our HoG-MBH detector.

For combining feature maps, we train a linear predictor on 500 randomly selected frames from the Hollywood-2 training set, using our fixation annotations. We exclude the first 8 frames of each video from the sampling process, in order to avoid the effects of the initial central fixation in our data collection setup. We also randomly select 250 and 500 frames for validation and testing, respectively. To avoid correlations, the video sets used to sample training, validation and testing frames are disjoint.\\

\vspace{-2.7mm}

\noindent\textbf{Findings:} When evaluated at $10^6$ random locations, half of which were fixated by the subjects and half not, the average precision of our detector is 76.7\%, when both MBH and HoG descriptors are used. HoG descriptors used in isolation perform better (73.4\% average precision) than MBH descriptors alone (70.5\%), indicating that motion structure contributes less to detector performance than does static image structure. There is, however, significant advantage in combining both sources of information.

When evaluated under the AUC metric, combining predictors always improves performance (Table \ref{t:judd_measures}). As a general trend, low-level features are better predictors than high level ones. The low level motion features (flow, pb edges with flow, flow bimodality, Harris cornerness), provide similar performance to static low-level features. Our HoG-MBH detector is comparable to the best static feature, the Horizon detector, under the AUC metric.

Interestingly, when evaluated according to KL divergence, the ranking of the saliency maps changes: the HoG-MBH detector performs best and the only other predictor that significantly outperforms central bias is the horizon detector. Under this metric, combining features does not always improve performance, as the linear combination method of \cite{JuddEtAl2009} optimizes pixel-level classification accuracy, and as such is not able to account for the inherent competition that takes place among these predictions due to image-level normalization. We conclude by noticing that fusing our predicted maps as well as our static and dynamic features gives the highest results under AUC metrics. Moreover, the HoG-MBH detector, trained using our eye movement data is the best predictor of visual saliency from our candidate set, under the probabilistic measure of matching the spatial distribution of human fixations. 

\begin{table}
\caption{Evaluation of Individual Feature Maps and Combinations for Human Saliency Prediction.} \label{t:judd_measures}
\begin{minipage}{0.497\linewidth}
\begin{center}
\scalebox{0.81}{
\begin{tabular}{|c|c|c|}
\hline
\multicolumn{3}{|c|}{\textbf{baselines}}
\\
\hline
\textbf{feature}               & \textbf{AUC}        & \textbf{KL}
\\
                               & (a)                 & (b)
\\
\hline
uniform baseline               & $0.500$             & $18.63$
\\
\hline
central bias (CB)              & $0.840$             & $15.93$
\\
\hline
human                          & $0.936$             & $10.12$
\\
\hline
\hline
\multicolumn{3}{|c|}{\textbf{static features (SF)}}
\\
\hline
color features \cite{JuddEtAl2009}         & $0.644$             & $17.90$
\\
\hline
subbands \cite{SimocelliFreeman1995}       & $0.634$             & $17.75$
\\
\hline
Itti\&Koch channels \cite{IttiKoch2000}    & $0.598$             & $16.98$
\\
\hline
saliency map \cite{OlivaTorralba2001}      & $0.702$             & $17.17$
\\
\hline
horizon detector \cite{OlivaTorralba2001}  & $0.741$             & $15.45$
\\
\hline
face detector \cite{ViolaJones2001}        & $0.579$             & $16.43$
\\
\hline
car detector \cite{FelzenswalbEtAl2008}    & $0.500$             & $18.40$
\\
\hline
person detector \cite{FelzenswalbEtAl2008} & $0.566$             & $17.13$
\\
\hline
\end{tabular}
}
\end{center}
\end{minipage}
\begin{minipage}{0.495\linewidth}
\begin{center}
\scalebox{0.81}{
\begin{tabular}{|c|c|c|}
\hline
\multicolumn{3}{|c|}{\textbf{our motion features (MF)}}
\\
\hline
\textbf{feature}               & \textbf{AUC}        & \textbf{KL}
\\
                               & (a)                 & (b)
\\
\hline
flow magniture                             & $0.626$             & $18.57$
\\
\hline
pb edges with  flow                        & $0.582$             & $17.74$
\\
\hline
flow bimodality                            & $0.637$             & $17.63$
\\
\hline
Harris cornerness                          & $0.619$             & $17.21$
\\
\hline
HOG-MBH detector                           & $0.743$             & $\mathbf{14.95}$
\\
\hline
\hline
\multicolumn{3}{|c|}{\textbf{feature combinations}}
\\
\hline
SF \cite{JuddEtAl2009}                     & $0.789$             & $16.16$
\\
\hline
SF + CB \cite{JuddEtAl2009}                & $0.861$             & $15.96$
\\
\hline
MF                                         & $0.762$             & $15.62$
\\
\hline
MF + CB                                    & $0.830$             & $15.97$
\\
\hline
SF + MF                                    & $0.812$             & $15.94$
\\
\hline
SF + MF + CB                               & $\mathbf{0.871}$    & $15.89$
\\
\hline
\end{tabular}
}
\end{center}
\end{minipage}

\vspace{2mm}

 We show area under the curve (AUC) and KL divergence. AUC and KL induce different saliency map rankings, but for visual recognition measures that emphasize spatial localization are essential (see also \tabref{t:classification} for action recognition results and \figref{fig:features} for  illustration).
\end{table}

\begin{figure*}
\begin{center}
\scalebox{0.81}{
\begin{tabular}{cccccc}
\scalebox{0.156}{\includegraphics[viewport=0cm 0.0cm 21.3cm 11.6cm]{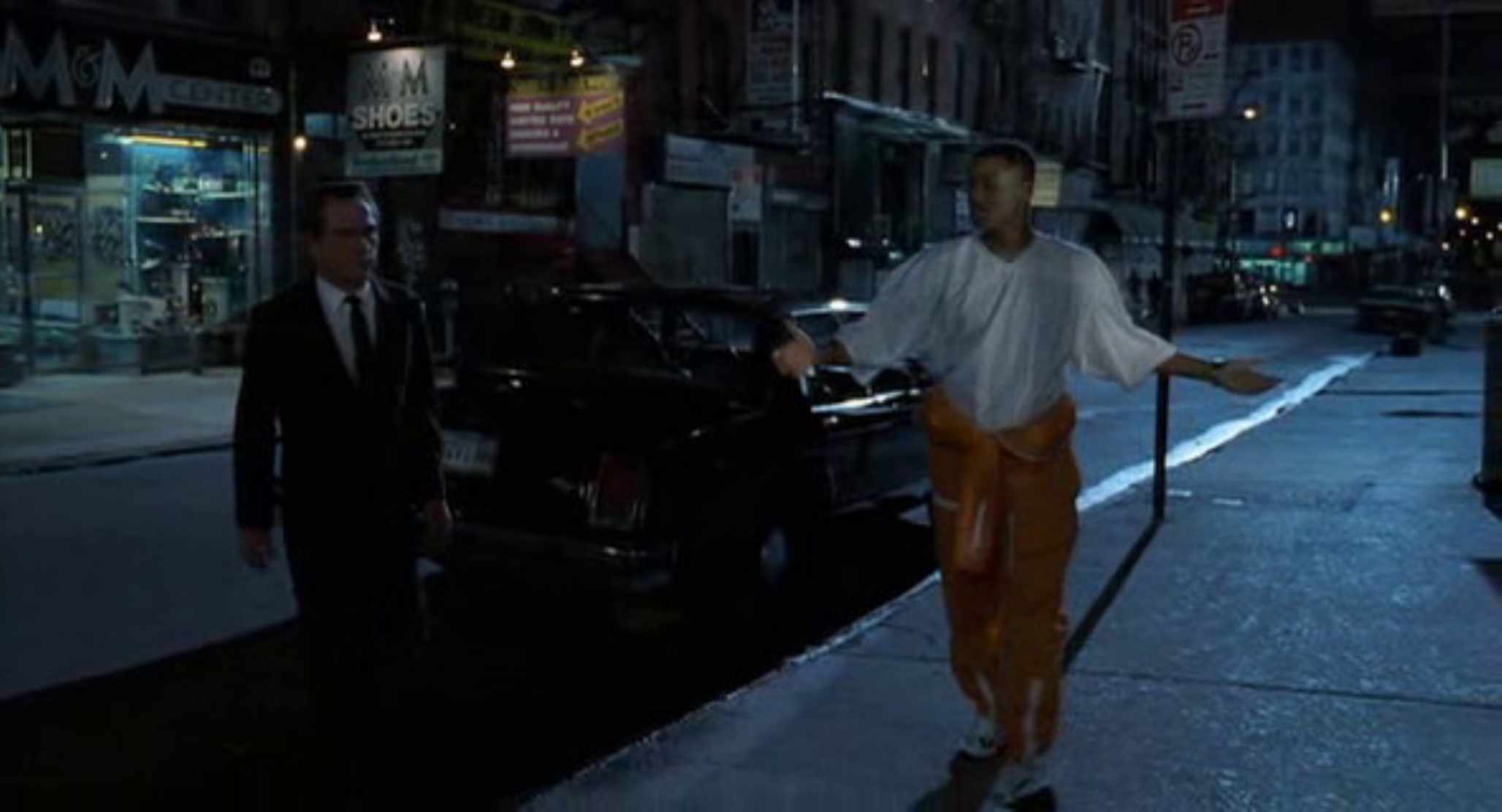}}
&
\scalebox{0.2}{\includegraphics[viewport=0cm 0.0cm 16.5cm 8.6cm]{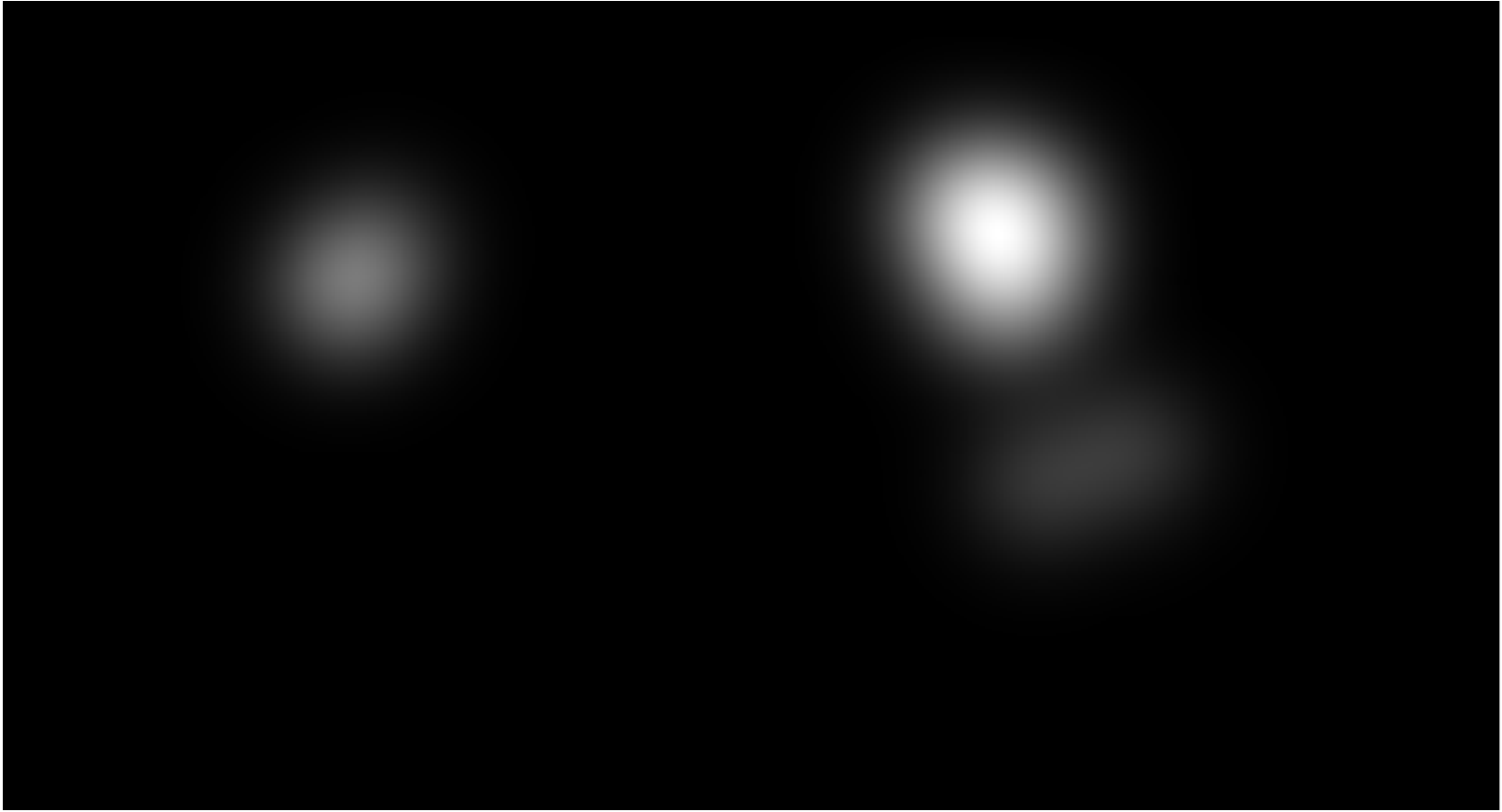}}
&
\scalebox{0.2}{\includegraphics[viewport=0cm 0.0cm 16.5cm 8.6cm]{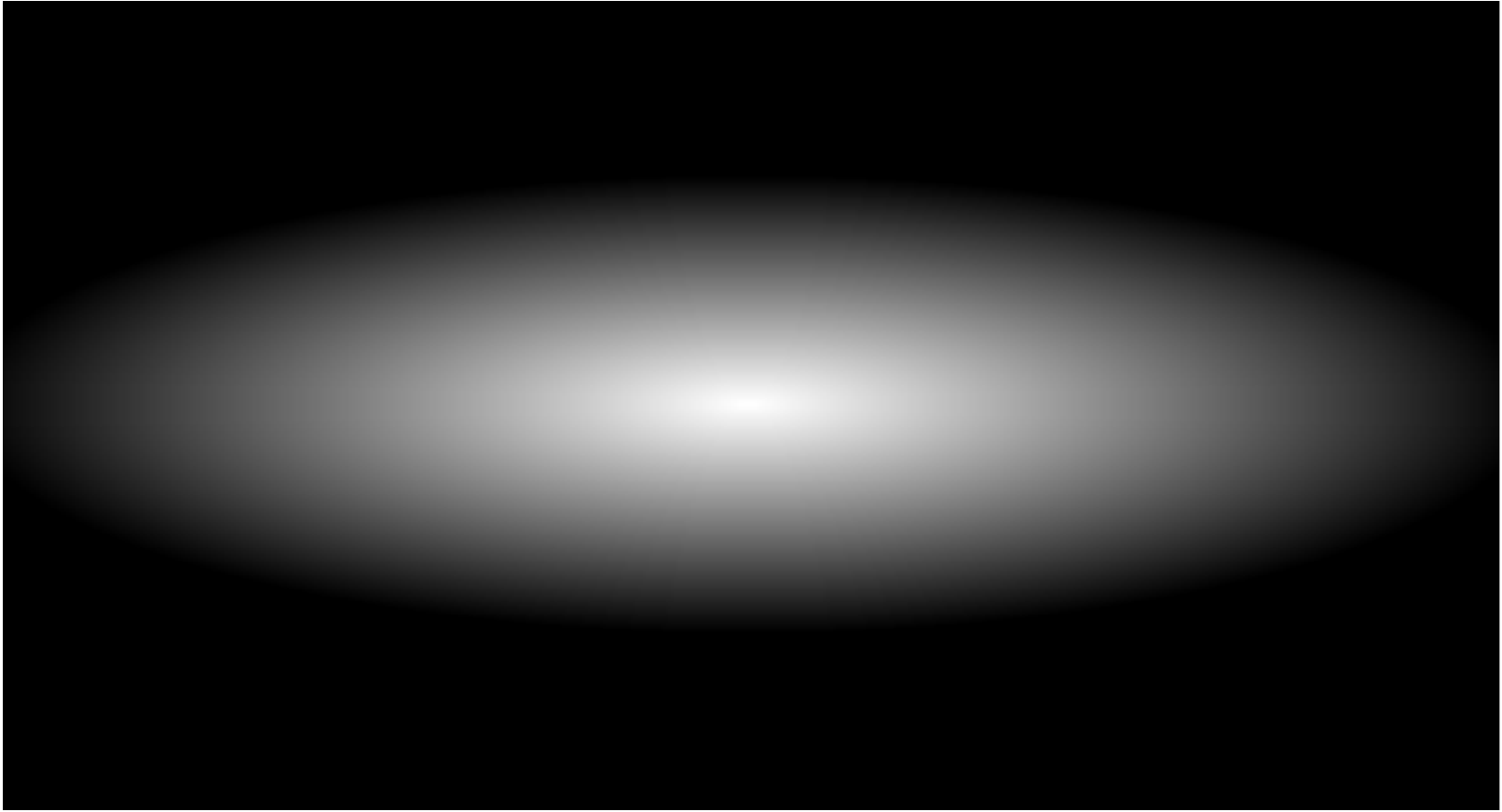}}
&
\scalebox{0.2}{\includegraphics[viewport=0cm 0.0cm 16.5cm 8.6cm]{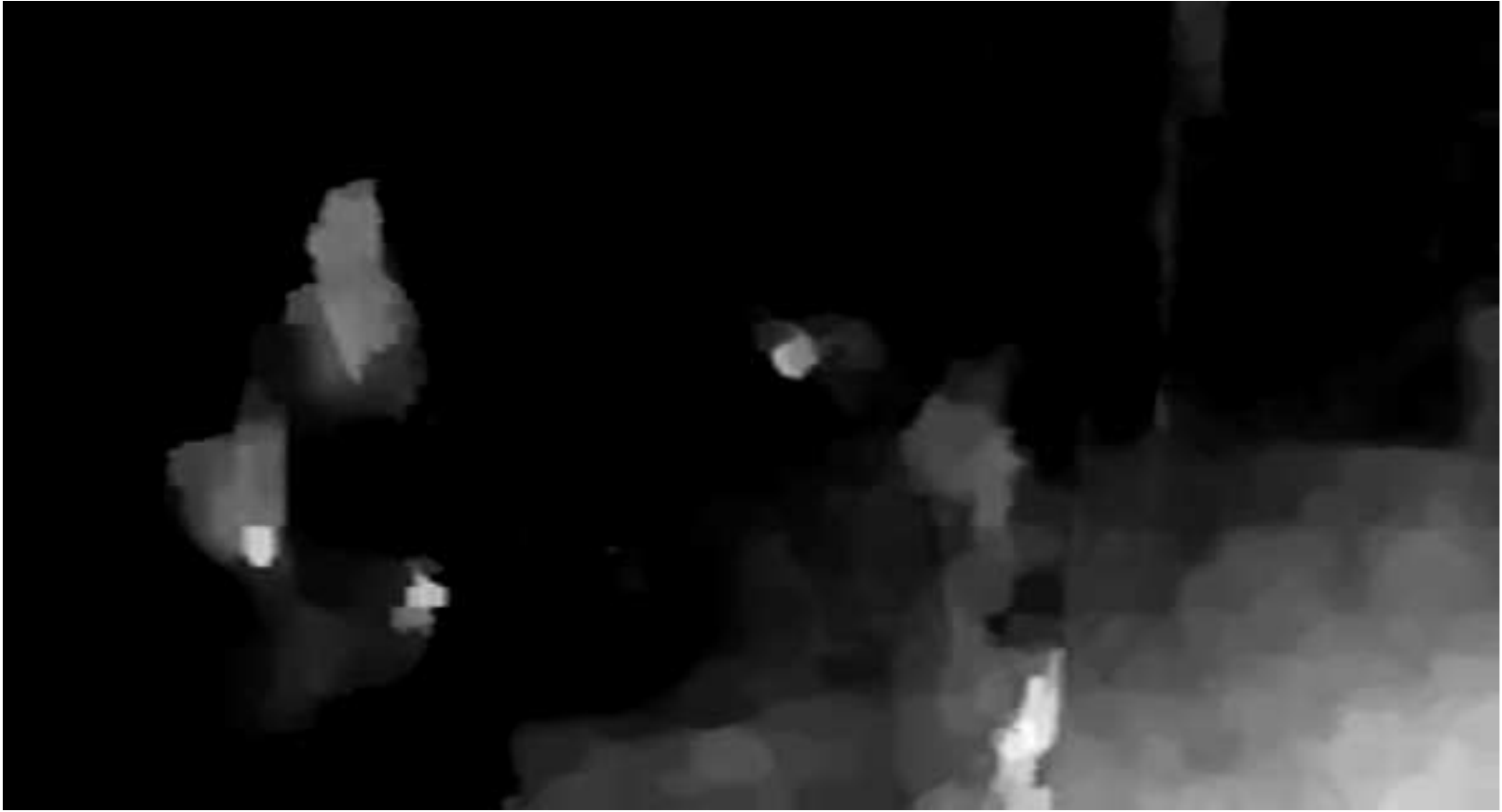}}
&
\scalebox{0.2}{\includegraphics[viewport=0cm 0.0cm 16.5cm 8.6cm]{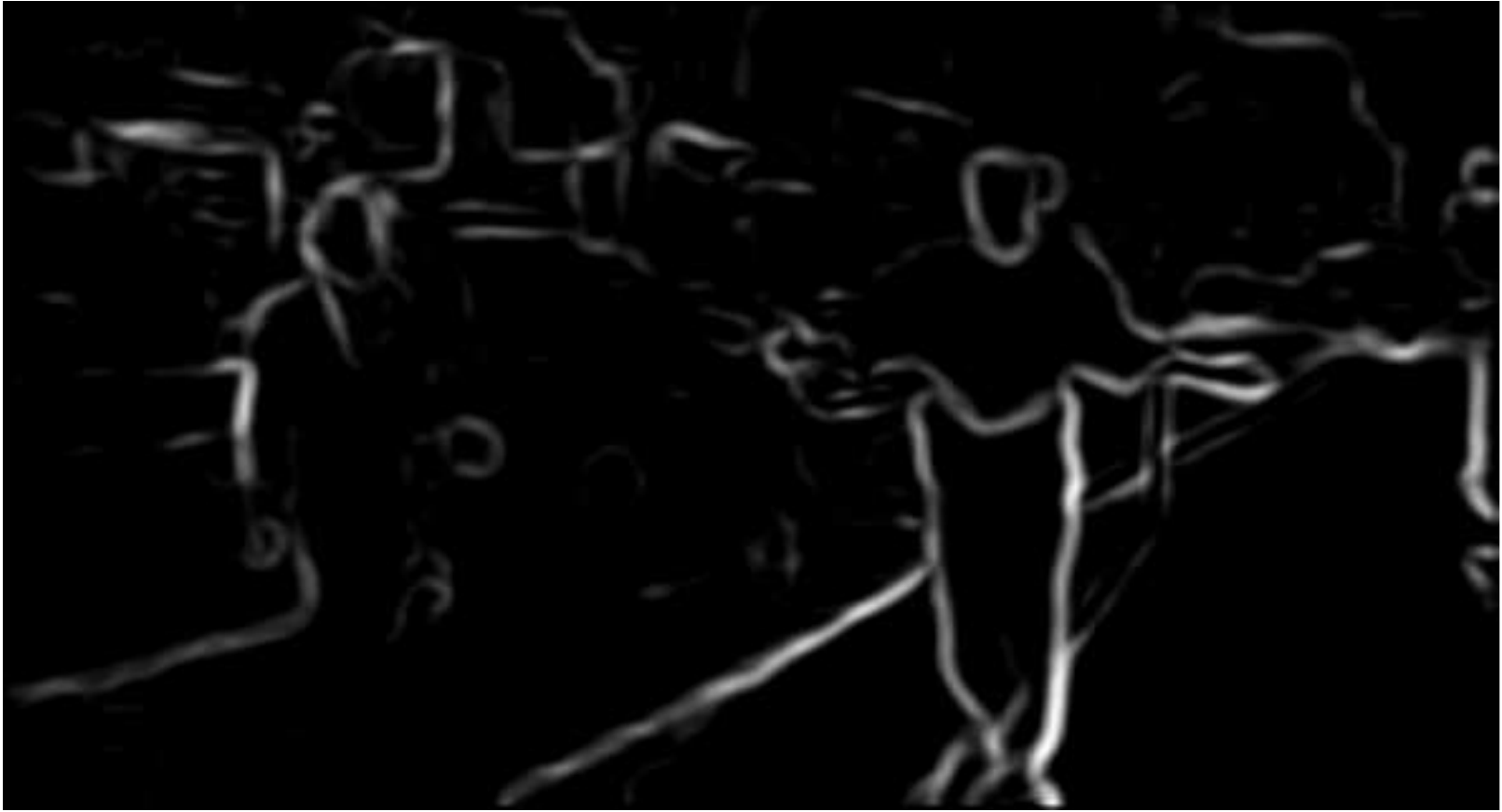}}
&
\scalebox{0.2}{\includegraphics[viewport=0cm 0.0cm 16.5cm 8.6cm]{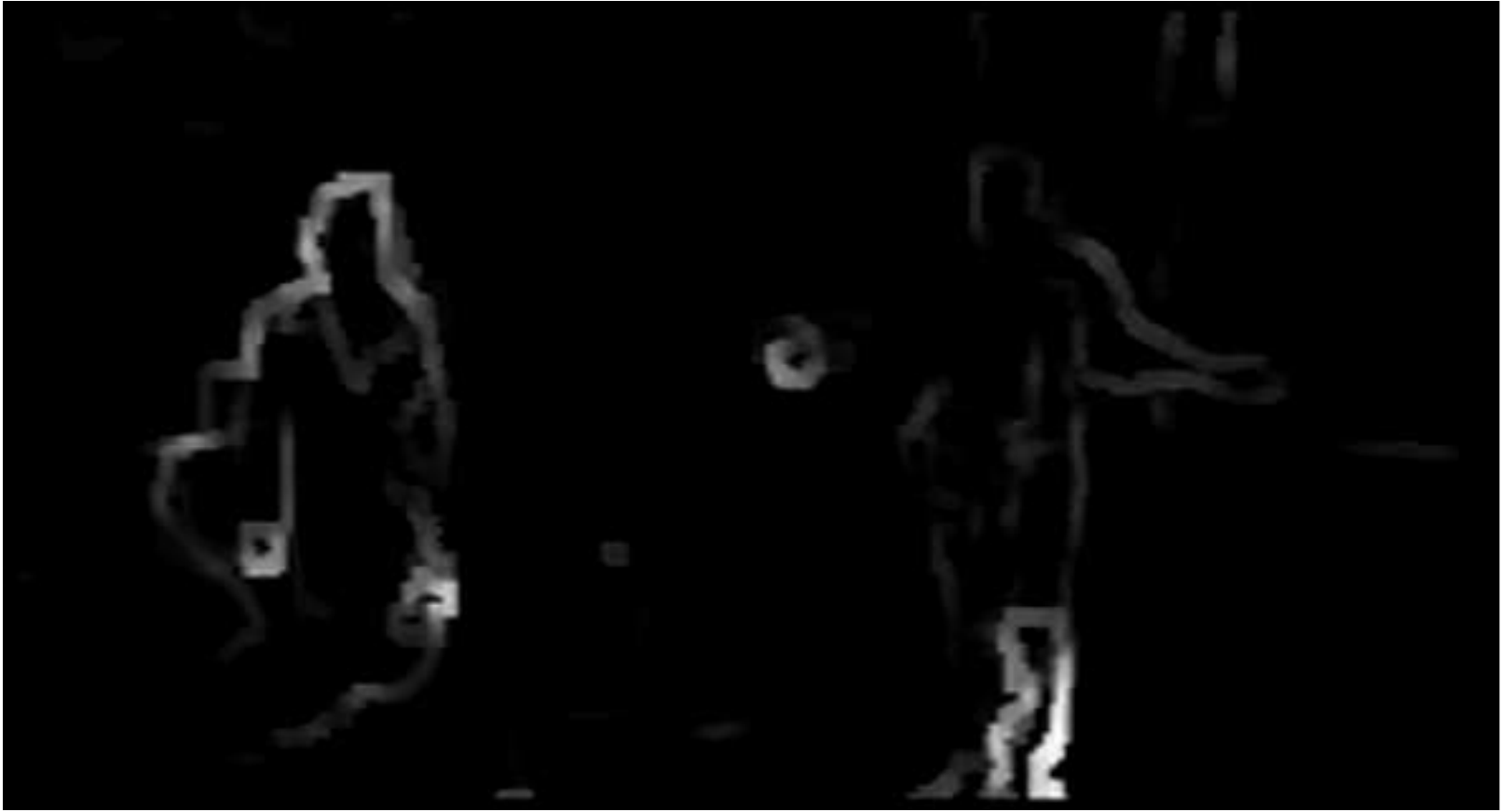}}
\\
(a) original image
%\vspace{1mm}
&
(b) ground truth saliency
%\vspace{1mm}
&
(c) CB
%central bias map
%\vspace{1mm}
&
(d) flow magnitude
%\vspace{1mm}
&
(e) pb edges with flow
%\vspace{1mm}
&
(f) flow bimodality
\\
\scalebox{0.156}{\includegraphics[viewport=0cm 0.0cm 20.9cm 11.6cm]{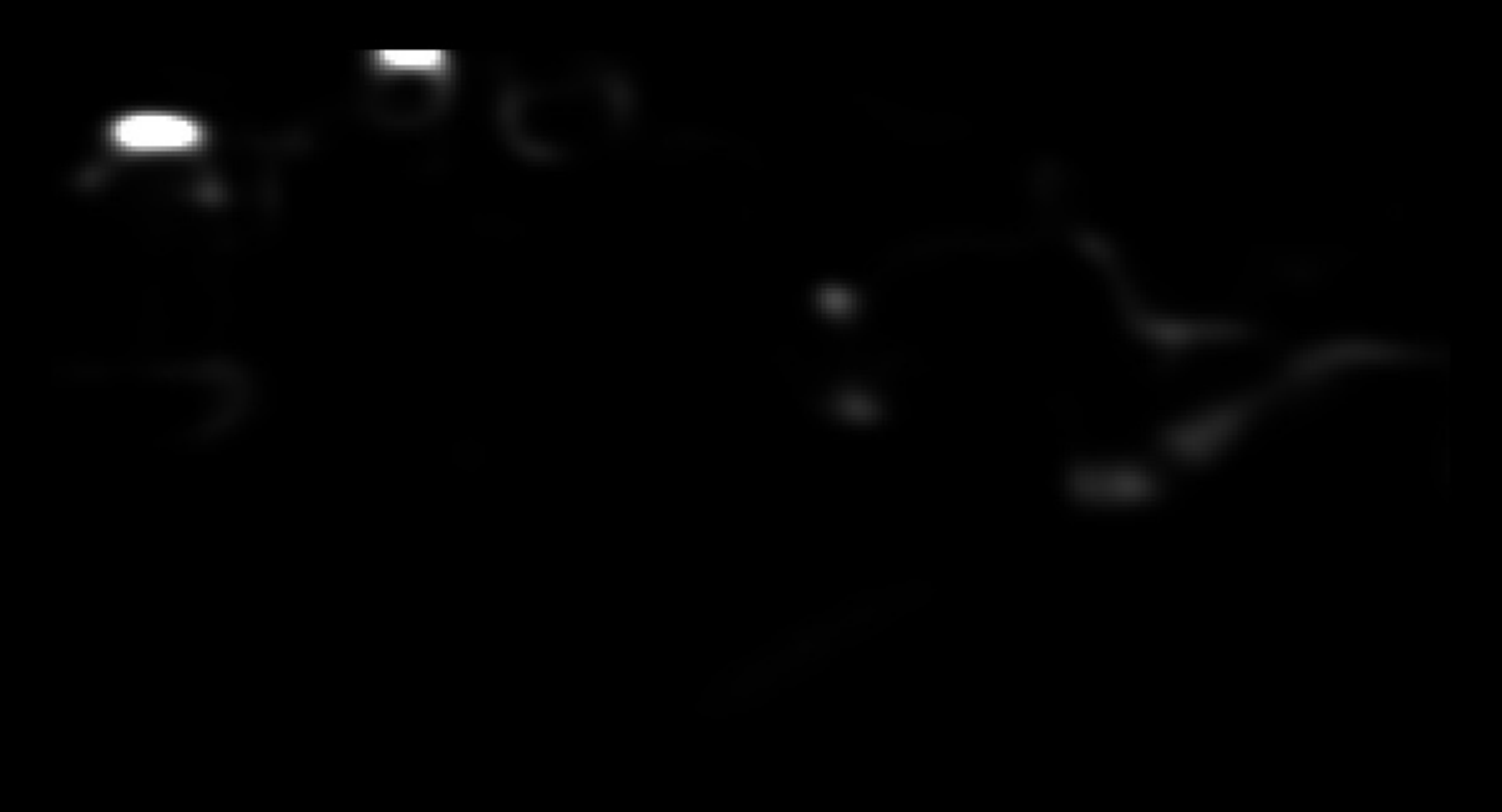}}
&
\scalebox{0.2}{\includegraphics[viewport=0cm 0.0cm 16.5cm 8.6cm]{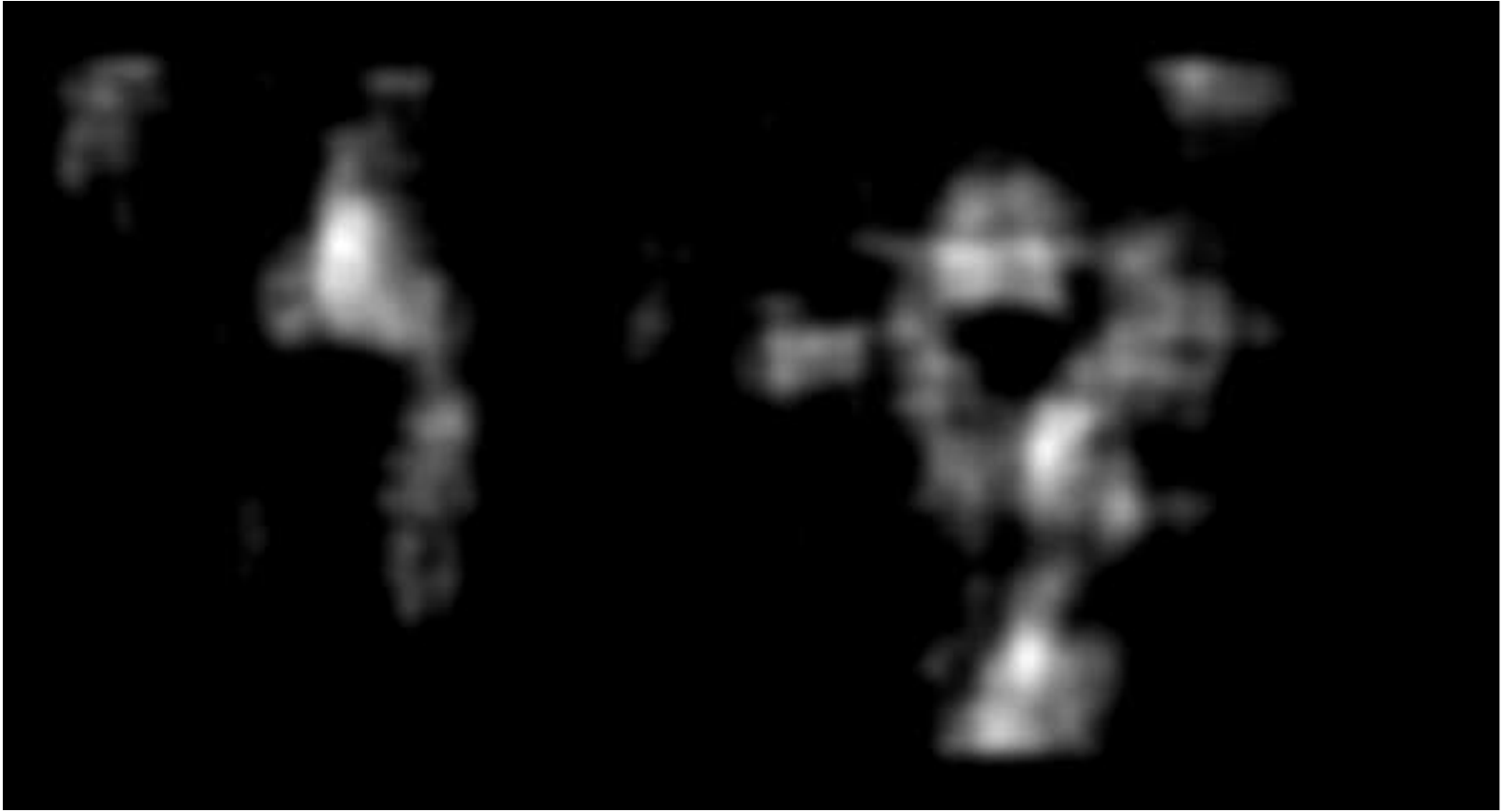}}
&
\scalebox{0.2}{\includegraphics[viewport=0cm 0.0cm 16.5cm 8.6cm]{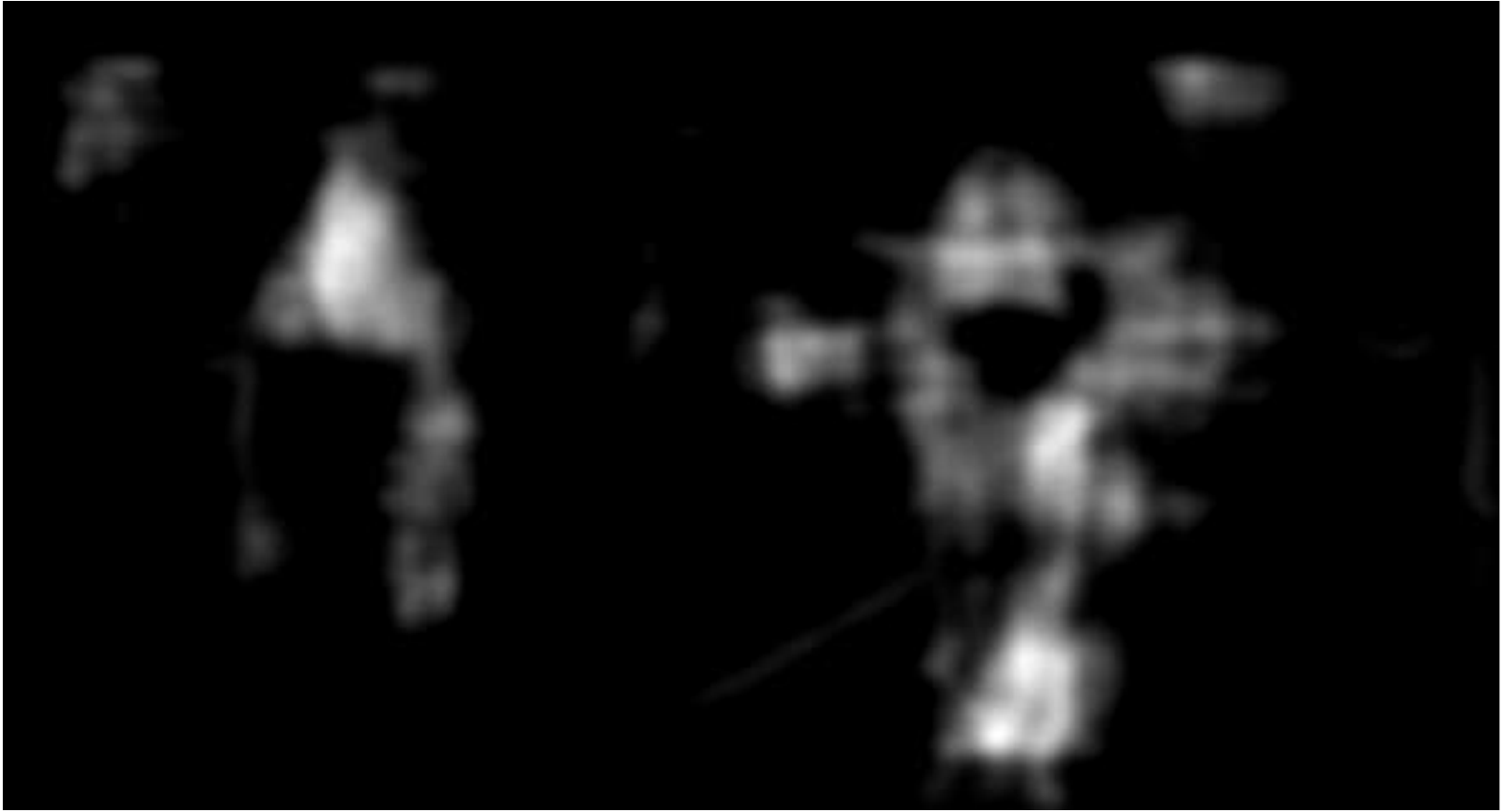}}
&
\scalebox{0.2}{\includegraphics[viewport=0cm 0.0cm 16.5cm 8.6cm]{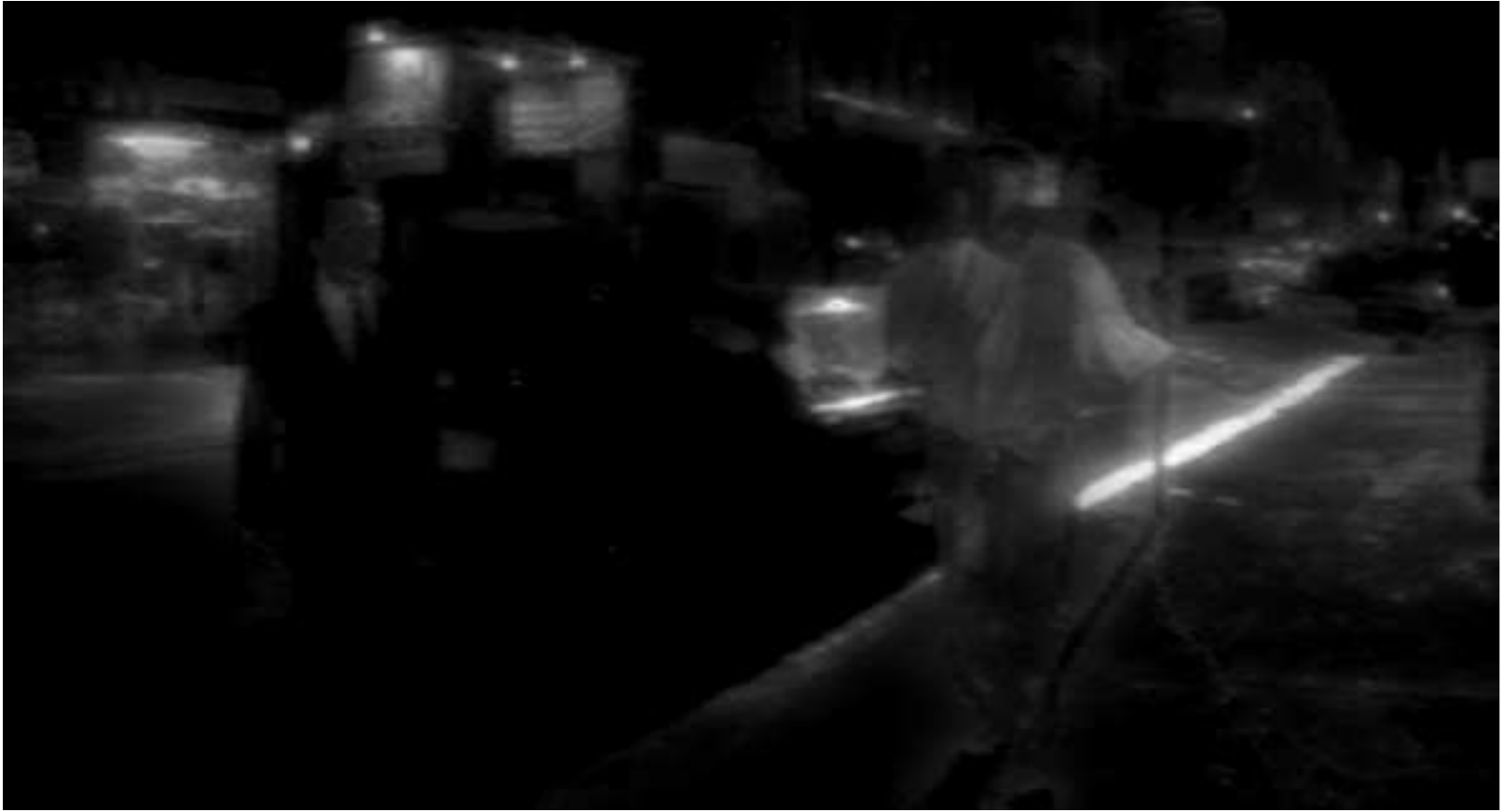}}
&
\scalebox{0.2}{\includegraphics[viewport=0cm 0.0cm 16.5cm 8.6cm]{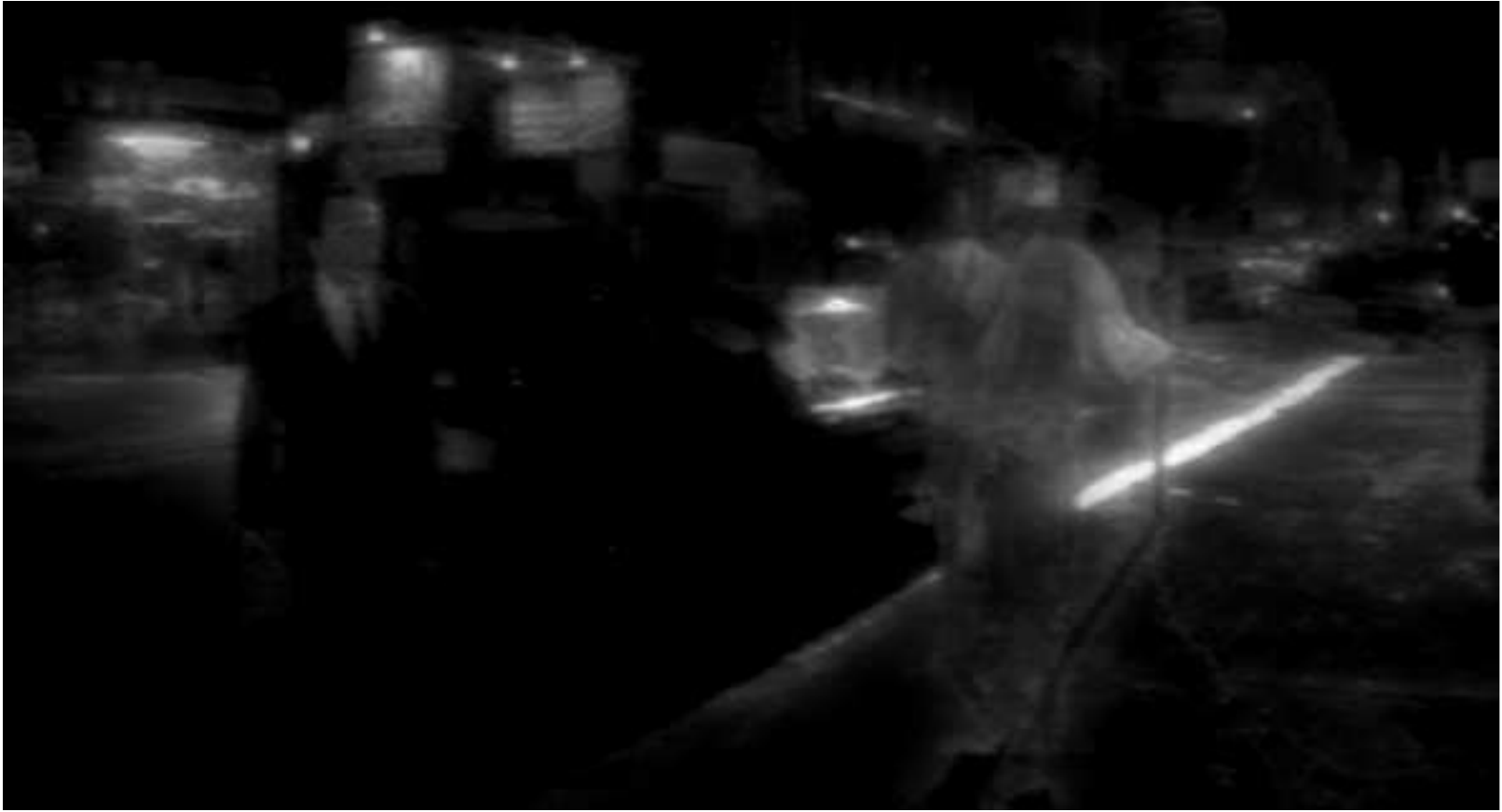}}
&
\scalebox{0.2}{\includegraphics[viewport=0cm 0.0cm 16.5cm 8.6cm]{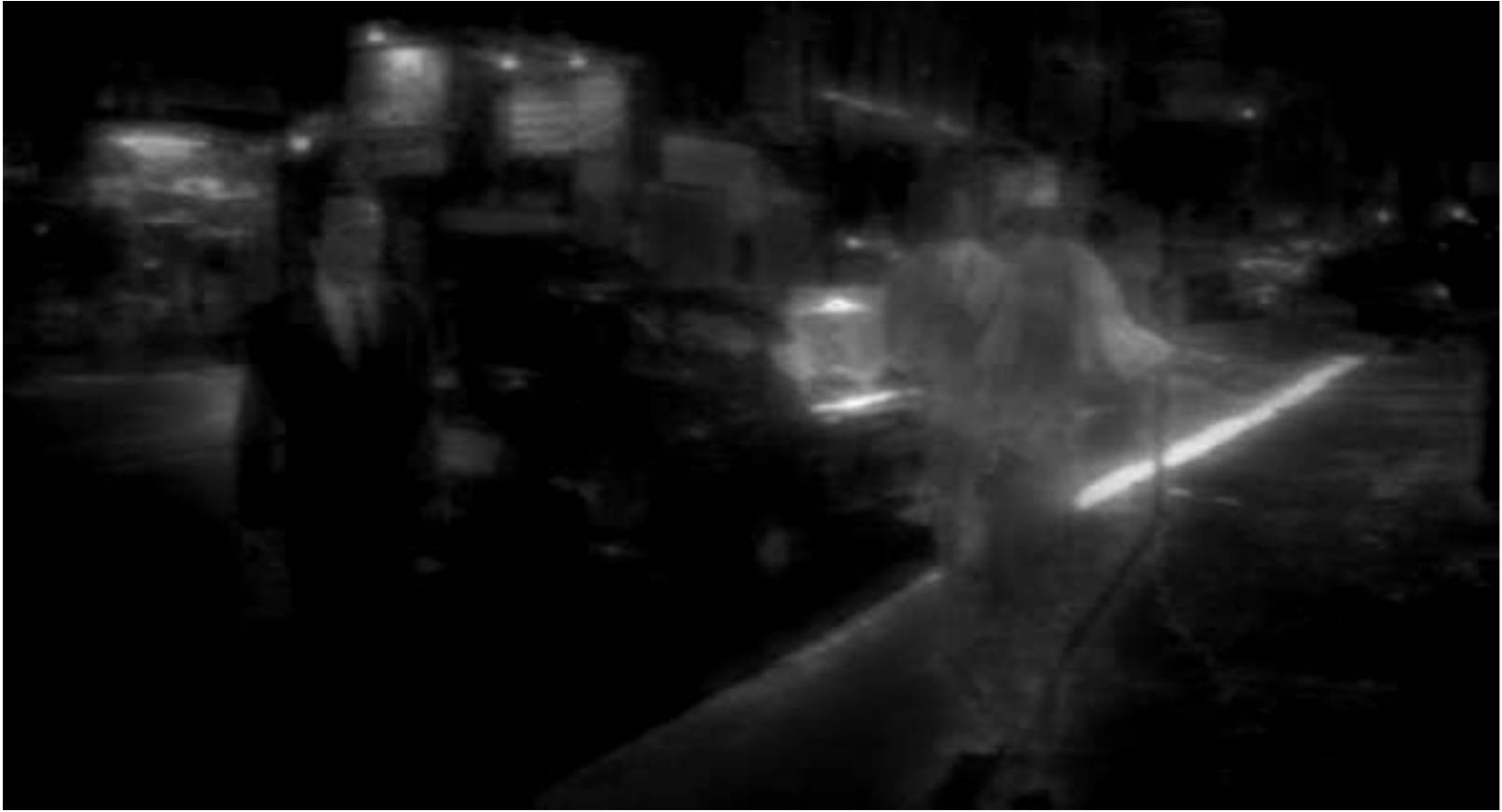}}
\\
(g) Harris cornerness
%\vspace{1mm}
&
(h) HoG-MBH detector
%\vspace{1mm}
&
(i) MF
%\vspace{1mm}
& 
(j) SF
%\vspace{1mm}
&
(k) SF + MF
%\vspace{1mm}
&
(l) SF + MF + CB
%\vspace{1mm}
\end{tabular}}
\end{center}

\caption{Saliency predictions for a video frame (a), both motion-based features in isolation (d-h) and combinations (i-l).
HoG-MBH detector maps are closest to the ground truth (b), consistent with Table \ref{t:judd_measures}b.
}\label{fig:features}
\end{figure*}

\begin{figure*}
\begin{center}
\scalebox{0.72}{
\begin{tabular}{cccccc}
\scalebox{0.22}{\includegraphics[viewport=0.0cm 0.1cm 16.4cm 9.1cm]{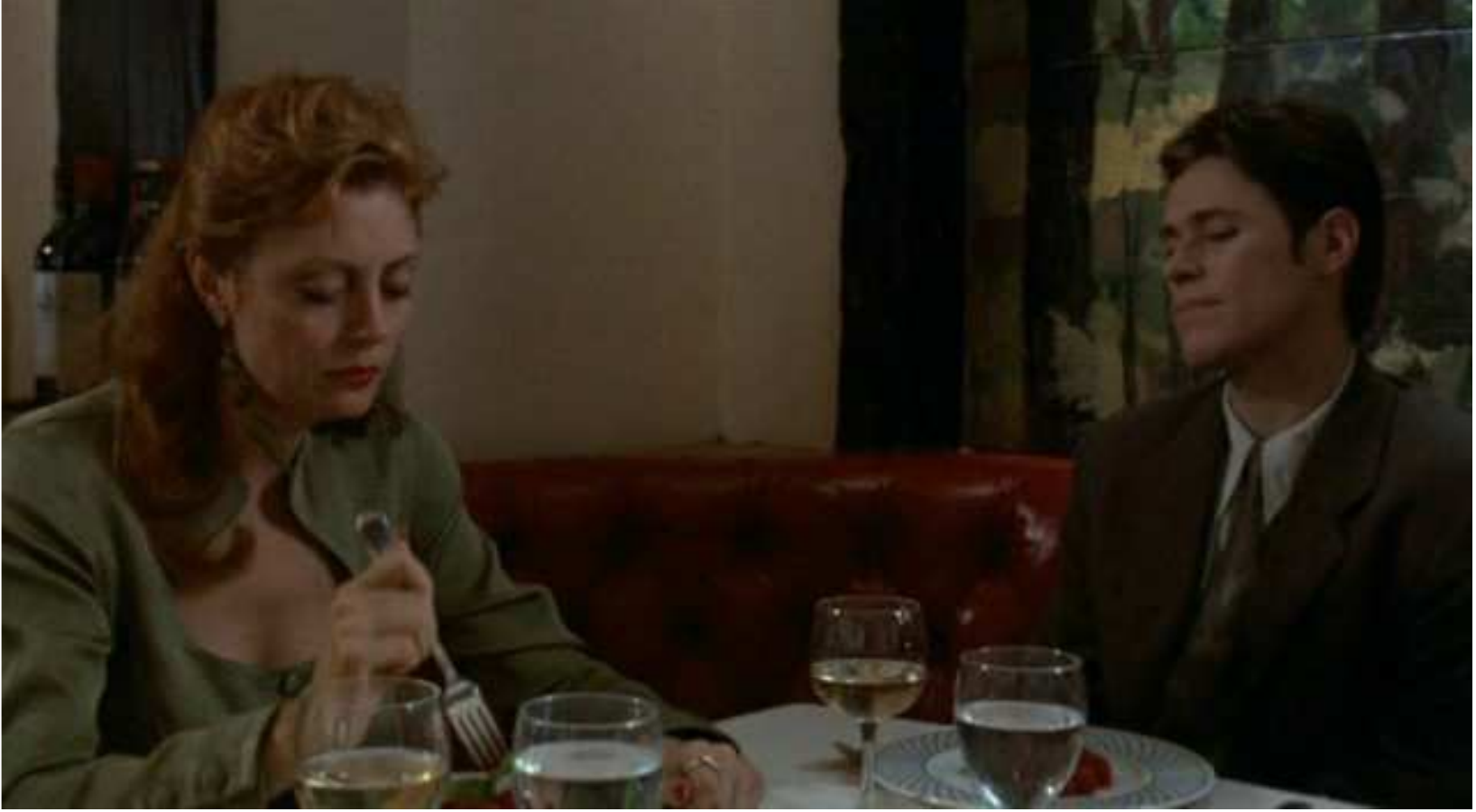}}
\hspace{1mm}
&
\scalebox{0.22}{\includegraphics[viewport=0.0cm 0.1cm 16.4cm 9.1cm]{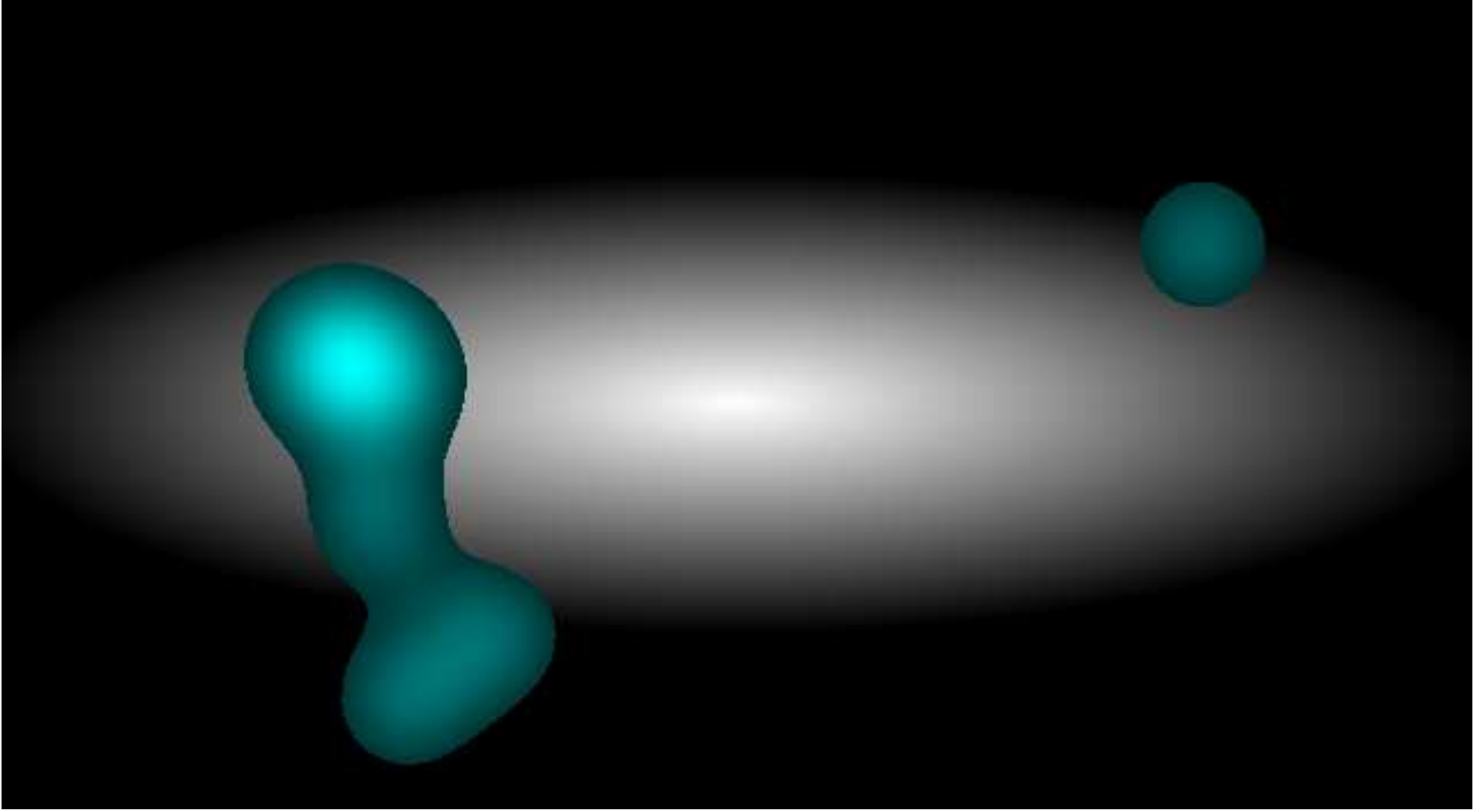}}
\hspace{1mm}
&
\scalebox{0.22}{\includegraphics[viewport=0.0cm 0.1cm 16.4cm 9.1cm]{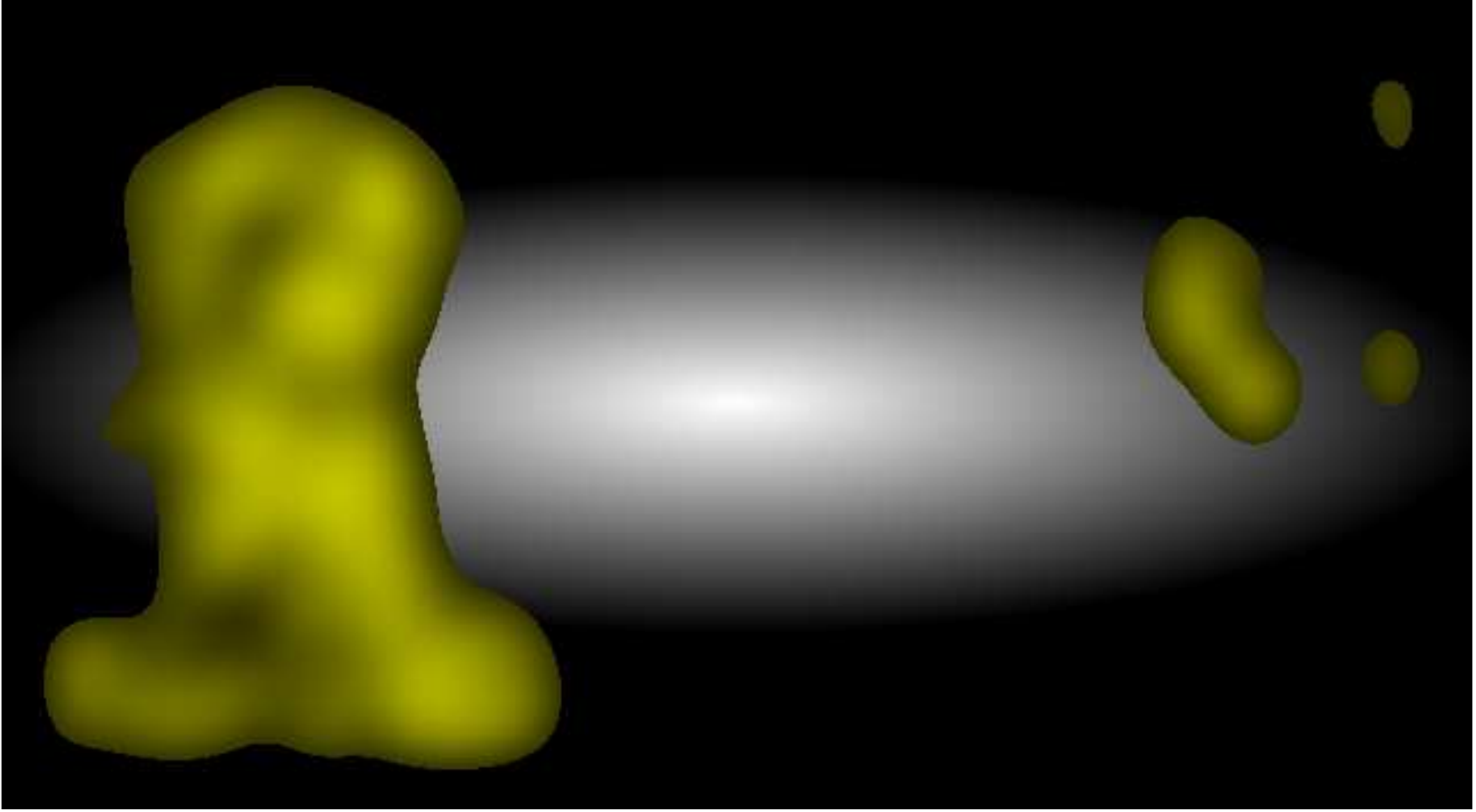}}
\hspace{1mm}
&
\scalebox{0.22}{\includegraphics[viewport=0.0cm 0.1cm 16.3cm 8.8cm]{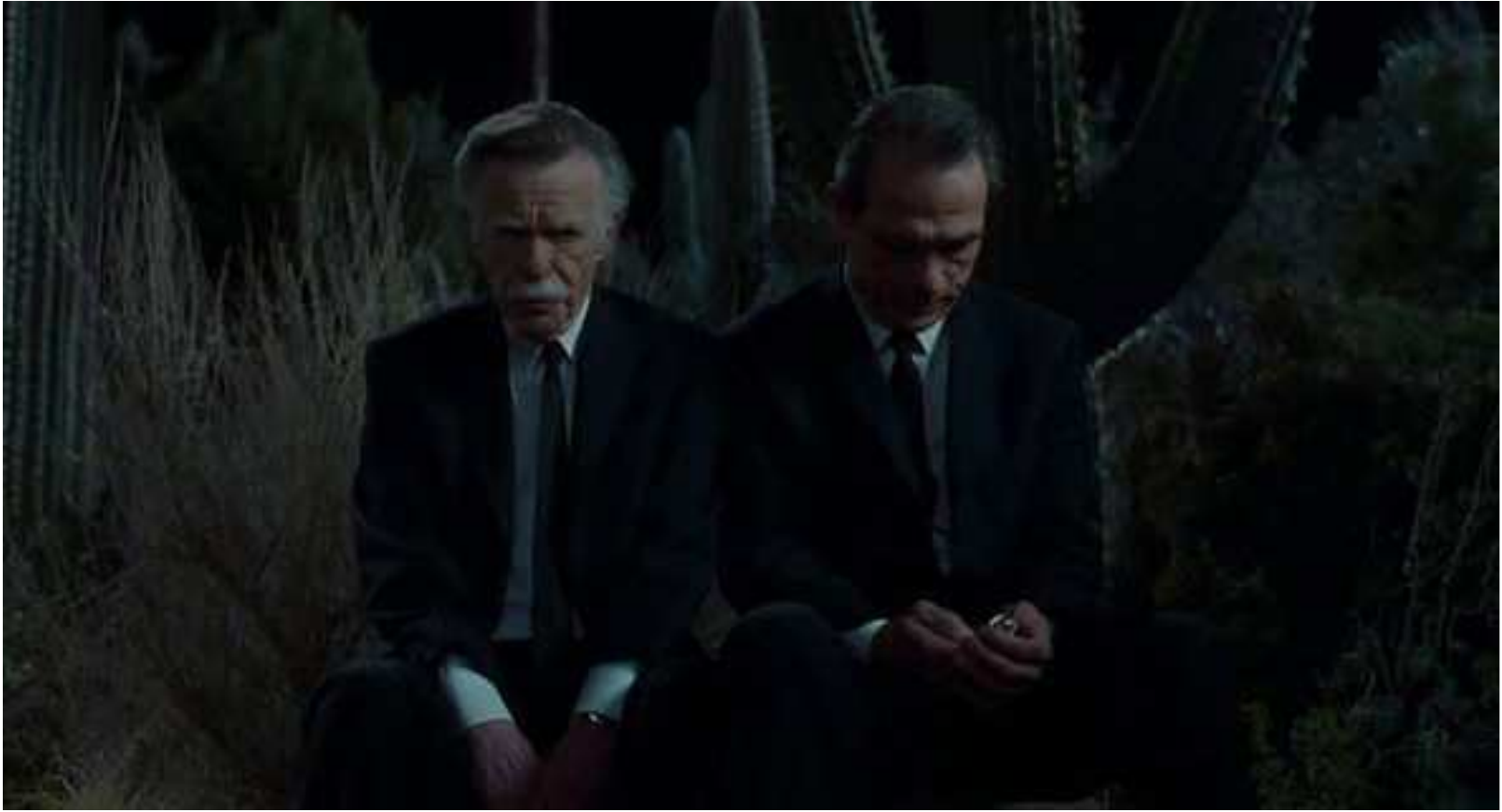}}
\hspace{1mm}
&
\scalebox{0.22}{\includegraphics[viewport=0.0cm 0cm 16.3cm 8.9cm]{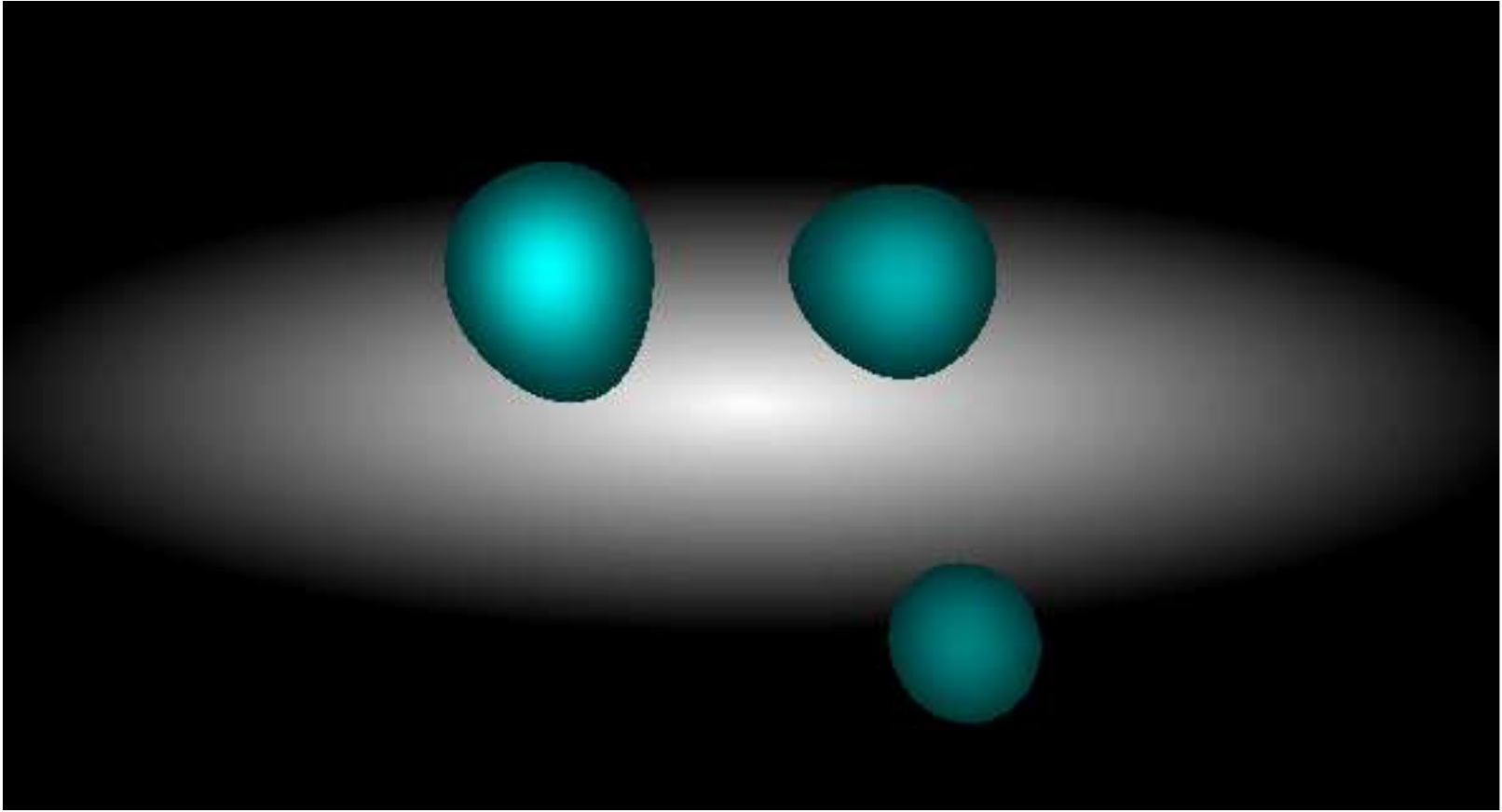}}
\hspace{1mm}
&
\scalebox{0.22}{\includegraphics[viewport=0.0cm 0cm 16.3cm 8.9cm]{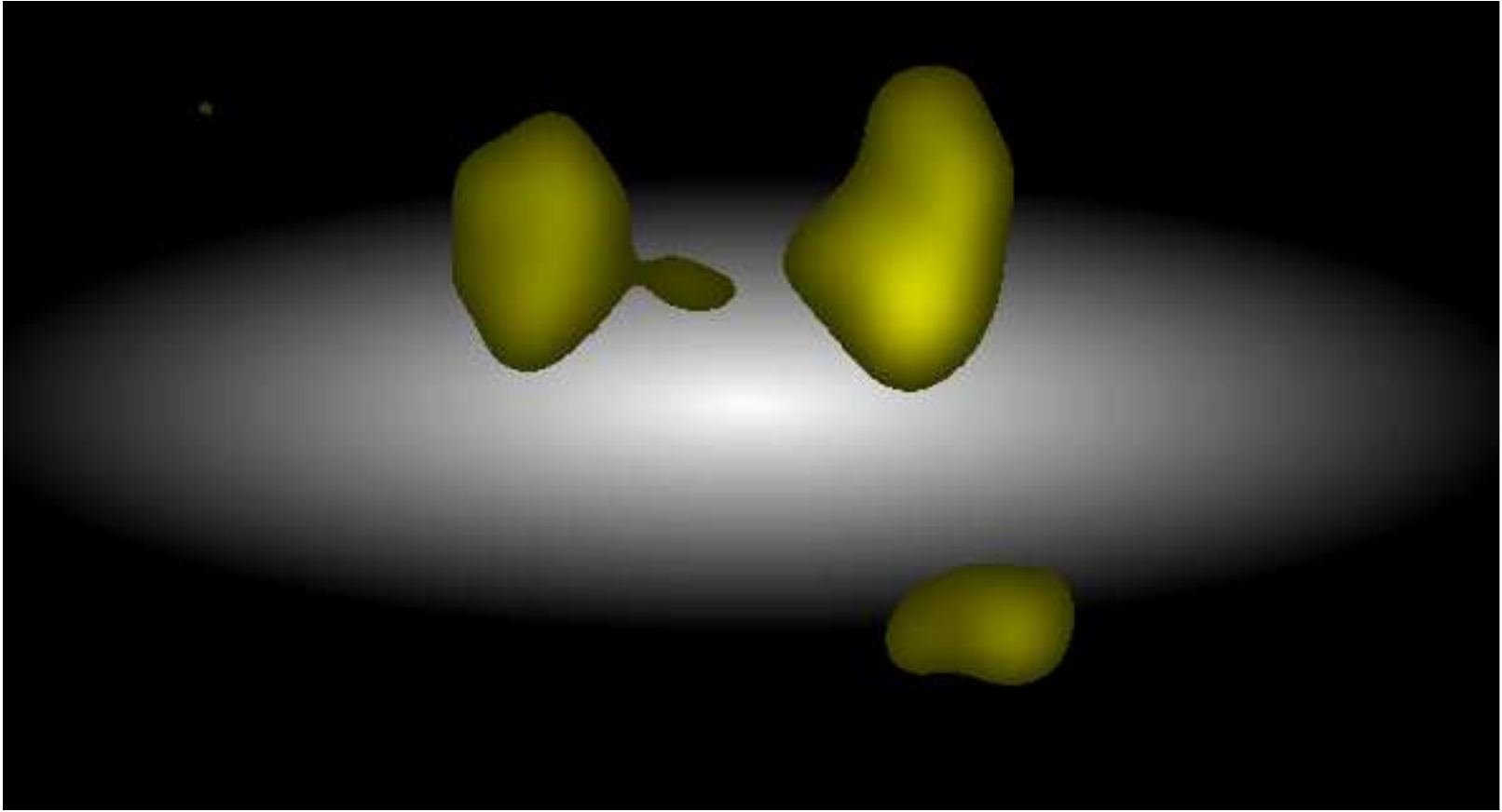}}
\hspace{1mm}

\\
image
&
ground truth/CB
&
HoG-MBH detector/CB
&
image
&
ground truth/CB
&
HoG-MBH detector/CB
\\
\scalebox{0.22}{\includegraphics[viewport=0.0cm 0cm 16.4cm 6.9cm]{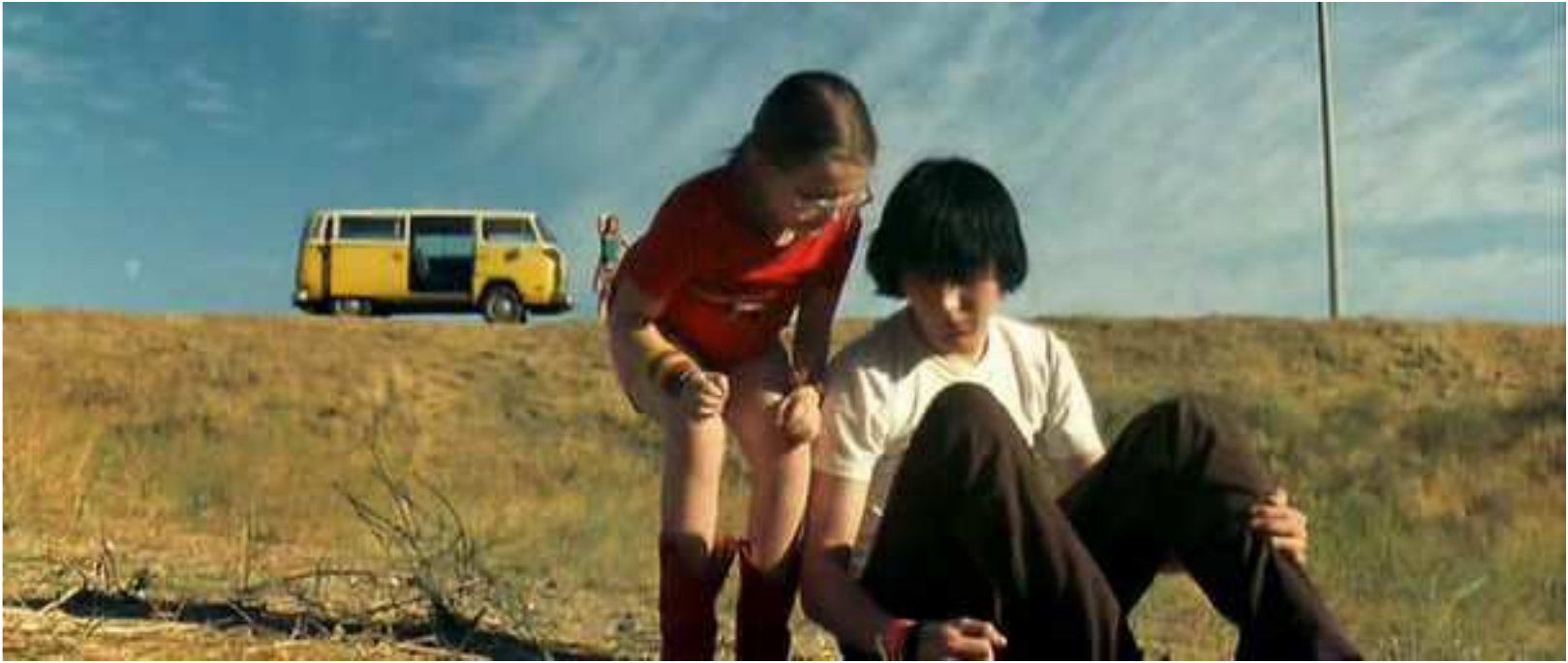}}
\hspace{1mm}
&
\scalebox{0.22}{\includegraphics[viewport=0.0cm 0cm 16.4cm 6.9cm]{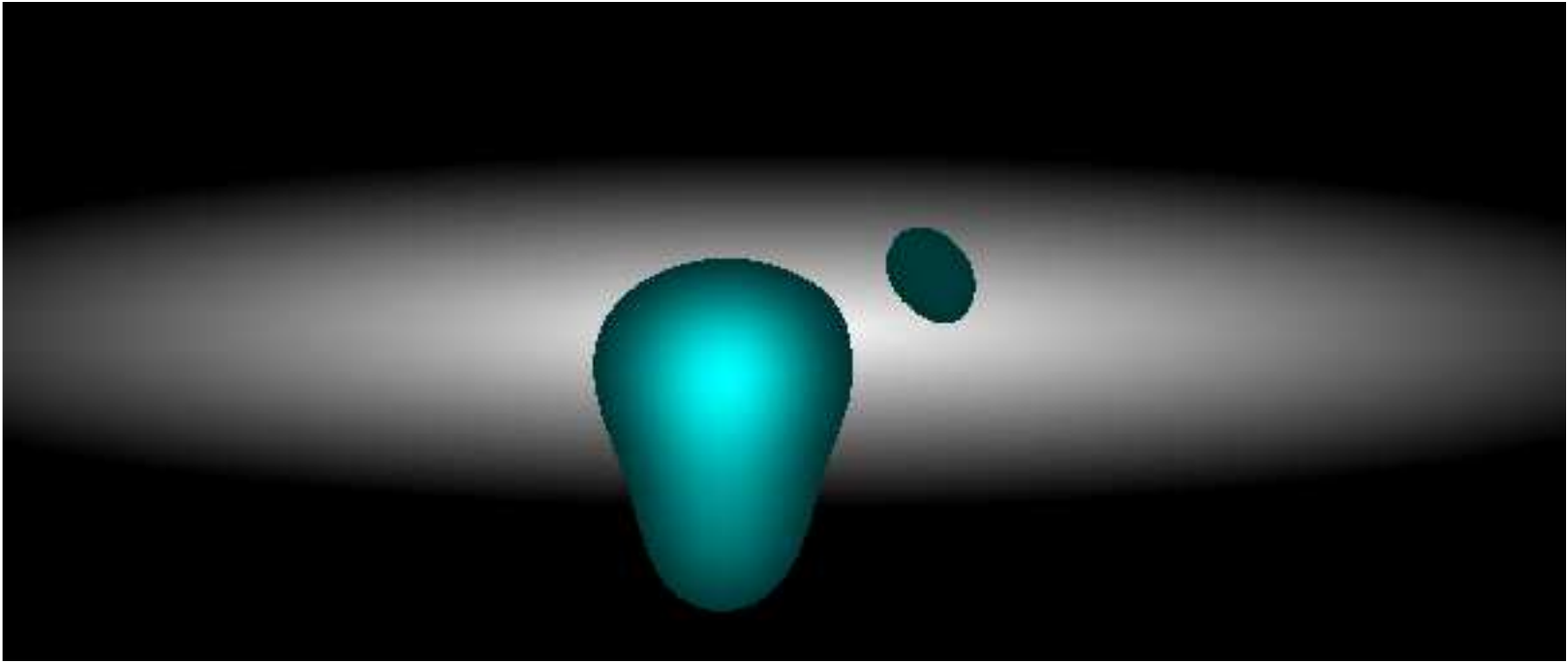}}
\hspace{1mm}
&
\scalebox{0.22}{\includegraphics[viewport=0.0cm 0cm 16.4cm 6.9cm]{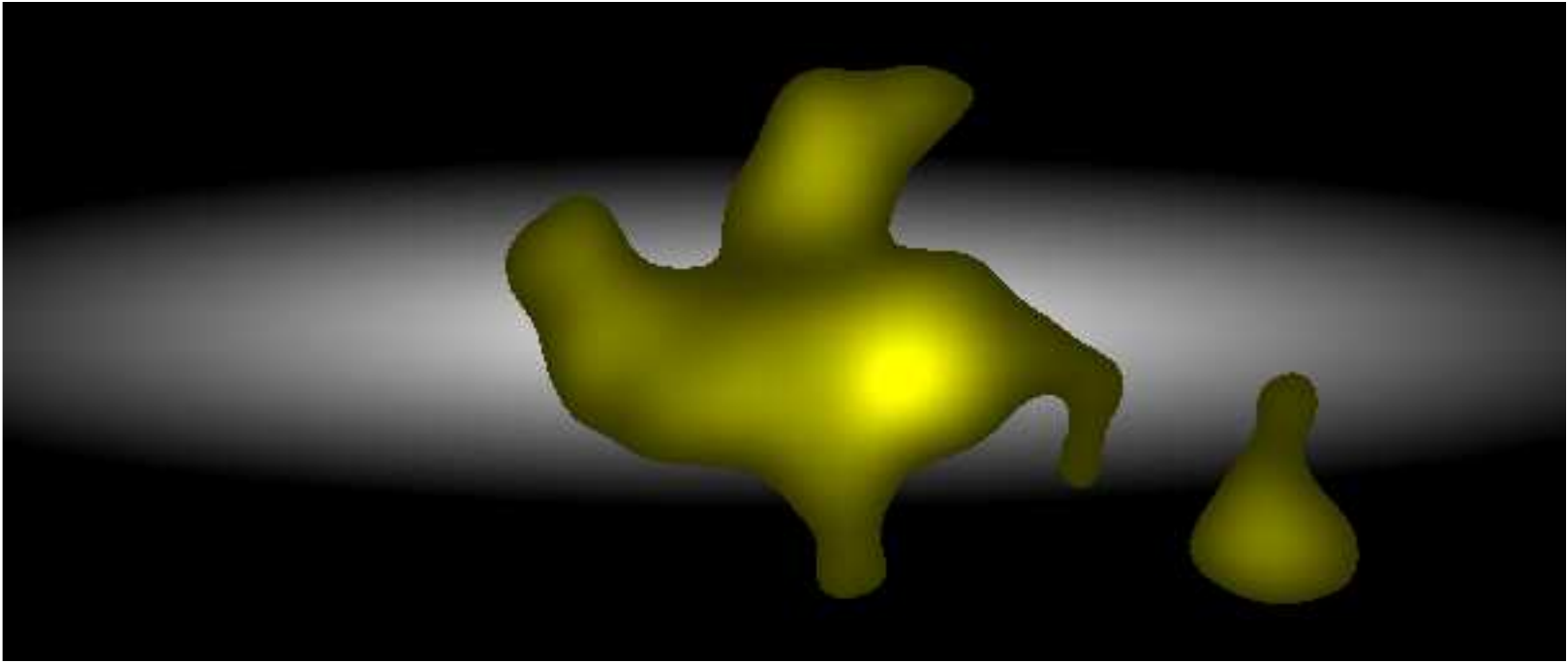}}
\hspace{1mm}
&
\scalebox{0.22}{\includegraphics[viewport=0.0cm 0.1cm 16.5cm 6.8cm]{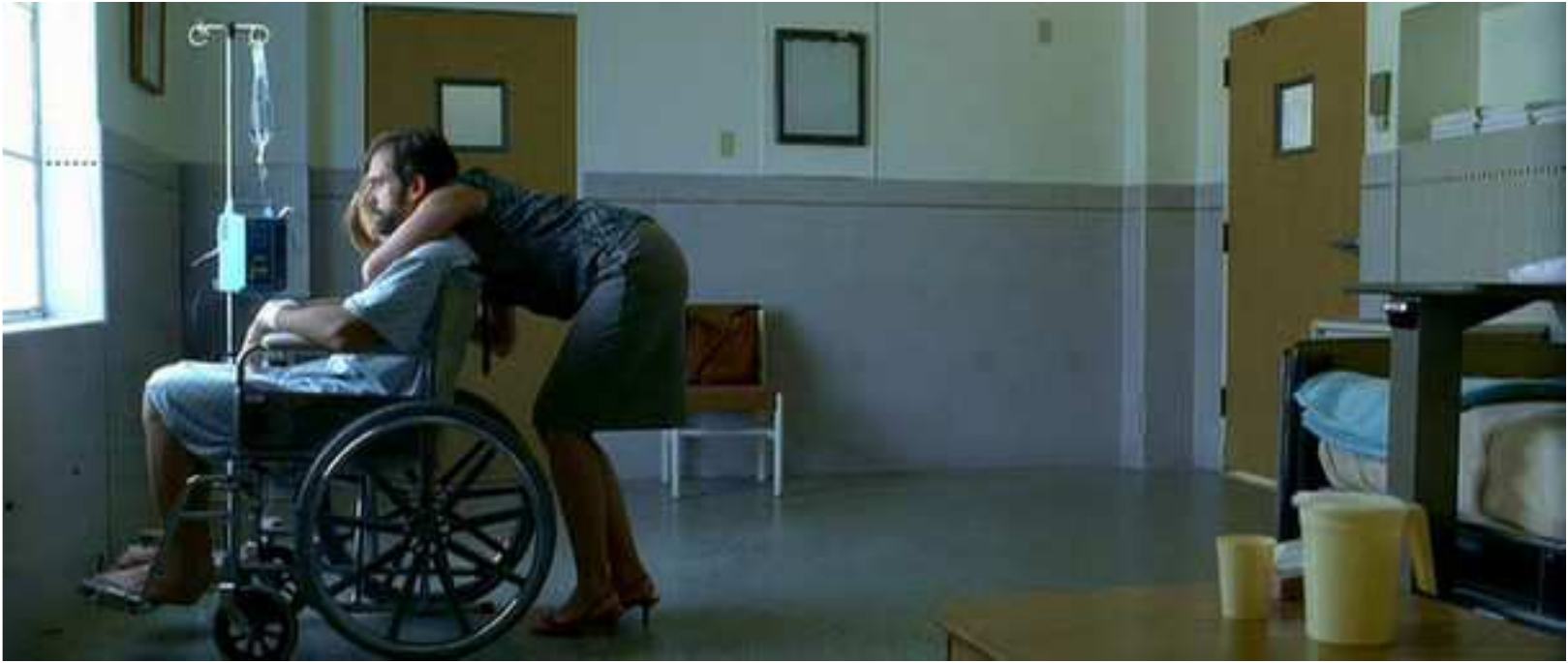}}
\hspace{1mm}
&
\scalebox{0.22}{\includegraphics[viewport=0.0cm 0cm 16.5cm 7.0cm]{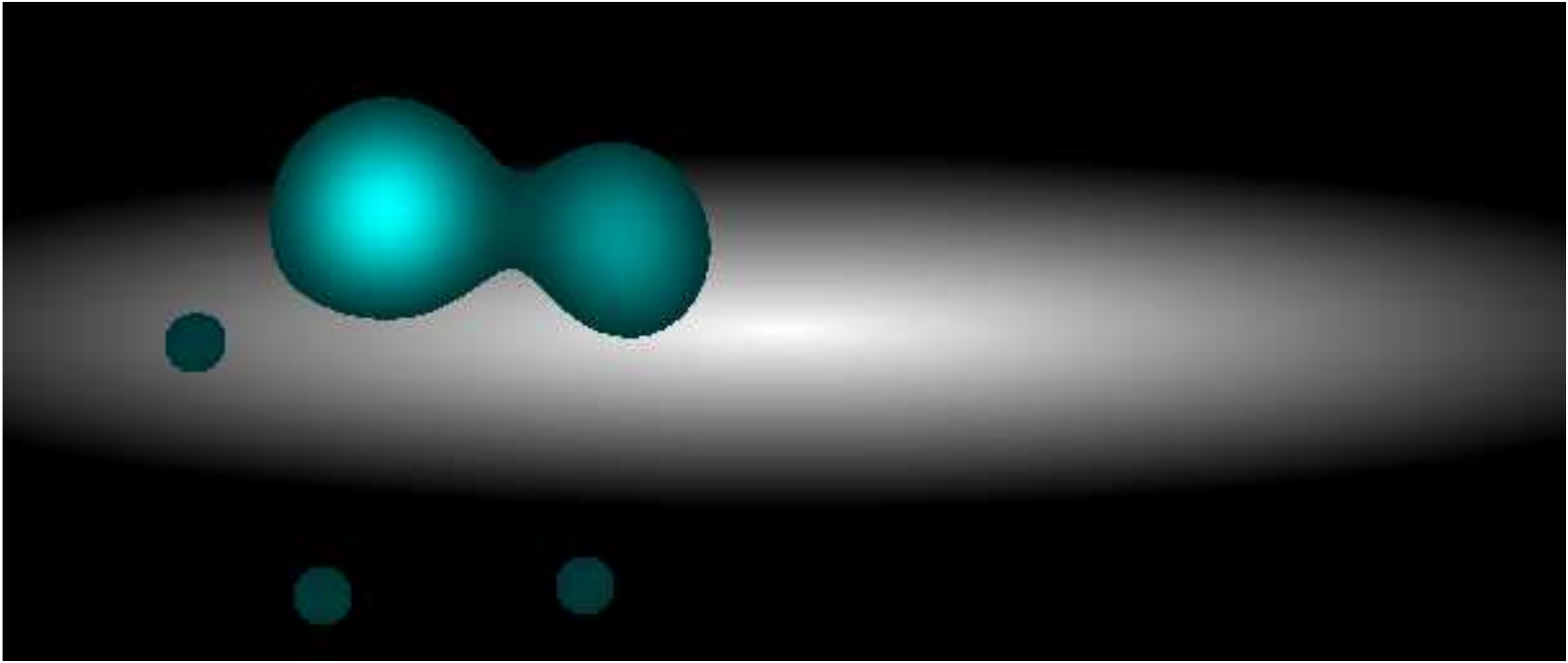}}
\hspace{1mm}
&
\scalebox{0.22}{\includegraphics[viewport=0.8cm 0cm 16.5cm 7.0cm]{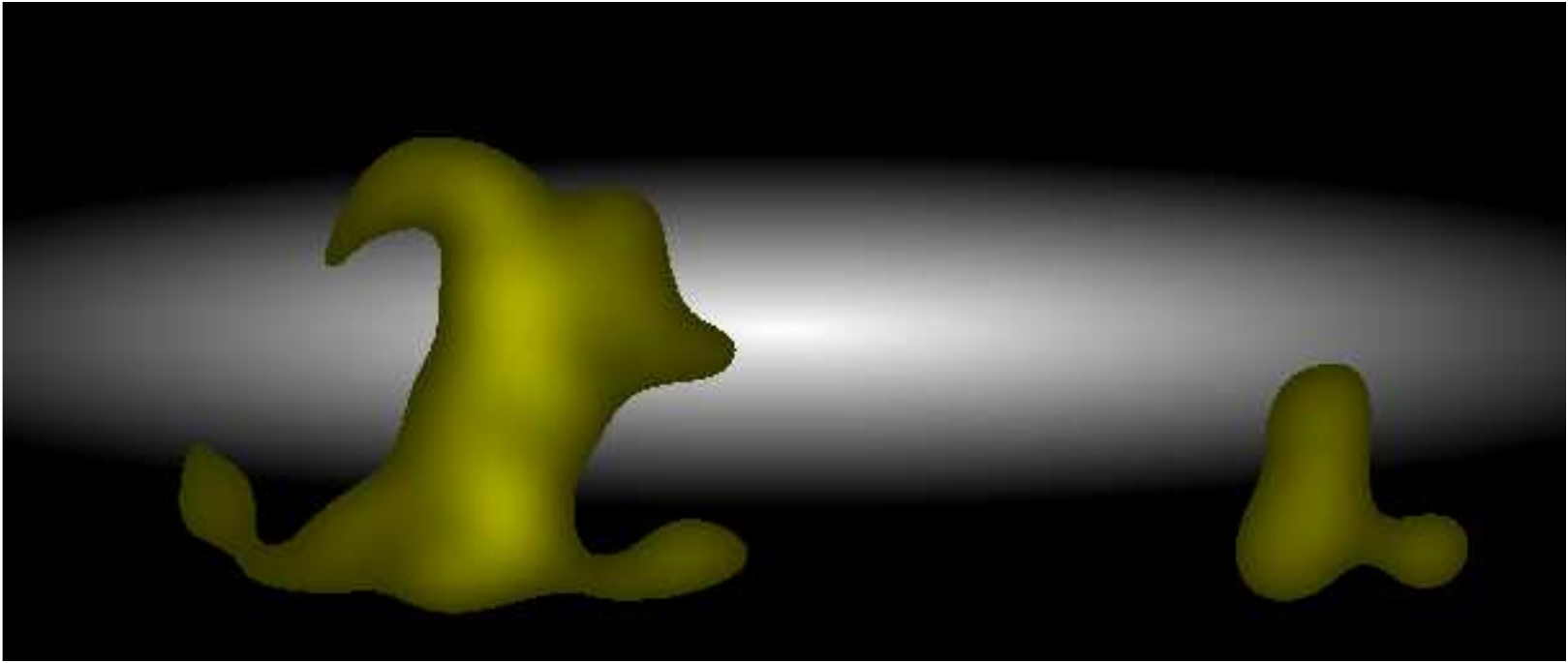}}
\\
image
&
ground truth/CB
&
HoG-MBH detector/CB
&
image
&
ground truth/CB
&
HoG-MBH detector/CB
\end{tabular}
}
\end{center}
\caption{Note that ground truth saliency maps (cyan) and output of our HoG-MBH detector (yellow) are similar to each other, but qualitatively different from the central bias map (gray). This gives visual intuition for the significantly higher performance of the HoG-MBH detector, over the central bias saliency sampling, when used in an end-to-end computer visual action recognition system (Table \ref{t:classification}c,d,e).}\label{fig:cb_det_comparison}
\end{figure*}

\section{Automatic Visual Action Recognition}\label{s:end_to_end_recognition}

We next investigate action recognition performance when interest points are sampled from the saliency maps predicted by our HoG-MBH detector, which we choose because it best approximates the ground truth saliency map spatially, under the KL divergence metric. Apart from sampling from the uniform and ground truth distributions, as a second baseline, we also sample interest points using the central bias saliency map, which was also shown to approximate to a reasonable extent human fixations under the less intuitive AOI measure (Table \ref{t:judd_measures}). We also investigate whether our end-to-end recognition system can be combined with the state-of-the-art approach of \cite{WangEtAl2011} to obtain better performance.

\noindent\textbf{Experimental Protocol}: We first run our HoG-MBH detector over the entire Hollywood-2 data set and obtain our automatic saliency maps. We then configure our recognition pipeline with an interest point operator that samples locations using these saliency maps as probability distributions.  We also run the pipeline of \cite{WangEtAl2011} and combine the four kernel matrices produced in the final stage of their classifier with the ones we obtain for our 14 descriptors, sampled from the saliency maps, using MKL.

We also test our recognition pipeline on the UCF Sports dataset, which is substantially different in terms of action classes, scene clutter, shooting conditions and the evaluation procedure. Unlike Hollywood-2, this database provides no training and test sets, and classifiers are generally evaluated by cross-validation. We follow the standard procedure by first extending the dataset with horizontally flipped versions of each video. For each cross-validation fold, we leave out one original video and its flipped version and train a multi-class classifier. We test on the original video, but not its flipped version. We compute the confusion matrix and report the average accuracy over all classes.

Our experimental procedure for the UCF Sports dataset closely follows the one we use for Hollywood-2. We re-train our HoG-MBH detector on a subset of 50 video pairs (original and flipped), while we use the rest of 100 pairs for testing. The average precision of our detector is $92.5\%$ for the training set and $93.1\%$ on our test set, which confirms that our detector does not overfit the data. We use the re-trained detector to run the same pipelines and baselines as for Hollywood-2.
%We run our recognition pipeline sampling from the saliency maps produced by our detector on the videos in the UCF dataset, with the same baselines. %
%As with the Hollywood-2 dataset, we run four baselines: the Harris operator and the uniform, central bias and ground truth saliency map random sampling operators.
%Because this dataset is much smaller, randomness could have a significant impact on performance. We run all pipelines involving random interest point sampling operators three times, with different seeds, and report standard deviations for classification accuracy.\\

\noindent\textbf{Results}: On both datasets, our saliency map based pipelines -- both predicted and ground truth -- perform markedly better than the pipeline sampling interest points uniformly (Tables \ref{t:classification}c and \ref{t:classification_ucfsa}). Although central bias is a relatively close approximation of human visual saliency on our datasets, it does not lead to performance that closely matches the one produced by these maps. Our automatic pipeline based on predicted saliency maps and our ground-truth based pipeline have similar performance, with a slight advantage being observed for predicted saliency in the case of Hollywood-2 and for ground truth saliency for UCF Sports (Tables \ref{t:classification}d,e and \ref{t:classification_ucfsa}).

Our results confirm that approximations produced by the HoG-MBH detector are qualitatively different from a central bias distribution, focusing on local image structure that frequently co-occurs in its training set of fixations, disregarding on whether it is close to the image center or not (see Fig.\ref{fig:cb_det_comparison}). These structures are highly likely to be informative for predicting the actions in the dataset (\textit{e.g.} the frames containing the \emph{eat} and \emph{hug} actions in Fig.\ref{fig:cb_det_comparison}). Therefore, the detector will tend to emphasize these locations as opposed to less relevant ones. This also explains why our predicted saliency maps can be as informative for action recognition as ground truth maps, even exceeding their performance on certain action classes: while humans will also fixate on structures not relevant for action recognition, fixated structures that are relevant to this task will occur at a higher frequency in our datasets. Hence, they will be well approximated by a detector trained in a bottom-up manner. This can explain why the performance ballance is even more inclined towards predicted saliency maps on the UCF Sports Actions dataset, where motion and image patterns are more stable and easier to predict compared to the Hollywood-2 dataset.

Finally, we note that even though our pipeline is sparse, it achieves near state of the art performance when compared to a pipeline that uses dense trajectories. When the sparse descriptors obtained from our automatic pipeline were combined with the kernels associated to a dense trajectory representation \cite{WangEtAl2011} using an MKL framework with 18 kernels (14 kernels associated with sparse features sampled from predicted saliency maps + 4 kernels associated to dense trajectories from \cite{WangEtAl2011}), we were able to go beyond the state-of-the-art (Tables \ref{t:classification}f,i and \ref{t:classification_ucfsa}). This demonstrates that an end-to-end automatic system incorporating both human and computer vision technology can deliver high performance on a challenging problem such as action recognition in unconstrained video.

\begin{table}
\caption{Recognition results using second order pooling\cite{CarreiraEtAl2012} on the \textbf{UCF Sports Actions} Data Set}\label{t:classification_ucfsa_o2p}
\begin{center}
\scalebox{0.9}{
\begin{tabular}{|c|c|c|c|}
\hline \multirow{2}{*}{\textbf{method}}  & \multirow{2}{*}{\textbf{distribution}} & \multicolumn{2}{|c|}{\textbf{accuracy}}              \\
\cline{3-4}
                                         &                                     & \textbf{mean}              & \textbf{stdev}             \\
\hline                                   & Harris corners (a)                  & 84.3\%                     & 0.00\%                     \\
\cline{2-4}                              & uniform sampling (b)                & 83.9\%                     & 0.57\%                     \\
\cline{2-4} interest points              & central bias sampling (c)           & 83.7\%                     & 0.80\%                     \\
\cline{2-4}                              & predicted saliency sampling (d)     & 86.3\%                     & 0.66\%                     \\
\cline{2-4}                              & ground truth saliency sampling (e)  & 86.1\%                     & 0.63\%                     \\
\hline\multicolumn{2}{|c|}{trajectories\cite{WangEtAl2011} only (f)}           & 85.4\%                     & 0.00\%                     \\
\hline trajectories +                    & \multirow{2}{*}{predicted saliency sampling (h)}  & \textbf{\multirow{2}{*}{87.5\%}}    & \multirow{2}{*}{0.59\%}    \\
       interest points                   &                                     &                                                   &                            \\
\hline
\end{tabular}
}
\end{center}
Performance comparison among several classification methods (see table \ref{t:classification} for description) on the UCF Sports Dataset using second order pooling. Statistics are computed using 10 random seeds.
\end{table}

\noindent\textbf{Action recognition using second order pooling}: Our saliency based interest point operators are defined by randomly sampling specific spatio-temporal probability distributions. One way to estimate the variability induced in the recognition performance it to run the pipelines many times for different random seeds. Unfortunately, this is not practical, due to the high cost of training an end-to-end recognition pipeline in the bag-of-visual words framework. Consequently, results reported in tables \ref{t:classification} and \ref{t:classification_ucfsa} are obtained for a single random seed.

However, due to the large number of random samples used to train the system (\textit{e.g.} 28 million interest points for Hollywood-2), we expect little variance in performance for these datasets, with somewhat higher variability for UCF Sports due to its smaller size. To verify this intuition, we experiment with a faster version of the pipeline, in which we replace the vocabulary building and binning steps (the most computationally expensive) with \emph{second order pooling} as described in \cite{CarreiraEtAl2012}. To encode a particular video, we compute the covariance matrix of its descriptors and apply the matrix logarithm and power scaling operators, with an exponent of $0.5$, as used in \cite{CarreiraEtAl2012}. These operators map our feature sets into a high dimensional descriptor space and make the application of additional non-linear kernels unnecessary. For additional speed, we concatenate descriptors obtained for various HoG and MBH configurations into a single representation (as opposed to applying MKL) and use a fixed value of 10 for the SVM C penalty parameter.

We re-run each pipeline on the UCF Sports dataset with 10 different random seeds and compute the mean and standard deviation of the leave-one-out classification accuracy measure. Our results, shown in Table \ref{t:classification_ucfsa_o2p}, indicate that the standard deviation of the accuracy for the UCF Sports Actions dataset is less than $0.8\%$ for all pipelines. All the trends we found for the bag-of-visual words pipeline (Table \ref{t:classification_ucfsa}) are confirmed, with recognition results being somewhat lower, mainly due to the removal  of the expensive MKL step. We conclude that randomness has little impact on the performance of the pipelines presented in this paper.

\section{Conclusions}\label{s:conclusions}

We have presented experimental and computational modelling work at the incidence of human visual attention and computer vision, with emphasis on action recognition in video. Inspired by earlier psychophysics and visual attention findings, not validated quantitatively at large scale until now and not pursued for video, we have collected, and made available to the research community, a set of comprehensive human eye-tracking annotations for Hollywood-2 and UCF Sports, some of the most challenging, recently created action recognition datasets in the computer vision community. Besides the collection of large datasets, we have performed quantitative analysis and introduced novel models for evaluating the spatial and the sequential consistency of human fixations across different subjects, videos and actions. 
%We have found good inter-subject fixation agreement but, perhaps surprisingly, only moderate evidence of task influence.

We have also performed a large scale analysis of automatic visual saliency models and end-to-end automatic visual action recognition systems. Our studies are performed with particular focus on computer vision techniques and interest point operators and descriptors. In particular, we propose new accurate saliency operators that can be effectively trained based on human fixations. Finally, we show that such automatic saliency predictors can be used within end-to-end computer visual action recognition systems to achieve state of the art results in some of the hardest benchmarks in the field.

%We hope that our work will foster further communication, benchmarks and methodology sharing between human vision and computer vision, and ultimately lead to improved end-to-end artificial visual action recognition systems.

\ifCLASSOPTIONcompsoc
  \section*{Acknowledgments}
\else
  \section*{Acknowledgment}
\fi

This work was supported by CNCS-UEFICSDI, under PNII RU-RC-2/2009 and PCE-2011-3-0438.

% Left out CT-ERC-2012-1 for now.

\ifCLASSOPTIONcaptionsoff
  \newpage
\fi



%\bibliographystyle{IEEEtran}
\bibliographystyle{IEEEtran}
\bibliography{pami_video_eye}

\begin{thebibliography}{10}
\providecommand{\url}[1]{#1}
\csname url@samestyle\endcsname
\providecommand{\newblock}{\relax}
\providecommand{\bibinfo}[2]{#2}
\providecommand{\BIBentrySTDinterwordspacing}{\spaceskip=0pt\relax}
\providecommand{\BIBentryALTinterwordstretchfactor}{4}
\providecommand{\BIBentryALTinterwordspacing}{\spaceskip=\fontdimen2\font plus
\BIBentryALTinterwordstretchfactor\fontdimen3\font minus
  \fontdimen4\font\relax}
\providecommand{\BIBforeignlanguage}[2]{{%
\expandafter\ifx\csname l@#1\endcsname\relax
\typeout{** WARNING: IEEEtran.bst: No hyphenation pattern has been}%
\typeout{** loaded for the language `#1'. Using the pattern for}%
\typeout{** the default language instead.}%
\else
\language=\csname l@#1\endcsname
\fi
#2}}
\providecommand{\BIBdecl}{\relax}
\BIBdecl

\bibitem{MarszalekEtAl2009}
M.~Marszalek, I.~Laptev, and C.~Schmid, ``Actions in context,'' in \emph{IEEE
  International Conference on Computer Vision and Pattern Recognition}, 2009.

\bibitem{RodriguezEtAl2008}
M.~D. Rodriguez, J.~Ahmed, and M.~Shah, ``Action mach a spatio-temporal maximum
  average correlation height filter for action recognition,'' in \emph{IEEE
  International Conference on Computer Vision and Pattern Recognition}, 2008.

\bibitem{EveringhamEtAl2010}
M.~Everingham, L.~V. Gool, C.~Williams, J.~Winn, and A.~Zisserman, ``The
  {P}ascal visual object classes ({VOC}) challenge,'' \emph{International
  Journal on Computer Vision}, 2010.

\bibitem{Rolls2011}
E.~Rolls, \emph{Memory, Attention, and Decision-Making: A Unifying
  Computational Neuroscience Approach.}\hskip 1em plus 0.5em minus 0.4em\relax
  Oxford University Press, 2008.

\bibitem{JuddEtAl2009}
T.~Judd, K.~Ehinger, F.~Durand, and A.~Torralba, ``Learning to predict where
  humans look,'' in \emph{International Conference on Computer Vision}, 2009.

\bibitem{LarochelleHinton2010}
H.~Larochelle and G.~Hinton, ``Learning to combine foveal glimpses with a
  third-order boltzmann machine,'' in \emph{Advances in Neural Information and
  Processing Systems}, 2010.

\bibitem{Laptev2005}
I.~Laptev, ``On space-time interest points,'' in \emph{International Journal on
  Computer Vision}, 2005.

\bibitem{DongEtAl2009}
D.~Han, L.~Bo, and C.~Sminchisescu, ``Selection and context for action
  recognition,'' in \emph{International Conference on Computer Vision}, 2009.

\bibitem{SminchisescuTriggs2005}
C.~Sminchisescu and B.~Triggs, ``Building roadmaps of minima and transitions in
  visual models,'' \emph{International Journal on Computer Vision}, vol.~61,
  no.~1, 2005.

\bibitem{MatheSmi12}
S.~Mathe and C.~Sminchisescu, ``Dynamic eye movement datasets and learnt
  saliency models for visual action recognition,'' in \emph{European Conference
  on Computer Vision}, 2012.

\bibitem{Yarbus1967}
A.~Yarbus, \emph{Eye Movements and Vision}.\hskip 1em plus 0.5em minus
  0.4em\relax New York Plenum Press, 1967.

\bibitem{IttiEtAl1998}
L.~Itti, C.~Koch, and E.~Niebur, ``A model of saliency-based visual attention
  for rapid scene analysis,'' \emph{IEEE Transactions on Pattern Analysis and
  Machine Intelligence}, vol.~20, 1998.

\bibitem{IttiReesTsotsos2005}
L.~Itti, G.~Rees, and J.~K. Tsotsos, Eds., \emph{Neurobiology of
  Attention}.\hskip 1em plus 0.5em minus 0.4em\relax Academic Press, 2005.

\bibitem{LandTatler2009}
M.~F. Land and B.~W. Tatler, \emph{Looking and Acting}.\hskip 1em plus 0.5em
  minus 0.4em\relax Oxford University Press, 2009.

\bibitem{IttiKoch2000}
L.~Itti and C.~Koch, ``A saliency-based search mechanism for overt and covert
  shifts of visual attention,'' \emph{Vision Research}, vol.~40, 2000.

\bibitem{BruceTsotsos2005}
N.~Bruce and J.~Tsotsos, ``Saliency based on information maximization,'' in
  \emph{Advances in Neural Information and Processing Systems}, 2005.

\bibitem{KienzleEtAl2006}
W.~Kienzle, F.~Wichmann, B.~Scholkopf, and M.~Franz, ``A nonparametric approach
  to bottom-up visual saliency,'' in \emph{Advances in Neural Information and
  Processing Systems}, 2006.

\bibitem{TorralbaEtAl2006}
A.~Torralba, A.~Oliva, M.~Castelhano, and J.~Henderson, ``Contextual guidance
  of eye movements and attention in real-world scenes: The role of global
  features in object search,'' \emph{Psychological Review}, vol. 113, 2006.

\bibitem{JuddEtAl2011}
T.~Judd, F.~Durand, and A.~Torralba, ``Fixations on low resolution images,'' in
  \emph{International Conference on Computer Vision}, 2009.

\bibitem{BorjiItti2011}
A.~Borji and L.~Itti, ``Scene classification with a sparse set of salient
  regions,'' in \emph{IEEE International Conference on Robotics and
  Automation}, 2011.

\bibitem{ElazaryItti2010}
L.~Elazary and L.~Itti, ``A {B}ayesian model for efficient visual search and
  recognition,'' \emph{Vision Research}, vol.~50, 2010.

\bibitem{EhingerEtAl2009}
K.~A. Ehinger, B.~Sotelo, A.~Torralba, and A.~Oliva, ``Modeling search for
  people in 900 scenes: A combined source model of eye guidance,'' \emph{Visual
  Cognition}, vol.~17, 2009.

\bibitem{SerreEtAl2007}
T.~Serre, L.~Wolf, S.~Bileschi, M.~Riesenhuber, and T.~Poggio, ``Robust object
  recognition with cortex-like mechanisms,'' \emph{IEEE Transactions on Pattern
  Analysis and Machine Intelligence}, vol.~29, 2007.

\bibitem{WaltherEtAl2005}
D.~Walther, U.~Rutishauser, C.~Koch, and P.~Perona, ``Selective visual
  attention enables learning and recognition of multiple objects in cluttered
  scenes,'' \emph{Computer Vision and Image Understanding}, vol. 100, 2005.

\bibitem{HanVasconcelos2010}
S.~Han and N.~Vasconcelos, ``Biologically plausible saliency mechanisms improve
  feedforward object recognition,'' \emph{Vision Research}, 2010.

\bibitem{SerreEtAlICCV2007}
H.~Jhuang, T.~Serre, L.~Wolf, and T.~Poggio, ``A biologically inspired system
  for action recognition,'' in \emph{International Conference on Computer
  Vision}, 2007.

\bibitem{KienzleEtAl2007}
W.~Kienzle, B.~Scholkopf, F.~Wichmann, and M.~Franz, ``How to find interesting
  locations in video: a spatiotemporal interest point detector learned from
  human eye movements,'' in \emph{Lecture Notes in Computer Science (DAGM)},
  2007.

\bibitem{MaratEtAl2007}
S.~Marat, M.~Guironnet, and D.~Pellerin, ``Video summarization using a visual
  attention model,'' in \emph{European Signal Processing Conference}, 2007.

\bibitem{LeMeurEtAl2007}
O.~Le~Meur, P.~Le~Callet, and D.~Barba, ``Predicting visual fixations on video
  based on low-level visual features,'' \emph{Vision Research}, vol.~47, pp.
  2483--2498, 2007.

\bibitem{FathiLiRehg2012}
A.~Fathi, Y.~Li, and J.~M. Rehg, ``Learning to recognize daily actions using
  gaze,'' in \emph{European Conference on Computer Vision}, 2012.

\bibitem{WinklerSubramanian2013}
S.~Winkler and R.~Subramanian, ``Overview of eye tracking datasets,'' in
  \emph{International Workshop on Quality of Multimedia Experience (QoMEX)},
  2013.

\bibitem{FeiFeiEtAl2007}
L.~Fei-Fei, A.~Iyer, C.~Koch, and P.~Perona, ``What do we perceive in a glance
  of a real-world scene?'' \emph{Journal of Vision}, 2007.

\bibitem{JhuangEtAl2007}
H.~Jhuang, T.~Serre, L.~Wolf, and T.~Poggio, ``A biologically inspired system
  for action recognition,'' in \emph{International Conference on Computer
  Vision}, 2007.

\bibitem{WangEtAl2011}
H.~Wang, A.Klaser, C.Schmid, and C.~Liu, ``Action recognition by dense
  trajectories,'' in \emph{IEEE International Conference on Computer Vision and
  Pattern Recognition}, 2011.

\bibitem{YamatoEtAl1992}
J.~Yamato, J.~Ohya, and K.~Ishii, ``Recognizing human action in time-squential
  images using hidden markov model,'' in \emph{IEEE International Conference on
  Computer Vision and Pattern Recognition}, 1992.

\bibitem{LiEtAl2008}
W.Li, Z.Zhang, and Z.Liu, ``Expandable data-driven graphical modeling of human
  actions based on salient postures,'' \emph{IEEE TCSVT}, vol.~18, 2008.

\bibitem{HoaiDeLaTorre2012}
M.~Hoai and F.~{De la Torre}, ``Max-margin early event detectors,'' in
  \emph{IEEE International Conference on Computer Vision and Pattern
  Recognition}, 2012.

\bibitem{DorrEtAl2010}
M.~Dorr, T.~Martinetz, K.~Gegenfurtner, and E.~Barth, ``Variability of eye
  movements when viewing dynamic natural scenes,'' \emph{Journal of Vision},
  vol.~10, 2010.

\bibitem{GreeneEtAl2012}
M.~R. Greene, T.~Liu, and J.~M. Wolfe, ``Reconsidering yarbus: A failure to
  predict observers' task from eye movement patterns,'' \emph{Vision Research},
  vol.~62, 2012.

\bibitem{LeMeurBaccino2012}
O.~Le~Meur and T.~Baccino, ``Methods for comparing scanpaths and saliency maps:
  strengths and weaknesses,'' \emph{Behavior Research Methods}, 2012.

\bibitem{YaoFei2010}
B.~Yao and L.~Fei-Fei, ``Modeling mutual context of object and human pose in
  human-object interaction activities,'' in \emph{IEEE International Conference
  on Computer Vision and Pattern Recognition}, 2010.

\bibitem{PrestEtAl2011}
A.~Prest, C.~Schmid, and V.~Ferrari, ``Weakly supervised learning of
  interactions between humans and objects,'' \emph{IEEE Transactions on Pattern
  Analysis and Machine Intelligence}, 2011.

\bibitem{HwangEtAl2011}
A.~Hwang, H.~Wang, and M.~Pomplun, ``Semantic guidance of eye movements in
  real-world scenes,'' \emph{Vision Research}, 2011.

\bibitem{NeedleManEtAl1970}
B.~Needleman and C.~Wunsch, ``A general method applicable to the search for
  similarities in the amino acid sequence of two proteins,'' \emph{Journal of
  Molecular Biology}, 1970.

\bibitem{LaptevLindeberg2003}
I.~Laptev and T.~Lindeberg, ``Space-time interest points,'' in
  \emph{International Conference on Computer Vision}, 2003.

\bibitem{DalalEtAl2006}
N.~Dalal, B.Triggs, and C.Schmid, ``Human detection using oriented histograms
  of flow and appearance,'' in \emph{European Conference on Computer Vision},
  2006.

\bibitem{DalalTriggs2005}
N.~Dalal and B.~Triggs, ``Histograms of oriented gradients for human
  detection,'' in \emph{IEEE International Conference on Computer Vision and
  Pattern Recognition}, 2005.

\bibitem{HariharanEtAl2010}
B.~Hariharan, L.~Zelnik-Manor, S.~V.~N. Vishwanathan, and M.~Varma, ``Large
  scale max-margin multi-label classification with priors,'' in
  \emph{International Conference on Machine Learning}, 2010.

\bibitem{JuddEtAl2012}
T.~Judd, F.~Durand, and A.~Torralba, ``A benchmark of computational models of
  saliency to predict human fixations,'' MIT, Tech. Rep.~1, 2012.

\bibitem{OlivaTorralba2001}
A.~Oliva and A.~Torralba, ``Modeling the shape of the scene: A holistic
  representation of the spatial envelope,'' \emph{International Journal on
  Computer Vision}, no.~42, 2001.

\bibitem{Rosenholtz1999}
R.~Rosenholtz, ``A simple saliency model predicts a number of motion popout
  phenomena,'' \emph{Vision Research}, no.~39, 1999.

\bibitem{ViolaJones2001}
P.~Viola and M.~Jones, ``Robust real-time object detection.''
  \emph{International Journal on Computer Vision}, 2001.

\bibitem{FelzenswalbEtAl2008}
P.~Felzenswalb, D.~McAllester, and D.~Ramanan, ``A discriminatively trained,
  multiscale, deformable part model,'' in \emph{IEEE International Conference
  on Computer Vision and Pattern Recognition}, 2008.

\bibitem{SunEtAl2010}
D.~Sun, S.~Roth, and M.~J. Black, ``Secrets of optical flow and their
  principles,'' in \emph{IEEE International Conference on Computer Vision and
  Pattern Recognition}, 2010.

\bibitem{ArbelaezEtAl2011}
P.~Arbelaez, M.~Maire, C.~Fowlkes, and J.~Malik, ``Occlusion boundary detection
  and figure/ground assignment from optical flow,'' \emph{IEEE Transactions on
  Pattern Analysis and Machine Intelligence}, 2011.

\bibitem{VedaldiZisserman2011}
A.~Vedaldi and A.~Zisserman, ``Efficient additive kernels via explicit feature
  maps,'' \emph{IEEE Transactions on Pattern Analysis and Machine
  Intelligence}, 2011.

\bibitem{LibLinear}
R.-E. Fan, K.-W. Chang, C.-J. Hsieh, X.-R. Wang, and C.-J. Lin, ``{LIBLINEAR}:
  A library for large linear classification,'' \emph{JMLR}, vol.~9, 2008.

\bibitem{SimocelliFreeman1995}
E.~Simocelli and W.~Freeman, ``The steerable pyramid: A flexible architecture
  for multi-scale derivative computation,'' in \emph{IEEE International
  Conference on Image Processing}, 1995.

\bibitem{CarreiraEtAl2012}
J.~Carreira, R.~Caseiro, J.~Batista, and C.~Sminchisescu, ``Semantic
  segmentation with second-order pooling,'' in \emph{European Conference on
  Computer Vision}, 2012.

\end{thebibliography}
%\begin{biography}[{\includegraphics[width=1in,height=1.25in,clip,keepaspectratio]{mshell}}]{Michael Shell}

\vspace{-10mm}

\begin{IEEEbiography}[{\scalebox{0.65}{\includegraphics{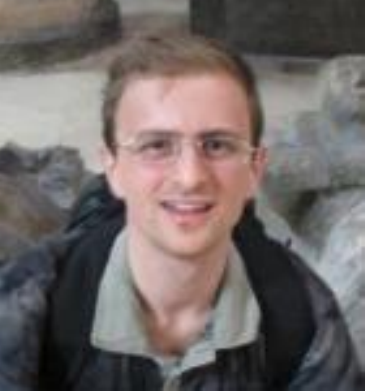}}}]{Stefan Mathe} has obtained his MSc. degree in Computer Science from the Technical University of Cluj Napoca, Romania.
He is currently a Research Assistant at the Institute of Mathematics of the Romanian Academy and a Phd student in the Artificial Intelligence Laboratory at the University of Toronto. His research interests span the areas of both low level and high level computer vision, machine learning, natural language understanding and computer graphics. His Phd research is focused on the problems of action recognition, with emphasis of developing higher level invariant features and exploiting contextual information. 
\end{IEEEbiography}
\vspace{-10mm}
\begin{IEEEbiography}[{\scalebox{0.45}{\includegraphics{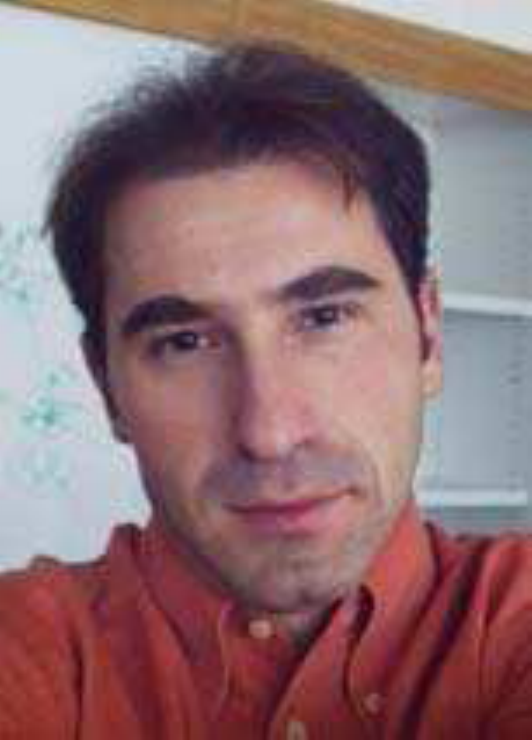}}}]{Cristian Sminchisescu} has obtained a doctorate in Computer Science and Applied Mathematics with an emphasis on imaging, vision
and robotics at INRIA, France, under an Eiffel excellence doctoral fellowship, and has done postdoctoral research in the Artificial Intelligence
Laboratory at the University of Toronto. He is a member in the program committees of the main conferences in computer vision and machine learning (CVPR, ICCV, ECCV, NIPS, AISTATS), area chair for ICCV07-13, and a member of
the Editorial Board (Associate Editor) of IEEE Transactions on Pattern Analysis and Machine Intelligence (PAMI). He has given more than 50 invited talks and presentations and has offered tutorials on 3d tracking,
recognition and optimization at ICCV and CVPR, the Chicago Machine Learning Summer School, the AERFAI Vision School in Barcelona and the Computer Vision Summer School (VSS) in Zurich. Sminchisescu's
research goal is to train computers to see. His research interests are in the area of computer vision (articulated objects, 3d reconstruction, segmentation, and object and action recognition) and machine learning
(optimization and sampling algorithms, structured prediction, sparse approximations and kernel methods). 
%Recent work by himself and his collaborators has produced state-of-the art results in the monocular 3d
%human pose estimation benchmark (HumanEva) and was the winner of the PASCAL VOC object segmentation and labeling challenge, over the past four editions 2009-2012.
\end{IEEEbiography}

%\vfill


\end{document}